\documentclass{article}

    \usepackage[preprint]{neurips_2023}

\usepackage[utf8]{inputenc} 
\usepackage[T1]{fontenc}    
\usepackage{hyperref}       
\usepackage{url}            
\usepackage{booktabs}       
\usepackage{amsfonts}       
\usepackage{nicefrac}       
\usepackage{microtype}      
\usepackage[table]{xcolor}
\usepackage{tabularx}
\usepackage{multirow}
\definecolor{LightCyan}{rgb}{0.88,1,1}
\usepackage{adjustbox}
\usepackage{subcaption}
\usepackage{wrapfig,lipsum}
\usepackage{hyperref}
\usepackage{todonotes}
\usepackage{amsmath}
\newtheorem{definition}{Definition}
\usepackage{titletoc}

\newcolumntype{C}{>{\centering\arraybackslash}X}

\hypersetup{
    bookmarks=true,         
    unicode=false,          
    pdftoolbar=true,        
    pdfmenubar=true,        
    pdffitwindow=false,     
    pdfstartview={FitH},    
    pdftitle={My title},    
    pdfauthor={Author},     
    pdfsubject={Subject},   
    pdfcreator={Creator},   
    pdfproducer={Producer}, 
    pdfkeywords={keyword1, key2, key3}, 
    pdfnewwindow=true,      
    colorlinks=false,       
    linkcolor=red,          
    citecolor=green,        
    filecolor=magenta,      
    urlcolor=cyan,           
}
\usepackage{tabularx}
\usepackage{cleveref}
\usepackage{pifont}
\newcommand{\cmark}{\ding{51}}%
\newcommand{\xmark}{\ding{55}}%

\title{Differentially Private Tabular Data Synthesis \\ using Large Language Models}

\author{%
  Toan V. Tran\\
  Emory University, GA, USA\\
  \texttt{vtran29@emory.edu} \\
  \And
  Li Xiong \\
  Emory University, GA, USA \\
  \texttt{lxiong@emory.edu} \\
}

\begin{document}

\maketitle

\begin{abstract}

Synthetic tabular data generation with differential privacy is a crucial problem to enable data sharing with formal privacy.  Despite a rich history of methodological research and development, developing differentially private tabular data generators that can provide realistic synthetic datasets remains challenging. This paper introduces \texttt{DP-LLMTGen} -- a novel framework for differentially private tabular data synthesis that leverages pretrained large language models (LLMs). \texttt{DP-LLMTGen} models sensitive datasets using a two-stage fine-tuning procedure with a novel loss function specifically designed for tabular data. Subsequently, it generates synthetic data through sampling the fine-tuned LLMs. Our empirical evaluation demonstrates that \texttt{DP-LLMTGen} outperforms a variety of existing mechanisms across multiple datasets and privacy settings. Additionally, we conduct an ablation study and several experimental analyses to deepen our understanding of LLMs in addressing this important problem. Finally, we highlight the controllable generation ability of \texttt{DP-LLMTGen} through a fairness-constrained generation setting.

\end{abstract}

\section{Introduction}
Tabular data is one of the most popular forms of data for data analytic applications \citep{9998482}. In many domains such as healthcare and finance, real data may be privacy sensitive and can not be directly shared and utilized. To address this issue, synthetic data, generated from real datasets, enables the use of realistic datasets without exposing real-world data. However, synthetic data can still retain sample-level patterns and details from the original datasets. This poses privacy risks for reidentification and attribute disclosure by membership inference or by linking attributes, especially with sensitive information \citep{zhu2024quantifying}. A promising approach is to integrate differential privacy (DP), which introduces  randomness in the data generation \citep{dp}. This process ensures that the data remain useful for analysis but significantly reduce the risk of re-identifying individuals \citep{Cummings2024Advancing, zhu2024quantifying}. Therefore, synthetic data generation with differential privacy is an important problem that has  received growing attention by the research community \citep{pmlr-v139-aydore21a, Vietri2022PrivateSD, gsd, liu2021iterative, yoon2018pategan}.

Generating synthetic tabular data in both non-DP and DP settings have been extensively investigated. One notable approach involves deep learning-based Generative Adversarial Networks (GANs). \citet{ctgan} has designed a suitable variant of GANs for tabular data (named CTGAN) which outperformed previous statistical methods \citep{1054142, JunBayes}. However, the utility of GANs struggles under DP settings when the models are trained by differentially private stochastic gradient descent (DPSGD) \citep{dpsgd} due to the complexity of the model architecture and adversarial training of GANs. Consequently, an alternative approach has emerged by leveraging marginal distributions. This approach privately measures marginal distributions and then generates synthetic datasets from those distributions \citep{Ryanaim, liu2021iterative, Vietri2022PrivateSD, pmlr-v139-aydore21a}. The marginal-based methods significantly outperformed GAN-based methods and created a notable gap between marginal-based and deep learning-based methods.

Furthermore, inspired by the development of large language models (LLMs), \citet{borisov2023language} proposed a method that leverages pre-trained LLMs  for synthesizing tabular data in non-private settings. Compared to previous methods, pre-trained LLMs with background knowledge offer a basic understanding of contextual knowledge. For example, in the Adult dataset \cite{AdultIncomeDataset1994}, features like gender and occupation clearly have a correlation, e.g., a mechanic is more likely to be a male. This basic knowledge can be already gained by LLMs during their pretrained stage \citep{dong2023statistical}. As a result, LLMs outperformed GANs for synthesizing tabular data in non-DP settings \citep{borisov2023language}. However, whether LLMs can operate effectively under DP settings remains an open question.

In this paper, we demonstrate that naively enhancing the existing approach from \citep{borisov2023language} by DPSGD do not work well. This is mainly because such conventional DP fine-tuned LLMs fail at generating tabular data with format compliance due to the injected noise. Additionally, the standard cross entropy loss used in causal language modeling does not work well for tabular data. To tackle the challenges, we propose \texttt{DP-LLMTGen} (an abbreviation of \underline{D}ifferentially \underline{P}rivate \underline{L}arge \underline{L}anguage \underline{M}odels for \underline{T}abular data \underline{Gen}eration), which is the first study leveraging pretrained LLMs for synthetic tabular generation under DP settings.

In summary, our contributions are as the follows. 1) \textit{Technical novelty}: we propose a two-stage fine-tuning mechanism using a novel loss function. Our fine-tuning mechanism distinguishes learning targets (format compliance and data modeling) and treats them suitably.
The loss function is designed to align with the proposed fine-tuning procedure and improve numerical understanding of LLMs. 2) \textit{Empirical evaluation}: \texttt{DP-LLMTGen} outperforms marginal-based methods in tight differential privacy settings, closing the gap between marginal-based and deep learning-based methods. Additionally, our ablation study demonstrates the two-stage fine tuning is necessary for format compliance and the proposed loss function helps to improve their numerical understanding and boost the overall performance. We also perform several experimental analyses. Our findings include:~$\bullet$~Data contamination between the pre-training and experimental datasets of LLama-2 7b models is not a significant problem. $\bullet$~LLMs incorporate feature names to enable context-aware learning for tabular data synthesis. $\bullet$~Models without dialogue optimization generate better synthetic datasets. $\bullet$~Larger models are not always better. 3) \textit{Controllable generation}: we empirically demonstrate the effectiveness of controllable generation of \texttt{DP-LLMTGen} in a fairness setting. Our controllable generator can significantly reduce biases with only a minor utility tradeoff.

\section{Related Works}
\textbf{Synthetic tabular data generation}. This field has a rich history of methodological research and development. Classical methods treat each feature as a variable to model the data as a joint multivariate probability distribution, from which synthetic samples are drawn. Various modeling methods have been employed, such as trees \citep{1054142}, Bayesian networks \citep{Jimbnm}, and copulas \citep{Kamthe2021CopulaFF}. With the advent of advanced neural networks, Variational Autoencoders (VAEs) \citep{AkramiVAE, ctgan}, Generative Adversarial Networks (GANs) \citep{NoseongGAN, ctgan}, and diffusion models \citep{AkimTabDDPM} have enhanced the modeling of high-dimensional dependencies. Recent studies have explored transformer-based architectures \citep{NIPS2017_3f5ee243, badaro-etal-2023-transformers}. \cite{gulati2023tabmt} conducted masked pretraining for encoder-only models. \cite{canale2022generative} trained decoder-only models from scratch. Concurrently, \cite{borisov2023language} fine-tuned LLMs (GPT-2 \citep{radford2019language} and distilled GPT-2 \citep{sanh2019distilbert}) which have been pre-trained on a massive amount of text data.

\textbf{Differentially private tabular data generation.} The advancements in standard synthetic data generation have led to various mechanisms that can ensure differential privacy. Classical methods have been enhanced to achieve the DP property, e.g., private Bayesian networks \citep{JunBayes} and DP-copula \citep{Li2014DifferentiallyPS}. Additionally, many neural networks have been trained by differentially private stochastic gradient descent (DPSGD) \citep{dpsgd} such as DP-GAN \citep{Xie2018DifferentiallyPG} and DP-Conditonal GAN \citep{9025394}. To improve privacy guarantees, \cite{yoon2018pategan} trained GANs by the PATE framework \citep{Papernot2018ScalablePL} which utilizes non-labeled public data. Despite GANs' effectiveness in non-DP settings, they do not align well with DPSGD due to the complexity of the model architecture and adversarial training of GANs. Consequently, marginal-based methods have emerged as more effective under DP constraints \citep{Ryanaim, pmlr-v139-aydore21a, Vietri2022PrivateSD, gsd}. Recent efforts involve training from scratch transformer-based networks with DPSGD \citep{Castellon2023DPTBARTAT, sablayrolles2023privately}. However, training from scratch usually requires significant privacy budgets and thus still lags behind marginal-based methods in most cases. Our study is the first to leverage pre-trained LLMs and introduces a novel fine-tuning approach for tight DP settings. \textit{We provide additional related works and clarify the differences in Appendix~\ref{sec:add_related_work}.}

\section{Preliminaries}
\textbf{Differential Privacy (DP).} Differential privacy is a golden privacy notation for quantifying and bounding the privacy risks of releasing data statistics \citep{dp}. Informally, DP ensures the outputs of computational algorithms do not change significantly when a single record is added to or removed from the dataset. In this work, we use a popular notion of differential privacy -- $(\epsilon, \delta)$-DP.

\begin{definition}
  A mechanism $\mathcal{M}: {D} \rightarrow {S}$ is $(\epsilon, \delta)$-DP if for any two neighbouring datasets D and D', and for any subset of output responses $S \in Range(\mathcal{M})$,
  \[ P[\mathcal{M}(D)) \in S]  \leq e^\epsilon P[\mathcal{M}(D')) \in S] + \delta \]
\end{definition}

Smaller values of $\epsilon$ and $\delta$ (close to zero) provide stronger privacy guarantees. The post-processing property of DP mechanism ensures that any computations, which are applied to the output of $\mathcal{M}$ and do not access the original sensitive data, remain the same DP guarantee.

\section{\texttt{DP-LLMTGen}: \underline{D}ifferentially \underline{P}rivate \underline{LLM}-based \underline{T}abular data \underline{Gen}erators}
\begin{figure}[htp]
    \centering
    \includegraphics[width=\textwidth]{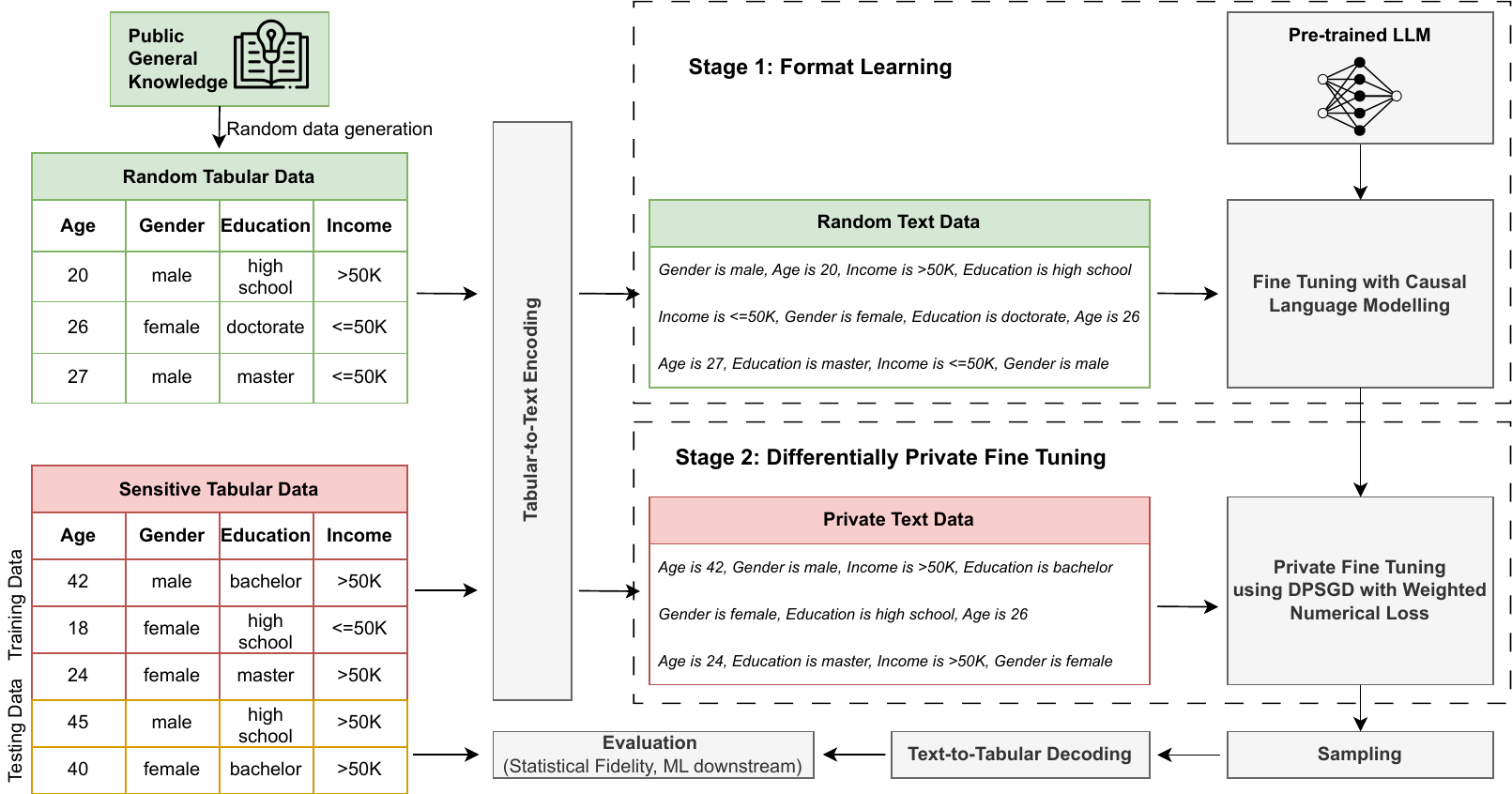}
    \caption{The process flow of \texttt{DP-LLMTGen}}
    \label{fig:process-flow}
    \vspace{-4mm}
\end{figure}

The previous work~\citep{borisov2023language} has leveraged pretrained LLMs for non-private tabular data synthesis by simply converting tabular data to text and fine tuning with standard causal language modeling. However, this conventional fine-tuning approach fails under DP training due to format compliance issues. To address this, we introduce DP-LLMTGen, as shown in Figure~\ref{fig:process-flow}. Initially, a random tabular dataset is generated using public general knowledge. Subsequently, both the original sensitive tabular dataset and the random tabular dataset are converted to textual data by the tabular-to-text encoding. Next, the LLM undergoes the two-stage fine-tuning process. Synthetic data are then generated by sampling from the fine-tuned LLM and converted back into tabular format through the text-to-tabular decoding. 

\subsection{Data processing}
\textbf{Random Tabular Data Generation.} In this paper, we assume that general knowledge, including feature names, categorical names, and numerical ranges, are not sensitive. Instead, the actual sensitive information comprises individual records. Consequently, we propose the generation of a dataset (referred to as "random tabular data") that lacks realistic feature distributions and exhibits random dependencies among features. The sole requirements for this random data are accurate feature names, correct categorical names, and numerically valid ranges. It is worth noting that this assumption is similar to previous transformer-based works \citep{Castellon2023DPTBARTAT, sablayrolles2023privately} which require creating a specialized vocabulary.

\textbf{Tabular-to-Text Encoding.} We use a common template -- \texttt{"\{feature\_1 name\} is \{value\}, \{feature\_2 name\} is \{value\}, etc."} which has been used in some previous works~\citep{borisov2023language, Dinh2022LIFTLF, hegselmann2023tabllm}. Moreover, we permute the features randomly to remove the feature-order dependencies which naturally does not exist in tabular data \citep{9998482, borisov2023language}. This permutation step makes the sampling step more flexible and controllable. Conversely, Text-to-Tabular Decoding is the reverse process of Tabular-to-Text Encoding.

\subsection{Two-stage Fine Tuning}
Our fine tuning include two stages: 1) Format Learning and 2) Differentially Private Fine Tuning. This procedure helps LLMs to ensure format compliance while achieving DP properties. We implement a parameter-efficient fine tuning method -- LoRA \citep{hu2022lora} for both stages.

\textbf{Stage 1: Format Learning.} This stage utilizes the random data that does not require any privacy protection. Therefore, we apply standard causal language modeling for fine tuning the pretrained LLM. This stage helps the LLM to learn the format, feature names, corresponding categorical names, and numerical ranges without spending any privacy budget. Let $\{t_1, t_2, t_3, ...\}$ denote the sequence of tokens $\mathbb{S}^{(i)}$, where each token is an element in vocabulary $V$. The loss function for this stage is the standard cross entropy loss across all tokens, shown in Equation~\ref{eq:loss-1}, where $N$ is the number of sequences, $\theta$ is the model parameters, and $p\left( t_j | t_{j-1}, t_{j-2}, ..., t_{1}, \theta \right)$ denotes the model output probability of predicting token $t_j$ given the previous tokens.

{ \small
\begin{equation}
    \mathcal{L}(\theta) = -\dfrac{1}{N} \sum_{i=1}^N \sum_{t_{j} \in \mathbb{S}^{(i)}} \log p\left( t_j | t_{j-1}, t_{j-2}, ..., t_{1}, \theta \right)
    \label{eq:loss-1}
\end{equation}
}

\textbf{Stage 2: Differentially Private Fine Tuning.} This stage aims to learn the feature distributions and dependencies of the actual datasets. To protect the sensitive data, the LLM is fine-tuned by  differential private stochastic gradient descent (DPSGD) \citep{dpsgd}.  To enhance the performance, we propose a novel loss function (denoted as $\mathcal{L}(\theta)$ in Equation \ref{eq:loss}) which includes two main components: \textit{Weighted Cross Entropy Loss (WCEL)} and \textit{Numerical-Understanding Loss (NUL)}. 

\underline{\textit{WCEL:}} Given a sequence of tokens $\mathbb{S}^{(i)} = \{t_1, t_2, t_3, ...\}$, the WCEL discriminates between format tokens $\mathbb{F}^{(i)}$ and tabular tokens $\mathbb{T}^{(i)}$. More specifically, words (such as feature names and "is") and commas are format tokens while tabular tokens are derived from the numerical and categorical values. For instance, considering the sentence \texttt{"Age is 20, Education is high school"}, the format and tabular tokens are \texttt{"Age is , Education is"} and \texttt{"20 high school"}, respectively. This discrimination helps the LLMs to efficiently focus on learning the feature distributions and dependencies instead of format compliance, which has been achieved in the first stage. The WCEL is defined in Equation~\ref{eq:wcel}.

{ \small
\begin{equation}
    \mathcal{L}_{\text{WCE}}(\mathbb{S}^{(i)}, \theta) = (1 - \alpha)\sum_{t_j \in \mathbb{F}^{(i)}} - \log p\left( t_j | t_{j-1}, ..., t_{1}, \theta \right) +  \alpha \sum_{t_j \in \mathbb{T}^{(i)}} - \log p\left( t_j | t_{j-1}, ..., t_{1}, \theta \right)
    \label{eq:wcel}
\end{equation}
}
\underline{\textit{NUL:}} There is a fundamental problem of cross entropy loss while dealing with numbers. The cross entropy loss does not differentiate well between numerical errors in different contexts. For instance, considering a ground truth of \texttt{10.0}, the cross entropy loss treats identically between predicting \texttt{10.1} and \texttt{99.9}. To address this, our Numerical-Understanding Loss considers the exact difference between the prediction and the ground truth numbers. Let $\hat{t}_j$ be the highest-probability prediction token for the token at index $j$, formalized in Equation~\ref{eq:llm-pred}, where $w$ is a token in the model vocabulary; $\theta$ is the model parameters; $f$ represents the model forwarding. Subsequently, Equation~\ref{eq:decoding} represents numeric token decoding, which converts the consecutive tokens from index $j$ to $k$ into a number, denoted by $\hat{n}_{\{j,k\}}$. For each sequence of tokens $\mathbb{S}^{(i)}$, $\mathbb{N}^{(i)}$ is the set of all pairs $\{j, k\}$ which represents the starting and ending indices of tabular numerical values. The NUL (Equation~\ref{eq:nul}) is the sum of the squared errors ($SE$ -- defined in Equation~\ref{eq:mse}) between predicted numerical values $\hat{n}_{j, k}$ and their corresponding ground truth values $n_{j, k}$. Notably, the numeric token decoding can fail if the prediction tokens do not form a number. In success cases, the errors are scaled with a factor $\lambda$. For failed cases, the error is set to 1.0, which is typically higher than the success squared errors to enhance format compliance.

{ \small
\begin{equation}
   \hat{t}_j = {\text{argmax}_{w} \left[f(w|t_{j-1}, t_{j-2}, ..., t_{1}, \theta)) \right]}
   \label{eq:llm-pred}
\end{equation}

\begin{equation}
   \hat{n}_{\{j,k\}} = \text{decoding(}\{\hat{t}_j; \hat{t}_{j+1}; ... \hat{t}_k\} \text{)}
   \label{eq:decoding}
\end{equation}

\begin{equation}
SE (\{j, k\}, \theta) =
    \begin{cases}
        1.0 & \text{if decoding failed} \\
        \dfrac{1}{2} \left( \dfrac{n_{\{j,k\}} - \hat{n}_{\{j,k\}}}{\lambda}  \right) ^ 2, & \text{otherwise}
    \end{cases}
    \label{eq:mse}
\end{equation}

\begin{equation}
    \mathcal{L}_{\text{NU}}(\mathbb{S}^{(i)}, \theta) = \sum_{\{j;k\} \in \mathbb{N}^{(i)}} SE(\{j, k\}, \theta)
    \label{eq:nul}
\end{equation}
}
The final loss function is the weighted sum of $\mathcal{L}_{\text{WCE}}$ and $\mathcal{L}_{\text{NU}}$ over $N$ sentences:
{ \small
\begin{equation}
    \mathcal{L}(\theta) = \dfrac{1}{N} \sum_{i=1}^N \left[ \mathcal{L}_{\text{WCE}}(\mathbb{S}^{(i)}, \theta) + \beta \mathcal{L}_{\text{NU}}(\mathbb{S}^{(i)}, \theta) \right]
    \label{eq:loss}
\end{equation}
}

\vspace{-8mm}
\subsection{Sampling}
\vspace{-2mm}
To generate synthetic samples, we initialize short prompts and ask the fine-tuned LLMs to generate the remainder. Following the approach outlined by \cite{borisov2023language}, our LLMs facilitate both types of sampling: random-initialized and value-specified. In random-initialized sampling, the prompts consist of a randomly chosen feature name, such as \texttt{"Age is"} or \texttt{"Workclass is"}. In contrast, value-specified sampling involves initializing prompts with specific feature values to enable controllable generation, e.g., \texttt{"Age is 26,"} or \texttt{"Income is >50K, Sex is female,"}. The fine-tuned LLMs can complete the remainder of the sentences. The full sentences can be converted back to tabular format by Text-to-Tabular Decoding, which is just a simple string processing algorithm.
\vspace{-3mm}
\begin{table}[ht!]
    \centering
    \caption{Comparative performance of \texttt{DP-LLMTGen} against the baselines across five datasets. For each metric, we report both the average value and standard deviation of three times running.}
    \vspace{1.5mm}
    \begin{adjustbox}{width=0.93\columnwidth,center}
    \begin{tabular}{c|c|l|ccccc|cc}
    \hline
      \multirow{2}{*}{\textbf{Dataset}} & \multirow{1}{*}{\textbf{Privacy}} & \multirow{2}{*}{\textbf{Method}} & \multicolumn{5}{c|}{\textbf{Statistical Fidelity (TVD) ($\downarrow$)}} & \multicolumn{2}{c}{\textbf{Xgboost} ($\uparrow$)}  \\ 
      & \multirow{1}{*}{\textbf{Budget}} &  & \textbf{1-way} & \textbf{2-way} & \textbf{3-way} & \textbf{4-way} & \textbf{5-way} & \textbf{ACC} & \textbf{AUC} \\ \hline
  \multirow{15}{*}{Bank}   & - & \textit{Training Set} & \textit{0.009{\footnotesize±0.001}} & \textit{0.021{\footnotesize±0.001}} & \textit{0.043{\footnotesize±0.001}} & \textit{0.076{\footnotesize±0.002}} & \textit{0.121{\footnotesize±0.002}}
 & \textit{0.853{\footnotesize±0.009}} & \textit{0.924{\footnotesize±0.005}} \\ \cline{2-10}
  & \multirow{7}{*}{$\epsilon$ = 0.5} & DP-GAN & 0.196{\footnotesize±0.019} & 0.302{\footnotesize±0.023} & 0.368{\footnotesize±0.023} & 0.411{\footnotesize±0.021} & 0.440{\footnotesize±0.017} & 0.528{\footnotesize±0.010} & 0.547{\footnotesize±0.032} \\
  & & DP-CTGAN & 0.175{\footnotesize±0.017} & 0.266{\footnotesize±0.019} & 0.327{\footnotesize±0.019} & 0.374{\footnotesize±0.018} & 0.411{\footnotesize±0.016} & 0.497{\footnotesize±0.024} & 0.509{\footnotesize±0.042} \\
  & & PATE-CTGAN & 0.200{\footnotesize±0.003} & 0.291{\footnotesize±0.004} & 0.351{\footnotesize±0.004} & 0.400{\footnotesize±0.003} & 0.442{\footnotesize±0.002} & 0.420{\footnotesize±0.061} & 0.331{\footnotesize±0.027}  \\
  & & RAP & 0.165{\footnotesize±0.006} & 0.268{\footnotesize±0.005} & 0.345{\footnotesize±0.003} & 0.404{\footnotesize±0.002} & 0.446{\footnotesize±0.001} & \underline{0.636{\footnotesize±0.006}} & 
  \textbf{0.699{\footnotesize±0.005}}  \\
  & & RAP++ & \underline{0.071{\footnotesize±0.005}} & \underline{0.129{\footnotesize±0.005}} & \underline{0.185{\footnotesize±0.005}} & \underline{0.240{\footnotesize±0.005}} & \underline{0.293{\footnotesize±0.006}} & 0.635{\footnotesize±0.068} & \underline{0.697{\footnotesize±0.062}}  \\
  & & GSD & 0.166{\footnotesize±0.001} & 0.245{\footnotesize±0.002} & 0.302{\footnotesize±0.001} & 0.354{\footnotesize±0.002} & 0.401{\footnotesize±0.001} & 0.635{\footnotesize±0.017} & 0.692{\footnotesize±0.017}  \\
  & & \cellcolor{LightCyan}DP-LLMTGen &\cellcolor{LightCyan}\textbf{0.050{\footnotesize±0.003}} & \cellcolor{LightCyan}\textbf{0.091{\footnotesize±0.003}} & \cellcolor{LightCyan}\textbf{0.134{\footnotesize±0.003}} & \cellcolor{LightCyan}\textbf{0.181{\footnotesize±0.004}} & \cellcolor{LightCyan}\textbf{0.232{\footnotesize±0.004}} & \cellcolor{LightCyan}\textbf{0.637{\footnotesize±0.007}} & \cellcolor{LightCyan}0.669{\footnotesize±0.032} \\
 \cline{2-10}
  & \multirow{7}{*}{$\epsilon$ = 1.0} & DP-GAN & 0.244{\footnotesize±0.013} & 0.357{\footnotesize±0.013} & 0.418{\footnotesize±0.013} & 0.453{\footnotesize±0.011} & 0.473{\footnotesize±0.009} & 0.460{\footnotesize±0.071} & 0.396{\footnotesize±0.056}\\
  & & DP-CTGAN & 0.157{\footnotesize±0.019} & 0.244{\footnotesize±0.020} & 0.303{\footnotesize±0.022} & 0.347{\footnotesize±0.023} & 0.382{\footnotesize±0.023} & 0.499{\footnotesize±0.025} & 0.518{\footnotesize±0.214}  \\
  & & PATE-CTGAN & 0.194{\footnotesize±0.007} & 0.287{\footnotesize±0.008} & 0.350{\footnotesize±0.008} & 0.402{\footnotesize±0.007} & 0.445{\footnotesize±0.006} & 0.533{\footnotesize±0.027} & 0.583{\footnotesize±0.111}  \\
  & & RAP & 0.159{\footnotesize±0.005} & 0.263{\footnotesize±0.003} & 0.341{\footnotesize±0.002} & 0.400{\footnotesize±0.001} & 0.443{\footnotesize±0.001} & 0.655{\footnotesize±0.008} & 0.705{\footnotesize±0.006}  \\
  & & RAP++ & \underline{0.058{\footnotesize±0.007}} & \underline{0.105{\footnotesize±0.009}} & \underline{0.151{\footnotesize±0.010}} & \underline{0.201{\footnotesize±0.011}} & \underline{0.252{\footnotesize±0.011}} & \textbf{0.674{\footnotesize±0.017}} & \textbf{0.747{\footnotesize±0.019}} \\
  & & GSD & 0.166{\footnotesize±0.001} & 0.244{\footnotesize±0.002} & 0.301{\footnotesize±0.001} & 0.353{\footnotesize±0.002} & 0.400{\footnotesize±0.001} & \underline{0.671{\footnotesize±0.041}} & \underline{0.726{\footnotesize±0.032}}  \\
  & & \cellcolor{LightCyan}DP-LLMTGen & \cellcolor{LightCyan}\textbf{0.047{\footnotesize±0.010}} & \cellcolor{LightCyan}\textbf{0.085{\footnotesize±0.012}} & \cellcolor{LightCyan}\textbf{0.125{\footnotesize±0.011}} & \cellcolor{LightCyan}\textbf{0.169{\footnotesize±0.010}} & \cellcolor{LightCyan}\textbf{0.218{\footnotesize±0.009}} & \cellcolor{LightCyan}0.638{\footnotesize±0.004} & \cellcolor{LightCyan}0.703{\footnotesize±0.008} \\ \hline

  \multirow{15}{*}{Adult}   & - & \textit{Training Set} & \textit{0.004{\footnotesize±0.001}} & \textit{0.011{\footnotesize±0.001}} & \textit{0.024{\footnotesize±0.001}} & \textit{0.044{\footnotesize±0.001}} & \textit{0.073{\footnotesize±0.001}} & \textit{0.872{\footnotesize±0.002}} & \textit{0.928{\footnotesize±0.001}} \\ \cline{2-10}
  & \multirow{7}{*}{$\epsilon$ = 0.5} & DP-GAN & 0.201{\footnotesize±0.013} & 0.316{\footnotesize±0.015} & 0.386{\footnotesize±0.014} & 0.429{\footnotesize±0.011} & 0.456{\footnotesize±0.008} & 0.765{\footnotesize±0.010} & 0.692{\footnotesize±0.166}  \\
  & & DP-CTGAN & 0.151{\footnotesize±0.012} & 0.239{\footnotesize±0.012} & 0.299{\footnotesize±0.010} & 0.343{\footnotesize±0.007} & 0.378{\footnotesize±0.005} & 0.759{\footnotesize±0.001} & 0.606{\footnotesize±0.089}  \\
  & & PATE-CTGAN & 0.262{\footnotesize±0.005} & 0.365{\footnotesize±0.004} & 0.425{\footnotesize±0.004} & 0.464{\footnotesize±0.003} & 0.488{\footnotesize±0.002} & 0.518{\footnotesize±0.248} & 0.379{\footnotesize±0.146} \\
  & & RAP & 0.152{\footnotesize±0.001} & 0.253{\footnotesize±0.002} & 0.326{\footnotesize±0.002} & 0.382{\footnotesize±0.002} & 0.425{\footnotesize±0.002} & \underline{0.791{\footnotesize±0.003}} & \underline{0.819{\footnotesize±0.004}} \\
  & & RAP++ & \underline{0.066{\footnotesize±0.003}} & \underline{0.116{\footnotesize±0.004}} & \underline{0.162{\footnotesize±0.004}} & \underline{0.208{\footnotesize±0.006}} & \underline{0.251{\footnotesize±0.008}} & 0.766{\footnotesize±0.012} & 0.758{\footnotesize±0.023}  \\
  & & GSD & 0.245{\footnotesize±0.001} & 0.337{\footnotesize±0.000} & 0.395{\footnotesize±0.000} & 0.438{\footnotesize±0.000} & 0.470{\footnotesize±0.000} & 0.759{\footnotesize±0.001} & 0.768{\footnotesize±0.019}  \\
  & & \cellcolor{LightCyan}DP-LLMTGen & \cellcolor{LightCyan}\textbf{0.058{\footnotesize±0.008}} & \cellcolor{LightCyan}\textbf{0.095{\footnotesize±0.010}} & \cellcolor{LightCyan}\textbf{0.126{\footnotesize±0.011}} & \cellcolor{LightCyan}\textbf{0.154{\footnotesize±0.011}} & \cellcolor{LightCyan}\textbf{0.183{\footnotesize±0.010}} & \cellcolor{LightCyan}\textbf{0.833{\footnotesize±0.005}} & \cellcolor{LightCyan}\textbf{0.887{\footnotesize±0.003}} \\
 \cline{2-10}
  & \multirow{7}{*}{$\epsilon$ = 1.0} & DP-GAN &0.259{\footnotesize±0.112} & 0.370{\footnotesize±0.091} & 0.427{\footnotesize±0.058} & 0.460{\footnotesize±0.034} & 0.479{\footnotesize±0.018} & 0.761{\footnotesize±0.002} & 0.589{\footnotesize±0.126} \\
  & & DP-CTGAN & 0.178{\footnotesize±0.026} & 0.277{\footnotesize±0.027} & 0.339{\footnotesize±0.024} & 0.380{\footnotesize±0.021} & 0.409{\footnotesize±0.018} & 0.550{\footnotesize±0.272} & 0.597{\footnotesize±0.110}  \\
  & & PATE-CTGAN & 0.256{\footnotesize±0.002} & 0.362{\footnotesize±0.001} & 0.423{\footnotesize±0.001} & 0.463{\footnotesize±0.001} & 0.487{\footnotesize±0.000} & 0.589{\footnotesize±0.295} & 0.486{\footnotesize±0.071}  \\
  & & RAP & 0.148{\footnotesize±0.000} & 0.248{\footnotesize±0.001} & 0.322{\footnotesize±0.001} & 0.378{\footnotesize±0.001} & 0.421{\footnotesize±0.001} & 0.801{\footnotesize±0.006} & \underline{0.831{\footnotesize±0.005}} \\
  & & RAP++ & \underline{0.054{\footnotesize±0.002}} & \underline{0.099{\footnotesize±0.003}} & \underline{0.142{\footnotesize±0.003}} & \underline{0.183{\footnotesize±0.002}} & \underline{0.225{\footnotesize±0.002}} & \underline{0.803{\footnotesize±0.003}} & 0.807{\footnotesize±0.017} \\
  & & GSD & 0.245{\footnotesize±0.001} & 0.337{\footnotesize±0.000} & 0.395{\footnotesize±0.000} & 0.438{\footnotesize±0.000} & 0.470{\footnotesize±0.000} & 0.759{\footnotesize±0.001} & 0.778{\footnotesize±0.006}  \\
  & & \cellcolor{LightCyan}DP-LLMTGen & \cellcolor{LightCyan}\textbf{0.038{\footnotesize±0.002}} & \cellcolor{LightCyan}\textbf{0.068{\footnotesize±0.002}} & \cellcolor{LightCyan}\textbf{0.096{\footnotesize±0.003}} & \cellcolor{LightCyan}\textbf{0.124{\footnotesize±0.003}} & \cellcolor{LightCyan}\textbf{0.153{\footnotesize±0.002}} & \cellcolor{LightCyan}\textbf{0.831{\footnotesize±0.003}} & \cellcolor{LightCyan}\textbf{0.879{\footnotesize±0.003}} \\ \hline
  
  \multirow{15}{*}{Food} & - & \textit{Training Set} & \textit{0.062{\footnotesize±0.002}} & \textit{0.134{\footnotesize±0.004}} & \textit{0.221{\footnotesize±0.010}} & \textit{0.299{\footnotesize±0.013}} & \textit{0.354{\footnotesize±0.014}} & \textit{0.902{\footnotesize±0.027}} & \textit{0.880{\footnotesize±0.143}} \\  \cline{2-10}
  & \multirow{7}{*}{$\epsilon$ = 0.5} & DP-GAN & 0.229{\footnotesize±0.026} & 0.345{\footnotesize±0.019} & 0.425{\footnotesize±0.012} & 0.472{\footnotesize±0.006} & 0.492{\footnotesize±0.003} & {0.658{\footnotesize±0.381}} & {0.401{\footnotesize±0.184}} \\
  & & DP-CTGAN & 0.229{\footnotesize±0.009} & 0.344{\footnotesize±0.004} & 0.427{\footnotesize±0.004} & 0.475{\footnotesize±0.003} & 0.494{\footnotesize±0.001} & {0.637{\footnotesize±0.186}} & {0.358{\footnotesize±0.145}} \\
  & & PATE-CTGAN & 0.175{\footnotesize±0.010} & 0.295{\footnotesize±0.006} & 0.393{\footnotesize±0.004} & 0.456{\footnotesize±0.002} & 0.486{\footnotesize±0.002} & \underline{0.868{\footnotesize±0.020}} & {0.346{\footnotesize±0.060}} \\
  & & RAP & \underline{0.152{\footnotesize±0.015}} & \underline{0.266{\footnotesize±0.017}} & \underline{0.377{\footnotesize±0.013}} & \underline{0.451{\footnotesize±0.008}} & \underline{0.485{\footnotesize±0.004}} & {0.739{\footnotesize±0.241}} & {0.489{\footnotesize±0.217}} \\
  & & RAP++ & 0.250{\footnotesize±0.007} & 0.369{\footnotesize±0.002} & 0.437{\footnotesize±0.003} & 0.476{\footnotesize±0.002} & 0.493{\footnotesize±0.001} & {0.650{\footnotesize±0.212}} & {0.609{\footnotesize±0.031}} \\
  & & GSD & 0.169{\footnotesize±0.012} & 0.287{\footnotesize±0.006} & 0.389{\footnotesize±0.002} & 0.455{\footnotesize±0.000} & \underline{0.485{\footnotesize±0.001}} & {0.641{\footnotesize±0.077}} & \textbf{0.693{\footnotesize±0.023}} \\
  & & \cellcolor{LightCyan}DP-LLMTGen & \cellcolor{LightCyan}\textbf{0.149{\footnotesize±0.012}} & \cellcolor{LightCyan}\textbf{0.264{\footnotesize±0.009}} & \cellcolor{LightCyan}\textbf{0.372{\footnotesize±0.004}} & \cellcolor{LightCyan}\textbf{0.446{\footnotesize±0.002}} & \cellcolor{LightCyan}\textbf{0.482{\footnotesize±0.001}} & \cellcolor{LightCyan}\textbf{0.872{\footnotesize±0.013}} & \cellcolor{LightCyan}\underline{0.665{\footnotesize±0.130}} \\
  \cline{2-10}
    & \multirow{7}{*}{$\epsilon$ = 1.0} & DP-GAN & 0.249{\footnotesize±0.023} & 0.364{\footnotesize±0.019} & 0.437{\footnotesize±0.012} & 0.477{\footnotesize±0.006} & 0.494{\footnotesize±0.003} & 0.372{\footnotesize±0.219} & \underline{0.650{\footnotesize±0.055}} \\
    & & DP-CTGAN & 0.242{\footnotesize±0.010} & 0.357{\footnotesize±0.010} & 0.436{\footnotesize±0.006} & 0.479{\footnotesize±0.002} & 0.495{\footnotesize±0.001} & 0.470{\footnotesize±0.368} & 0.355{\footnotesize±0.076} \\
    & & PATE-CTGAN & 0.135{\footnotesize±0.010} & 0.258{\footnotesize±0.008} & 0.362{\footnotesize±0.004} & 0.433{\footnotesize±0.002} & \underline{0.473{\footnotesize±0.001}} & \textbf{0.872{\footnotesize±0.013}} & 0.411{\footnotesize±0.004} \\
    & & RAP & \underline{0.132{\footnotesize±0.017}} & \underline{0.243{\footnotesize±0.020}} & \underline{0.354{\footnotesize±0.019}} & \underline{0.433{\footnotesize±0.016}} & \underline{0.473{\footnotesize±0.011}} & 0.812{\footnotesize±0.052} & 0.623{\footnotesize±0.088} \\
    & & RAP++ & 0.228{\footnotesize±0.024} & 0.346{\footnotesize±0.020} & 0.423{\footnotesize±0.014} & 0.468{\footnotesize±0.007} & 0.489{\footnotesize±0.003} & 0.850{\footnotesize±0.030} & 0.601{\footnotesize±0.045} \\
    & & GSD & 0.158{\footnotesize±0.010} & 0.272{\footnotesize±0.007} & 0.376{\footnotesize±0.004} & 0.446{\footnotesize±0.002} & 0.481{\footnotesize±0.001} & 0.632{\footnotesize±0.075} & \textbf{0.692{\footnotesize±0.193}} \\ 
    & & \cellcolor{LightCyan}DP-LLMTGen & \cellcolor{LightCyan}\textbf{0.113{\footnotesize±0.006}} & \cellcolor{LightCyan}\textbf{0.220{\footnotesize±0.008}} & \cellcolor{LightCyan}\textbf{0.333{\footnotesize±0.009}} & \cellcolor{LightCyan}\textbf{0.419{\footnotesize±0.007}} & \cellcolor{LightCyan}\textbf{0.468{\footnotesize±0.004}} & \cellcolor{LightCyan}\textbf{0.872{\footnotesize±0.013}} & \cellcolor{LightCyan}{0.482{\footnotesize±0.173}} \\
    \hline

  \multirow{15}{*}{Apple}   & - & \textit{Training Set} & \textit{0.025{\footnotesize±0.001}} & \textit{0.093{\footnotesize±0.000}} & \textit{0.242{\footnotesize±0.004}} & \textit{0.407{\footnotesize±0.005}} & \textit{0.482{\footnotesize±0.002}} & \textit{0.896{\footnotesize±0.016}} & \textit{0.959{\footnotesize±0.005}} \\ \cline{2-10}
  & \multirow{7}{*}{$\epsilon$ = 0.5} & DP-GAN & 0.270{\footnotesize±0.003} & 0.371{\footnotesize±0.003} & 0.433{\footnotesize±0.002} & 0.476{\footnotesize±0.000} & 0.494{\footnotesize±0.001} & 0.498{\footnotesize±0.006} & 0.579{\footnotesize±0.053} \\
  & & DP-CTGAN & 0.183{\footnotesize±0.020} & 0.272{\footnotesize±0.026} & 0.350{\footnotesize±0.023} & \underline{0.431{\footnotesize±0.013}} & 
\underline{0.483{\footnotesize±0.004}} & 0.498{\footnotesize±0.007} & 0.490{\footnotesize±0.031}  \\
  & & PATE-CTGAN & \underline{0.119{\footnotesize±0.003}} & \underline{0.199{\footnotesize±0.005}} & \textbf{0.288{\footnotesize±0.005}} & \textbf{0.400{\footnotesize±0.003}} & \textbf{0.478{\footnotesize±0.001}} & 0.505{\footnotesize±0.036} & 0.505{\footnotesize±0.065} \\
  & & RAP & 0.202{\footnotesize±0.002} & 0.315{\footnotesize±0.000} & 0.434{\footnotesize±0.001} & 0.491{\footnotesize±0.000} & 0.499{\footnotesize±0.000} & 0.467{\footnotesize±0.033} & 0.529{\footnotesize±0.069}  \\
  & & RAP++ & 0.120{\footnotesize±0.030} & 0.270{\footnotesize±0.028} & 0.411{\footnotesize±0.017} & 0.479{\footnotesize±0.006} & 0.497{\footnotesize±0.001} & \textbf{0.616{\footnotesize±0.047}} & \underline{0.665{\footnotesize±0.047}} \\
  & & GSD & 0.124{\footnotesize±0.005} & 0.243{\footnotesize±0.002} & 0.377{\footnotesize±0.001} & 0.468{\footnotesize±0.000} & 0.496{\footnotesize±0.000} & 0.526{\footnotesize±0.011} & 0.547{\footnotesize±0.010}  \\
  & & \cellcolor{LightCyan}DP-LLMTGen & \cellcolor{LightCyan}\textbf{0.076{\footnotesize±0.009}} & \cellcolor{LightCyan}\textbf{0.156{\footnotesize±0.008}} & \cellcolor{LightCyan}\underline{0.311{\footnotesize±0.006}} & \cellcolor{LightCyan}{0.448{\footnotesize±0.002}} & \cellcolor{LightCyan}{0.493{\footnotesize±0.000}} & \cellcolor{LightCyan}\underline{0.588{\footnotesize±0.041}} & \cellcolor{LightCyan}\textbf{0.684{\footnotesize±0.033}} \\ \cline{2-10}
    & \multirow{7}{*}{$\epsilon$ = 1.0} & DP-GAN & 0.295{\footnotesize±0.007} & 0.395{\footnotesize±0.005} & 0.451{\footnotesize±0.004} & 0.482{\footnotesize±0.002} & 0.496{\footnotesize±0.001} & 0.415{\footnotesize±0.023} & 0.372{\footnotesize±0.019} \\
    & & DP-CTGAN & 0.282{\footnotesize±0.027} & 0.373{\footnotesize±0.025} & 0.431{\footnotesize±0.018} & 0.473{\footnotesize±0.009} & 0.493{\footnotesize±0.003} & 0.491{\footnotesize±0.021} & 0.585{\footnotesize±0.057} \\
    & & PATE-CTGAN & {0.116{\footnotesize±0.006}} & \underline{0.193{\footnotesize±0.007}} & \underline{0.283{\footnotesize±0.006}} & \textbf{0.398{\footnotesize±0.008}} & \textbf{0.477{\footnotesize±0.005}} & 0.542{\footnotesize±0.009} & 0.597{\footnotesize±0.027} \\
    & & RAP & 0.199{\footnotesize±0.002} & 0.312{\footnotesize±0.001} & 0.432{\footnotesize±0.001} & 0.491{\footnotesize±0.000} & 0.499{\footnotesize±0.000} & 0.487{\footnotesize±0.048} & 0.506{\footnotesize±0.040} \\
    & & RAP++ & \underline{0.090{\footnotesize±0.006}} & 0.233{\footnotesize±0.005} & 0.386{\footnotesize±0.008} & 0.471{\footnotesize±0.004} & 0.495{\footnotesize±0.001} & \textbf{0.654{\footnotesize±0.011}} & \textbf{0.719{\footnotesize±0.010}} \\
    & & GSD & 0.093{\footnotesize±0.004} & 0.204{\footnotesize±0.002} & 0.345{\footnotesize±0.002} & 0.454{\footnotesize±0.002} & 0.493{\footnotesize±0.000} & 0.563{\footnotesize±0.016} & 0.582{\footnotesize±0.016} \\
    & & \cellcolor{LightCyan}DP-LLMTGen & \cellcolor{LightCyan}\textbf{0.068{\footnotesize±0.003}} & \cellcolor{LightCyan}\textbf{0.139{\footnotesize±0.003}} & \cellcolor{LightCyan}\textbf{0.277{\footnotesize±0.003}} & \cellcolor{LightCyan}\underline{0.425{\footnotesize±0.002}} & \cellcolor{LightCyan}\underline{0.488{\footnotesize±0.001}} & \cellcolor{LightCyan}\underline{0.639{\footnotesize±0.011}} & \cellcolor{LightCyan}\underline{0.687{\footnotesize±0.033}} \\
\hline
  \multirow{15}{*}{Shipping}   & - & \textit{Training Set} & \textit{0.010{\footnotesize±0.001}} & \textit{0.026{\footnotesize±0.001}} & \textit{0.060{\footnotesize±0.001}} & \textit{0.123{\footnotesize±0.003}} & \textit{0.218{\footnotesize±0.005}} & \textit{0.693{\footnotesize±0.005}} & \textit{0.756{\footnotesize±0.004}} \\ \cline{2-10} 
  & \multirow{7}{*}{$\epsilon = 0.5$} & DPGAN & 0.228{\footnotesize±0.023} & 0.334{\footnotesize±0.022} & 0.395{\footnotesize±0.018} & 0.435{\footnotesize±0.016} & 0.463{\footnotesize±0.012} & 0.513{\footnotesize±0.093} & 0.493{\footnotesize±0.081} \\
  & & DPCTGAN & 0.269{\footnotesize±0.015} & 0.368{\footnotesize±0.021} & 0.421{\footnotesize±0.019} & 0.452{\footnotesize±0.014} & 0.473{\footnotesize±0.009} & 0.537{\footnotesize±0.107} & 0.485{\footnotesize±0.039} \\
  & & PATECTGAN & 0.105{\footnotesize±0.001} & 0.173{\footnotesize±0.003} & 0.230{\footnotesize±0.003} & 0.296{\footnotesize±0.004} & 0.376{\footnotesize±0.004} & 0.593{\footnotesize±0.009} & 0.572{\footnotesize±0.030} \\
  & & RAP & 0.109{\footnotesize±0.006} & 0.177{\footnotesize±0.005} & 0.236{\footnotesize±0.004} & 0.303{\footnotesize±0.002} & 0.378{\footnotesize±0.002} & 0.577{\footnotesize±0.006} & 0.518{\footnotesize±0.012} \\
  & & RAP++ & 0.080{\footnotesize±0.002} & 0.145{\footnotesize±0.004} & 0.214{\footnotesize±0.006} & 0.290{\footnotesize±0.007} & 0.364{\footnotesize±0.007} & \underline{0.594{\footnotesize±0.011}} & \underline{0.624{\footnotesize±0.031}} \\
  & & GSD & \underline{0.057{\footnotesize±0.002}} & \underline{0.100{\footnotesize±0.002}} & \underline{0.149{\footnotesize±0.001}} & \underline{0.218{\footnotesize±0.002}} & \underline{0.306{\footnotesize±0.003}} & \textbf{0.606{\footnotesize±0.014}} & \textbf{0.663{\footnotesize±0.022}} \\
  & & \cellcolor{LightCyan}DP-LLMTGen & \cellcolor{LightCyan}\textbf{0.048{\footnotesize±0.002}} & \cellcolor{LightCyan}\textbf{0.084{\footnotesize±0.002}} & \cellcolor{LightCyan}\textbf{0.126{\footnotesize±0.003}} & \cellcolor{LightCyan}\textbf{0.186{\footnotesize±0.003}} & \cellcolor{LightCyan}\textbf{0.271{\footnotesize±0.005}} & \cellcolor{LightCyan}{0.590{\footnotesize±0.006}} & \cellcolor{LightCyan}{0.510{\footnotesize±0.031}} \\ \cline{2-10}
  & \multirow{7}{*}{$\epsilon = 1.0$} & DPGAN & 0.219{\footnotesize±0.067} & 0.327{\footnotesize±0.061} & 0.395{\footnotesize±0.044} & 0.439{\footnotesize±0.029} & 0.467{\footnotesize±0.017} & 0.510{\footnotesize±0.102} & 0.498{\footnotesize±0.133} \\
  & & DPCTGAN & 0.274{\footnotesize±0.025} & 0.372{\footnotesize±0.022} & 0.422{\footnotesize±0.018} & 0.453{\footnotesize±0.013} & 0.474{\footnotesize±0.009} & 0.468{\footnotesize±0.109} & 0.489{\footnotesize±0.036} \\
  & & PATECTGAN & 0.070{\footnotesize±0.002} & 0.124{\footnotesize±0.002} & 0.176{\footnotesize±0.002} & 0.243{\footnotesize±0.004} & 0.328{\footnotesize±0.005} & 0.592{\footnotesize±0.013} & 0.480{\footnotesize±0.152} \\
  & & RAP & 0.109{\footnotesize±0.005} & 0.178{\footnotesize±0.004} & 0.237{\footnotesize±0.003} & 0.303{\footnotesize±0.001} & 0.377{\footnotesize±0.001} & 0.582{\footnotesize±0.007} & 0.541{\footnotesize±0.060} \\
  & & RAP++ & 0.070{\footnotesize±0.007} & 0.123{\footnotesize±0.011} & 0.182{\footnotesize±0.013} & 0.253{\footnotesize±0.013} & 0.332{\footnotesize±0.010} & \textbf{0.623{\footnotesize±0.006}} & \underline{0.673{\footnotesize±0.014}} \\
  & & GSD & \underline{0.053{\footnotesize±0.002}} & \underline{0.093{\footnotesize±0.001}} & \underline{0.140{\footnotesize±0.000}} & \underline{0.208{\footnotesize±0.002}} & \underline{0.297{\footnotesize±0.003}} &  \underline{0.616{\footnotesize±0.009}} & \textbf{0.689{\footnotesize±0.011}} \\
  & & \cellcolor{LightCyan}DP-LLMTGen & \cellcolor{LightCyan}\textbf{0.042{\footnotesize±0.002}} & \cellcolor{LightCyan}\textbf{0.074{\footnotesize±0.002}} & \cellcolor{LightCyan}\textbf{0.114{\footnotesize±0.000}} & \cellcolor{LightCyan}\textbf{0.174{\footnotesize±0.004}} & \cellcolor{LightCyan}\textbf{0.262{\footnotesize±0.008}} & \cellcolor{LightCyan}{0.594{\footnotesize±0.010}} & \cellcolor{LightCyan}{0.452{\footnotesize±0.061}} \\ \hline
\end{tabular}
\end{adjustbox}
\label{tab:general-performance}
\vspace{-9mm}
\end{table}

\section{Experiments and Results}
\subsection{Overall Evaluation}
\textbf{Datasets.} We conduct experiments on five datasets, which vary in size from less than 400 to more than 40,000 samples. These datasets also exhibit a diverse range of ratios between numerical and categorical features. The details of datasets and their key characteristics are provided in Appendix~\ref{sec:apd_dataset}, Table~\ref{tab:dataset}. Notably, three of the five datasets were published after the knowledge cutoff date of the Llama 2 models \citep{touvron2023llama}. We split all datasets into 80:20 for training and testing. All our and baseline models are trained and tested on identical sets. To provide robust results, for each dataset, we split into three different train-test sets using multiple random seeds.

\textbf{Baselines.} Our baselines include state-of-the-art GAN-based and marginal-based methods. Regarding GANs, we implement DP-GAN as a baseline, which trains Generative Adversarial Networks by DPSGD. We also consider Conditional Tabular GAN (CTGAN) \citep{ctgan} which is a variant designed for tabular data. We train the CTGANs by both DPSGD (named DP-CTGAN) and Private Aggregation of Teacher Ensembles (PATE) framework \citep{Papernot2018ScalablePL} (referred as PATE-CTGAN). {For marginal-based methods, we implement three methods: RAP~\citep{pmlr-v139-aydore21a}, RAP++~\citep{Vietri2022PrivateSD}, and GSD~\citep{gsd}}.

\textbf{Metrics.} We focus on two main kinds of metrics: Statistical Fidelity (SF) and Machine Learning (ML) downstream performance, which are similar to previous works \citep{gsd, Vietri2022PrivateSD, pmlr-v139-aydore21a}. For statistical fidelity, we calculate the average of total variation distances (TVD) of single/joint distributions (ranging from 1-way to 5-way) between the synthetic and testing sets. For ML downstream performance, we train XGBoost models using the synthetic data and report their performance (accuracy and AUC) on the testing sets. To ensure robust results, we run each experiment three times and report the average values and standard deviations. Due to the space limit, we present the details of the evaluation metrics in Appendix~\ref{sec:app-impl-eval}.

\textbf{Results.} Table~\ref{tab:general-performance} presents the overall performance of \texttt{DP-LLMTGen} (using LLama-2-chat-hf 7B backbone \citep{touvron2023llama}) and the baselines under different privacy budget settings across five datasets. Generally, \texttt{DP-LLMTGen} outperforms the baselines in most cases. For the Bank and Adult datasets, which have large numbers of samples (11K and 48K, respectively), \texttt{DP-LLMTGen} delivers better statistical fidelity for capturing both low- and high-dimensional dependencies among features by around 15\% compared to the baselines. RAP++ is the only method that comes close. Furthermore, \texttt{DP-LLMTGen} not only achieves the highest machine learning performance on the Adult dataset but also demonstrates competitive results for the Bank dataset. For the Food dataset, which has the smallest number of samples (i.e., 388), while RAP++ fails to provide a high-quality synthetic dataset, \texttt{DP-LLMTGen} still achieves the best statistical fidelity. For the Apple dataset, which consists entirely of numerical features, \texttt{DP-LLMTGen} excels at capturing low-dimensional dependencies. However, the GAN-based methods outperform the LLMs in modeling high-dimensional joint relationships. Notably, better statistical fidelity usually leads to higher performance for ML downstream model but not always. This is consistent to the previous results \citep{tao2022benchmarking}. Due to the space limit, we provide distance to closest record histogram in Appendix~\ref{sec:dcr} which demonstrates that \texttt{DP-LLMTGen} does not memorize and replicate the training datasets. We also provides some visualizations regarding to 1-way and 2-way TVD in Appendices~\ref{sec:1tvd} and~\ref{sec:2tvd}.

\subsection{Ablation Study}
\vspace{-2mm}
\subsubsection{Two-stage fine tuning}
\vspace{-2mm}
\begin{wrapfigure}{r}{0.65\textwidth}
    \vspace{-2cm}
    \centering
    \begin{subfigure}{.26\textwidth}
  \includegraphics[width=\linewidth]{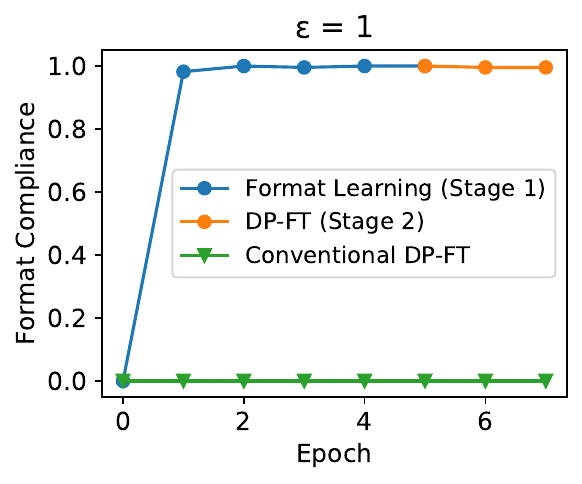}
  \caption{Format-compliance performance of our 2-stage FT and conventional DP-FT under a tight-privacy setting}
  \label{fig:TDPFT-format-1}
\end{subfigure}
\hspace{0.5mm}
\begin{subfigure}{.3\textwidth}
  \centering
  \includegraphics[width=\linewidth]{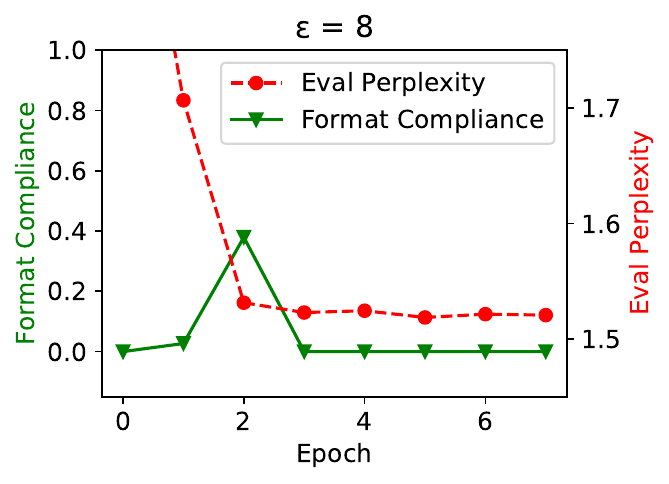}
  \caption{Performance on perplexity and format compliance of conventional DP-FT under a loose privacy setting}
  \label{fig:TDPFT-format-8}
\end{subfigure}
\caption{Analyzing Conventional Differentially-Private Fine Tuning (DP-FT) for Format Compliance on the Adult dataset}
\label{fig:TDPFT-format}
\vspace{-6mm}
\end{wrapfigure}
\textit{Two-stage fine tuning ensures format compliance while conventional DP fine tuning fails}. To verify the advantage of two-stage fine tuning versus conventional DP fine tuning, we conduct an experiment on the Adult dataset. Figure~\ref{fig:TDPFT-format-1} presents the format-compliance capability of conventional DP-FT (in green) and our 2-stage fine tuning (in orange and blue) under the privacy budget of $\left(1, 10^{-6}\right)$-DP. While the proposed 2-stage FT remains nearly perfect format compliance, conventional DP-FT fails entire time. We also investigate the conventional fine-tuning approach when the privacy setting is loose (i.e., $\epsilon = 8$ presented in Figure~\ref{fig:TDPFT-format-8}). The noise added is fairly small, so the model achieves the capability of 0.4 in the second epoch. However, this capability diminishes with continued training, although the evaluation perplexity consistently decreases. It can be because perplexity or cross entropy loss, which is used for standard causal language modeling, are calculated at token levels. However, format compliance is considered at sentence levels. For example, a missing comma in a 200-token sentence does not lead to high perplexity, but is considered as a failure at format compliance.

\vspace{-2mm}
\subsubsection{Weighted Cross Entropy Loss} 
\begin{figure}[h]
\centering
\begin{subfigure}{.237\textwidth}
  \centering
  \includegraphics[width=\linewidth]{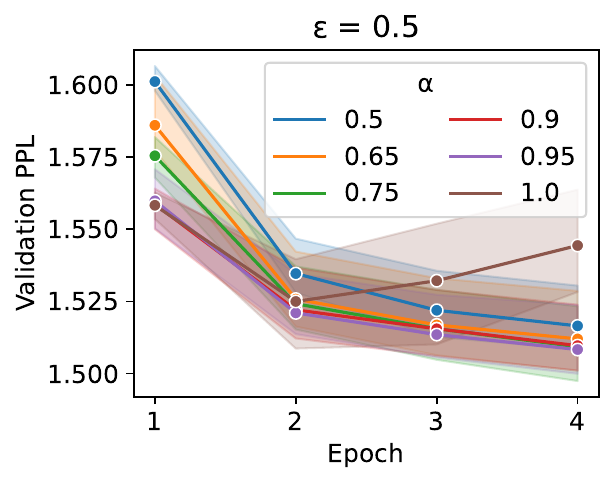}\\
  \includegraphics[width=\linewidth]{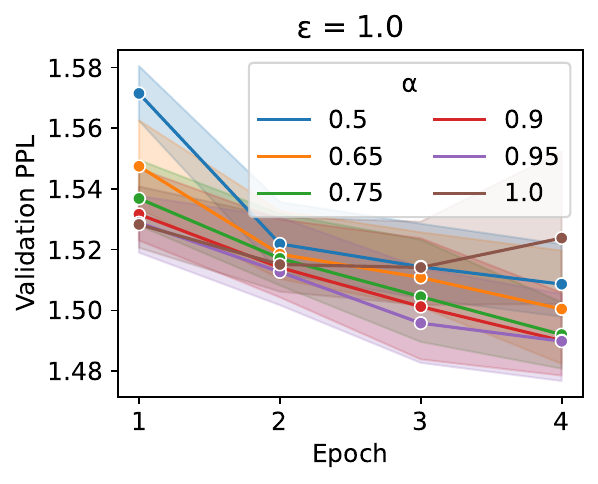}
  \caption{Perplexity}
  \label{fig:wl-ppl}
\end{subfigure}%
\begin{subfigure}{.24\textwidth}
  \centering
  \includegraphics[width=\linewidth]{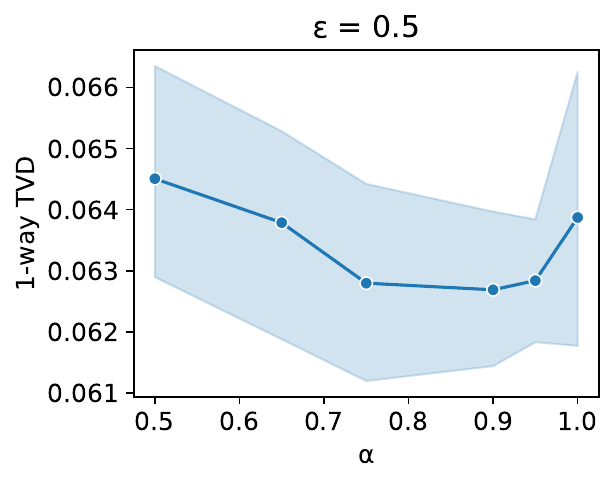}\\
  \includegraphics[width=\linewidth]{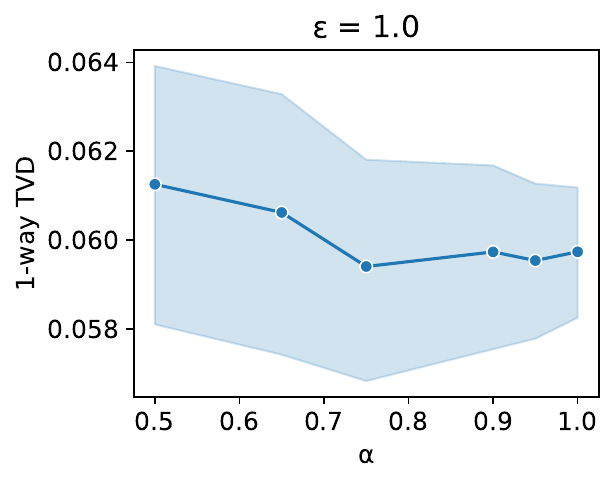}
  \caption{1-way TVD}
  \label{fig:wl-1tvd}
\end{subfigure}
\begin{subfigure}{.24\textwidth}
  \centering
  \includegraphics[width=\linewidth]{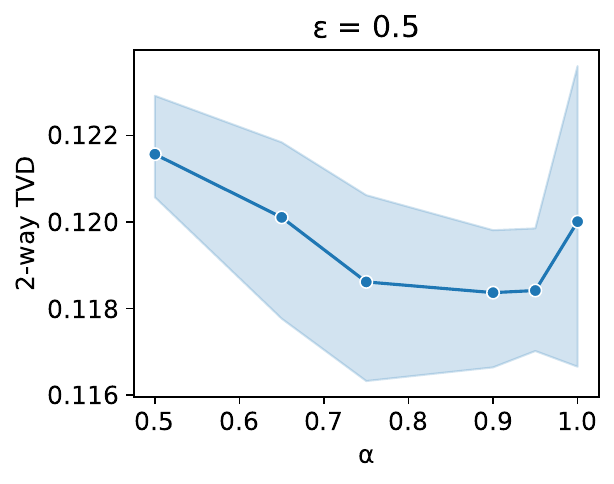}\\
  \includegraphics[width=\linewidth]{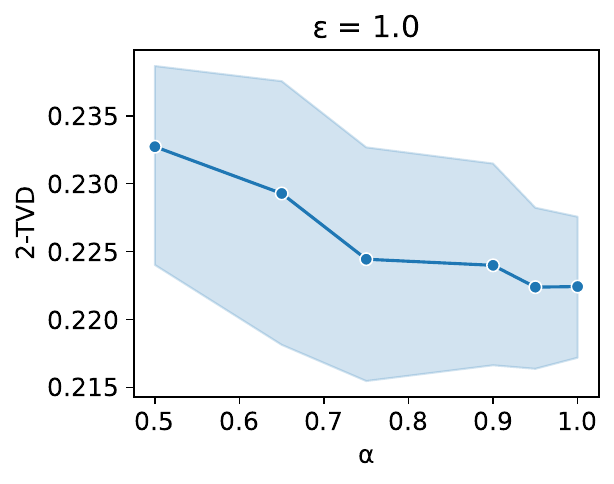}
  \caption{2-way TVD}
  \label{fig:wl-2tvd}
\end{subfigure}
\begin{subfigure}{.228\textwidth}
  \centering
  \includegraphics[width=\linewidth]{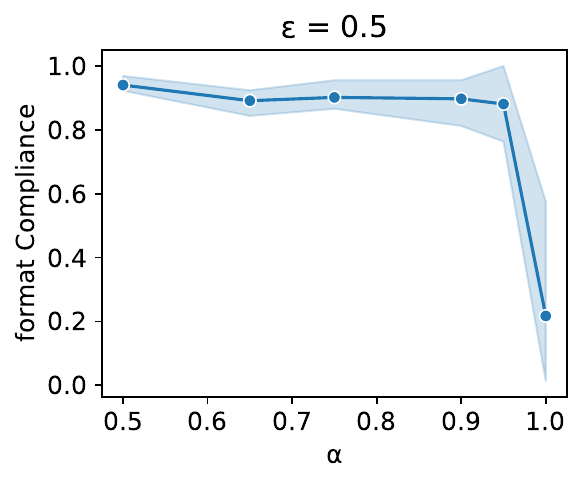}\\
  \includegraphics[width=\linewidth]{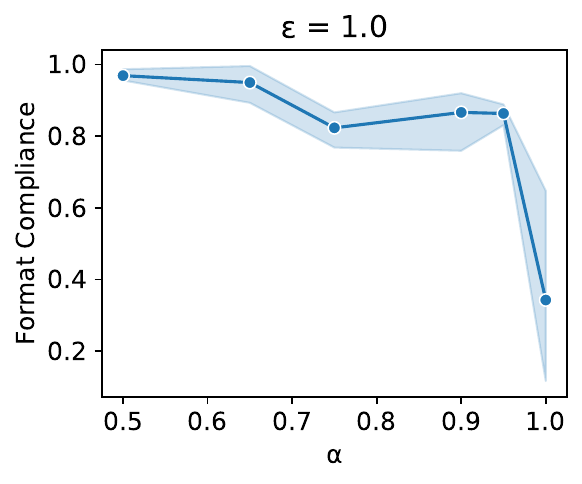}
  \caption{Format Compliance}
  \label{fig:wl-fl}
\end{subfigure}
\vspace{-2mm}
\caption{Performance of \texttt{DP-LLMTGen} while varying $\alpha$ on three subsets of the Adult dataset.}
\label{fig:wl}
\vspace{-3mm}
\end{figure}

\vspace{-2mm}
\textit{Weighted Cross Entropy Loss (WCEL) improves performance}. Notably, if $\alpha = 0.5$, WCEL (in Equation~\ref{eq:wcel}) becomes the standard cross entropy loss which is widely used in causal language modeling. Figure~\ref{fig:wl} depicts performance of \texttt{DP-LLMTGen} with varying $\alpha$. Generally, using any $\alpha$ s.t. $1 > \alpha > 0.5$, i.e., focusing more on the tabular tokens than the format tokens, the model provides better perplexity and statistical fidelity than the model trained by the standard cross entropy loss (i.e., $\alpha = 0.5$), as shown in Figures~\ref{fig:wl-ppl},~\ref{fig:wl-1tvd},~\ref{fig:wl-2tvd}. However, the larger $\alpha$ can lead to some loss of format-compliance capability, as demonstrated in Figure~\ref{fig:wl-fl}. At $\alpha = 1.0$, the second fine-tuning stage ignores entirely format tokens, causing a significant drop in format compliance capability.

\vspace{-2mm}
\subsubsection{Numerical-Understanding Loss}
\vspace{-2mm}
\textit{Numerical-Understanding Loss (NUL) helps the LLMs to deal better with numbers}. To verify the significance of the Numerical-Understanding Loss, we conduct an ablation study using the Apple dataset, where all features are numerical. Table~\ref{tab:NUL} presents the performance of \texttt{DP-LLMTGen} trained with and without NUL under different settings of privacy budget. Generally, \texttt{DP-LLMTGen} trained with NUL provides better statistical fidelity by about 5\%. Moreover, we observe that better Causal Language Modeling does not always lead to a better data generator for numerical features. Table~\ref{tab:NUL} provides the evaluation perplexity and statistical fidelity. There is no direct causal relationships between the two metrics. A higher-perplexity model can provide a better synthetic dataset. This result confirms our statement about the fundamental problem of LLMs using the standard cross entropy loss in representing and understanding numbers in the previous section.


\begin{table}[ht!]
    \vspace{-5mm}
    \centering
    \caption{Performance of \texttt{DP-LLMTGen} with and without the numerical-understanding loss}
    \vspace{0.5mm}
    \begin{adjustbox}{width=\columnwidth,center}
    \begin{tabular}{c|c|c|cccc}
      Privacy Budget & Method & Eval PPL ($\downarrow$) & 1-TVD ($\downarrow$) & 2-TVD ($\downarrow$) & 3-TVD ($\downarrow$) & 4-TVD ($\downarrow$) \\ \hline
     \multirow{2}{*}{$\epsilon = 0.5$} & \texttt{DP-LLMTGen} w.o. NUL & 3.882{\footnotesize±0.007} & 0.081{\footnotesize±0.005} & 0.162{\footnotesize±0.009} & 0.316{\footnotesize±0.012} & 0.452{\footnotesize±0.007} \\ 
      & \texttt{DP-LLMTGen} w. NUL & 3.886{\footnotesize±0.002} & 0.076{\footnotesize±0.008} & 0.156{\footnotesize±0.008} & 0.311{\footnotesize±0.006} & 0.448{\footnotesize±0.002}\\ \hline
     \multirow{2}{*}{$\epsilon = 1.0$} & \texttt{DP-LLMTGen} w.o. NUL & 3.865{\footnotesize±0.002} & 0.064{\footnotesize±0.003} & 0.139{\footnotesize±0.004} & 0.287{\footnotesize±0.007} & 0.434{\footnotesize±0.005} \\ 
      & \texttt{DP-LLMTGen} w. NUL & 3.859{\footnotesize±0.008} & 0.069{\footnotesize±0.003} & 0.139{\footnotesize±0.003} & 0.277{\footnotesize±0.003} & 0.425{\footnotesize±0.003} \\ \hline
    \end{tabular}
    \end{adjustbox}
    \label{tab:NUL}
    \vspace{-8mm}
\end{table}
\subsection{Additional Experimental Analyses}
\label{sec:exp-ana}
\vspace{-3mm}
\textbf{Data contamination testing.} \textit{Data contamination between the pre-training and experimental datasets of the LLama-2 7b model is not a significant problem.}\\
\vspace{-3mm}
\begin{wrapfigure}{r}{0.5\textwidth}
    \vspace{-5mm}
    \centering
    \begin{subfigure}{.24\textwidth}
    \includegraphics[width=\linewidth]{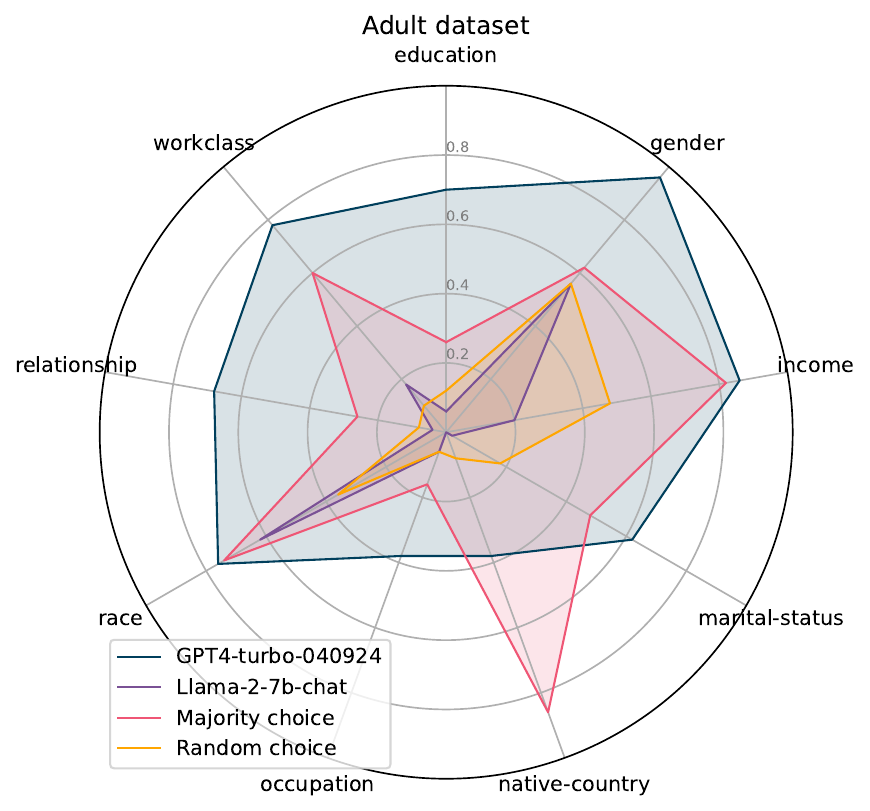}
    \end{subfigure}
    \begin{subfigure}{.24\textwidth}
      \centering
      \includegraphics[width=\linewidth]{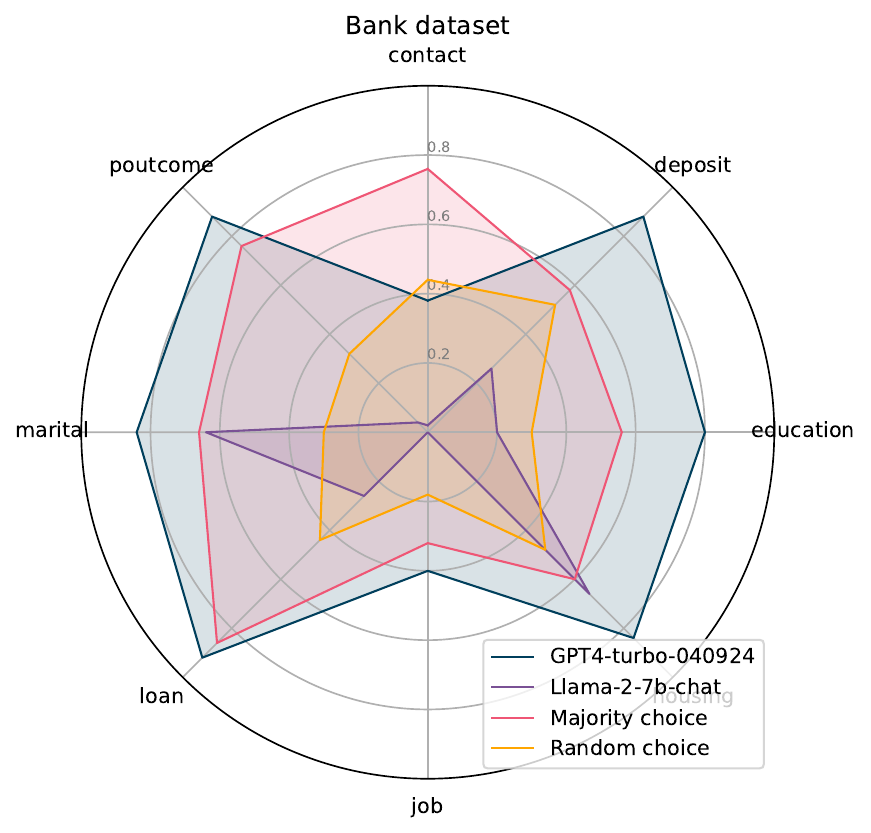}
    \end{subfigure}
    \caption{{Feature completion testing performance of GPT-4 and Llama-2-7b models. The values represent for the percentage of correct completion.}}
    \vspace{-4mm}
    \label{fig:dct}
\end{wrapfigure}
To ensure no DP violence, we conduct the feature completion testing proposed by \citet{bordt2023testing} for Llama-2 7b and GPT-4 models for the Adult and Bank datasets. The results (illustrated in Figure~\ref{fig:dct}) indicate that GPT-4 \citep{openai2024gpt4} could have been trained on those public datasets and actually memorizes the training data to achieve the superior performance compared to the baselines, including random-choice and majority-choice strategies. In contrast, Llama-2 performs comparably to the random-choice strategy but significantly worse than the majority-choice strategy. It indicates that Llama-2 has no memorization and zero knowledge about feature distributions of the datasets.

\textbf{Do LLMs incorporate feature names for context-aware learning?} \textit{LLMs achieve better performance on datasets with meaningful feature names compared to those lacking feature names}.\\
\vspace{-3mm}
\begin{wrapfigure}{l}{0.50\textwidth}
\vspace{-4mm}
\centering
\begin{subfigure}{.24\textwidth}
  \centering
  \includegraphics[width=\linewidth]{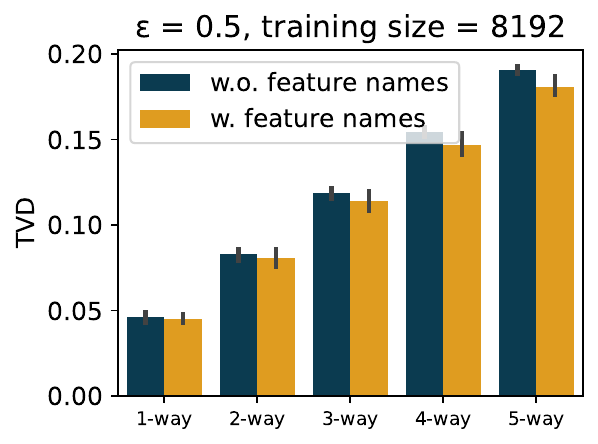}\\
\end{subfigure}%
\begin{subfigure}{.24\textwidth}
  \centering
  \includegraphics[width=\linewidth]{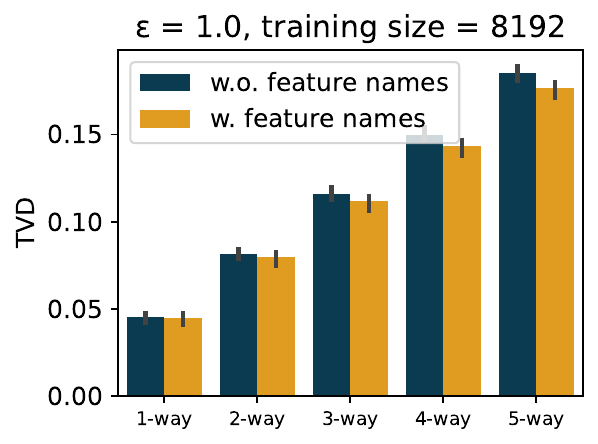}\\
\end{subfigure}%
\caption{Performance of \texttt{DP-LLMTGen} for the Adult subsets with and without feature names.}
\label{fig:noname}
\vspace{-4mm}
\end{wrapfigure}
We conduct an experiment using identical datasets: one dataset is maintained in its original form, while the feature names in the other are changed to \texttt{Feature-1, Feature-2, etc}. Figure~\ref{fig:noname} presents the performance of \texttt{DP-LLMTGen} for these two datasets. Generally, \texttt{DP-LLMTGen} provides better synthetic datasets when fine tuning on the datasets with feature names. This result demonstrates that LLMs utilize pre-trained general knowledge to incorporate both feature names and values. This ability is a unique characteristic of LLMs compared to earlier approaches like GANs. GANs generate outcomes that are independent of feature names, as they can only learn from the feature values. This aligns with the results in \citep{Dinh2022LIFTLF, hegselmann2023tabllm} in classification settings.

\textbf{Does dialogue optimization provide benefit for DP-LLMTabGen?} \textit{Models without dialogue optimization can generate better synthetic datasets}.\\
\vspace{-3mm}
\begin{wrapfigure}{r}{0.50\textwidth}
\centering
\begin{subfigure}{.24\textwidth}
  \centering
  \includegraphics[width=\linewidth]{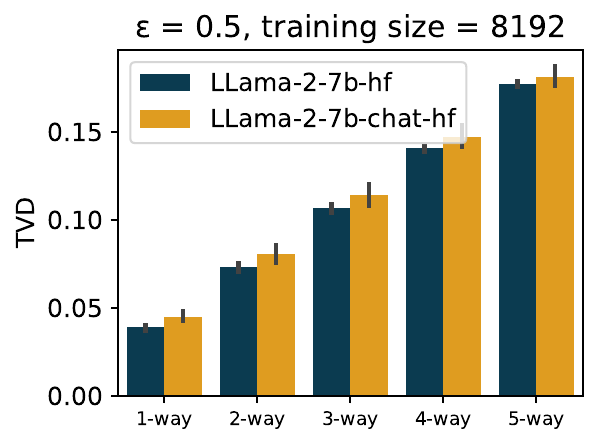}\\
\end{subfigure}%
\begin{subfigure}{.24\textwidth}
  \centering
  \includegraphics[width=\linewidth]{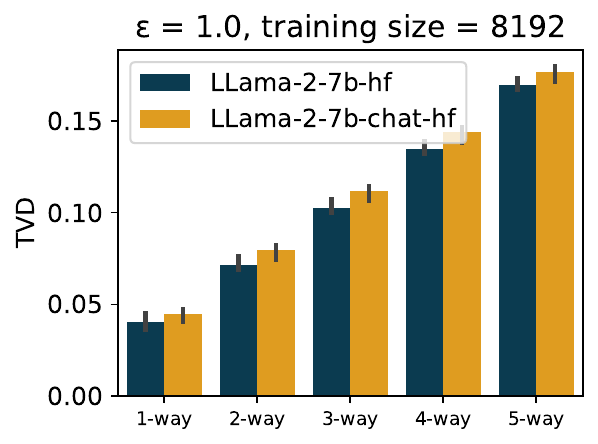}\\
\end{subfigure}%
\caption{Performance of \texttt{DP-LLMTGen} when using Llama-2-7b-hf and Llama-2-7b-chat-hf as the model backbone on subsets of the Adult dataset.}
\label{fig:chat-nonchat}
\vspace{-4mm}
\end{wrapfigure}
Most LLMs were developed for assistant tasks, which typically require dialogue optimization. To understand the effects of dialogue optimization, we conduct an experiments using Llama-2-hf and Llama-2-chat-hf models. Figure~\ref{fig:chat-nonchat} illustrates the performance of \texttt{DP-LLMTGen} using those two models. Generally, models which have not been optimized for conversational tasks exhibit a marginal (though not significant) improvement -- an average decrease of 2.85\% in 5-way TVD. This could be because synthetic data generation tasks might not require dialogue-related features. Therefore, models like Llama-2-hf, which have been optimized for more general tasks, could align better for synthesizing tabular data.


\vspace{-1mm}
\textbf{Are Larger LLMs always better?} \textit{Llama-2 7b can outperform LLama-2 13b under DP settings}.\\
\vspace{-3mm}
\begin{wrapfigure}{l}{0.50\textwidth}
\vspace{-4mm}
\centering
\begin{subfigure}{.24\textwidth}
  \centering
  \includegraphics[width=\linewidth]{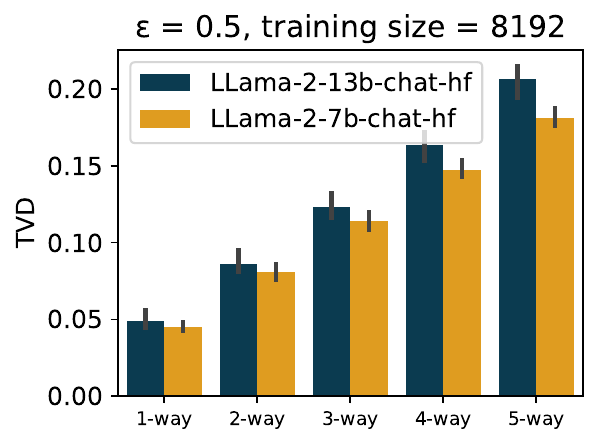}\\
\end{subfigure}%
\begin{subfigure}{.24\textwidth}
  \centering
  \includegraphics[width=\linewidth]{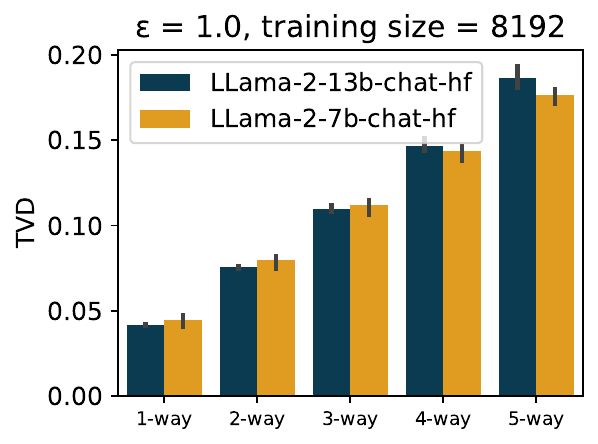}\\
\end{subfigure}%
\caption{Performance of \texttt{DP-LLMTGen} when using Llama-2-7b-chat-hf and Llama-2-13b-chat-hf on three subsets of the Adult dataset.}
\label{fig:7b-13b}
\vspace{-4mm}
\end{wrapfigure}
To understand the effects of model size, we perform an experiment using two Llama-2 models, 7b and 13b, on the Adult subsets. We configure identical hyperparameters of LoRA (i.e., scaling factor and rank). It is noteworthy that given the same rank in LoRA configuration, Llama-2 13b has bigger matrices which leads to more trainable parameters. More specifically, Llama-2 7b and 13b require 8M and 13M trainable parameters, respectively. Figure~\ref{fig:7b-13b} presents the performance of the two LLMs where LLama-2 7b provides better synthetic datasets. One of the reasons is that given fewer trainable parameters, DPSGD employs a lower gradient clipping factor, resulting in smaller total noise. This aligns with existing findings \citep{Kurakin2023HarnessingLM, yu2022differentially} which demonstrated parameter-efficient fine tuning outperforms full fine-tuning under differential privacy settings.

\vspace{-3mm}
\subsection{Controllable Generator -- Fairness-Aware Generation}
\vspace{-2mm}
\begin{wrapfigure}{r}{0.52\textwidth}
    \vspace{-6mm}
    \centering
    \begin{subfigure}{.25\textwidth}
    \includegraphics[width=\linewidth]{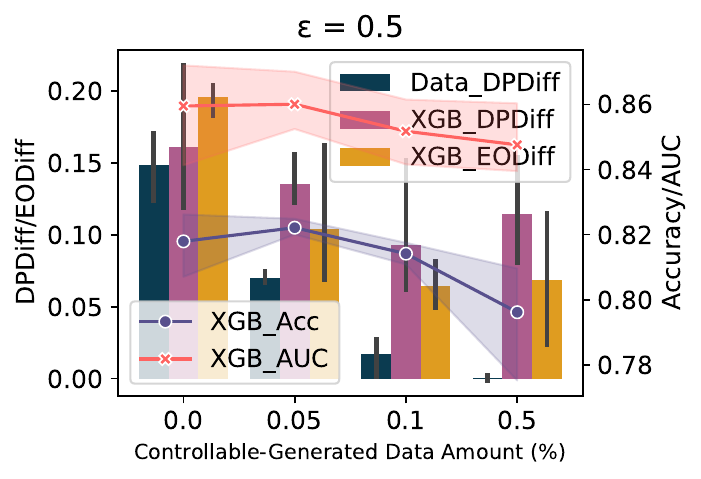}
    \end{subfigure}
    \hfill
    \begin{subfigure}{.25\textwidth}
      \centering
      \includegraphics[width=\linewidth]{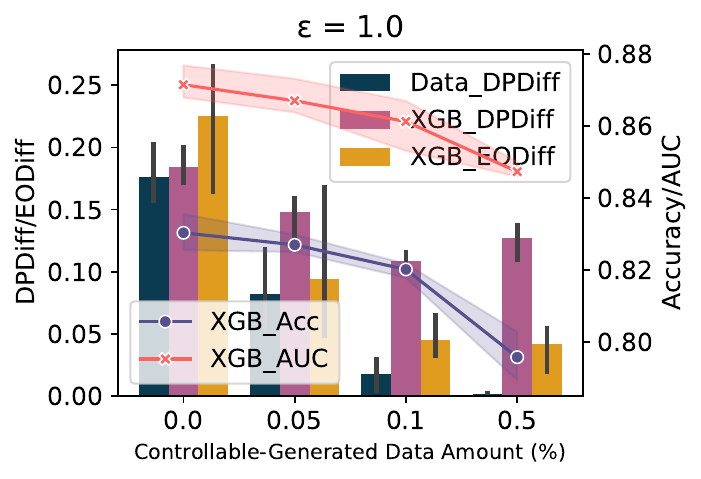}
    \end{subfigure}
    \caption{Fairness-Aware Generation Performance. DPDiff denotes \underline{D}emographic \underline{P}arity \underline{Diff}erence; EODiff stands for \underline{E}qualized of \underline{O}dd \underline{Diff}erence.}
    \vspace{-5mm}
    \label{fig:fairness}
\end{wrapfigure}
\textit{Controllable generator can significantly reduce biases with a minor tradeoff in utility}. For controllable generation, \texttt{DP-LLMTGen} uses value-specified sampling. For example, in the Adult dataset, to generate a sample that is a male earning more than \$50K/year, the initialized prompt can be \texttt{"Gender is male, Income is >50K,"}. We conduct an experiment where the amount of controllable-generated data is varied from 5\% to 50\%. The remainder is produced through random-initialized sampling. A value of 0.0\% denotes non-controllable generation. We also train XGBoost models on the generated datasets. Figure~\ref{fig:fairness} illustrates the fairness in data and ML downstream tasks. Utilizing a controllable generator for a small fraction of the data (e.g., 10\%), the demographic parity difference (detailed in Appendix~\ref{sec:app-fairness-metric}) in data can be decreased by $\sim$85\%. This also leads to a substantial improvement in fairness for ML downstream models while only reducing classification accuracy by $\sim$1\%. When the amount is significant, \texttt{DP-LLMTGen} can generate datasets with perfect demographic parity. However, data demographic parity does not guarantee the complete absence of bias in ML models, as bias elimination in ML still requires in-processing fairness-aware training models \citep{pmlr-v162-li22p, han2024ffb}.

\vspace{-3mm}



\section{Conclusion}
\vspace{-2mm}
We have presented \texttt{DP-LLMTGen} which leverages pretrained LLMs for differentially private tabular data synthesization. Overall, \texttt{DP-LLMTGen} outperformed the baselines across multiple datasets and privacy settings, closing the gap between deep learning-based and marginal-based methods. \texttt{DP-LLMTGen} offers controllable generators which are important for some applications that require fairness guarantees \citep{10.1145/3639283, 10.1145/3648609}. However, one key limitation of \texttt{DP-LLMTGen} is the high computational demand. Despite the limitations, with the rapid development of LLMs, we believe utilizing those models can offer a new paradigm in privacy-preserving tabular data synthesis.


\bibliographystyle{plainnat}
\bibliography{ref}

\clearpage
\appendix
\newpage
{\large \textbf{Appendix}} \par 
Due to the space limit, many details are presented in the Appendix. 

\startcontents[sections]
\printcontents[sections]{ }{1}{}
\newpage

\section{Additional Related Works}
\label{sec:add_related_work}
\textbf{Transformer-based approach for differentially private synthesizing tabular data.} Although our method (\texttt{DP-LLMTGen}) and the methods from \citep{Castellon2023DPTBARTAT, sablayrolles2023privately} (named DP-TBART and SynLM) are all based on transformer networks, there are significant differences. DP-TBART and SynLM train from scratch generative models specialized for each dataset. DP-TBART and SynLM create a small vocabulary including categorical values, and numerical quantiles. In contrast, \texttt{DP-LLMTGen} aims to fine tune pretrained large language models. \texttt{DP-LLMTGen} converts tabular data into textual sentences and uses a (human-language) semantic vocabulary. Additionally, the most important point is that DP-TBART and SynLM underperform the marginal-based methods in most cases \citep{Castellon2023DPTBARTAT, sablayrolles2023privately}. \texttt{DP-LLMTGen} is the first approach that leverages powerful LLMs for differentially private synthesizing tabular data and outperforms the marginal-based methods.

\textbf{Large Language Models for (differentially private) synthetic textual data generation.}  Large language models have been a powerful tool for synthetic data generation. \cite{li-etal-2023-synthetic} investigated in-context learning for text generation to enable text classifiers. \cite{xie2024differentially} utilized LLMs to privately paraphrase texts. Meanwhile, \citet{kurakin2024harnessing, mattern-etal-2022-differentially, yue-etal-2023-synthetic} fine tuning LLMs by DPSGD. \citet{kurakin2024harnessing} pointed out the data contamination problems of previous works and filtered the experimental datasets to address this problem. Non-fine-tuning approaches also work well for textual data because LLMs have been pretrained on massive texts. However, the performance of non-fine-tuning LLMs for tabular data is still far behind conventional methods \citep{carey2024dptabicl}. Therefore, fine-tuning is necessary for tabular data. This has been demonstrated in some tabular classification problems \citep{Dinh2022LIFTLF, hegselmann2023tabllm}.



\section{Details of datasets}
\label{sec:apd_dataset}

We use two main dataset sources: UCI and Kaggle. The UCI machine learning datasets are at larger scales and have been extensively studied by the research community. The datasets from Kaggle are more recent and have been published after the knowledge cutoff date of the Llama-2 models, which are used for \texttt{DP-LLMTGen}'s model backbone.
\begin{table}[h]
    \centering
    \caption{Key characteristics of our experimental datasets.}
    \vspace{2mm}
    \begin{tabularx}{\textwidth}{C|X|c|C|C|C|C}
     \textbf{Short Name} & \textbf{Full Name} & \textbf{Source} & \textbf{Number of samples} & \textbf{Number of categorical features} & \textbf{Number of numerical features} & \textbf{Published after Llama-2 knowledge cutoff date} \\ \hline
      Bank  & Bank marketing & \href{https://archive.ics.uci.edu/dataset/222/bank+marketing}{UCI} & 11162 & 9 & 7 & \xmark \\ \hline
      Adult & Adult census income & \href{https://archive.ics.uci.edu/dataset/2/adult}{UCI} & 48842 & 7 & 6 & \xmark \\ \hline
      Food  & Online food order & \href{https://www.kaggle.com/datasets/sudarshan24byte/online-food-dataset}{Kaggle} & 388 & 6 & 5 & \cmark \\ \hline
      Apple & Apple quality & \href{https://www.kaggle.com/datasets/nelgiriyewithana/apple-quality}{Kaggle} & 4000 & 0 & 7 & \cmark \\ \hline
      Shipping & Ontime Shipping Classification & \href{https://www.kaggle.com/code/nayanack/shipping-data-classification}{Kaggle} & 10999 & 4 & 7 & \cmark \\
    \end{tabularx}
    \label{tab:dataset}
\end{table}

\section{Evaluation Metrics}
\label{sec:app-impl-eval}
\subsection{Statistical fidelity.} 
Our work evaluates statistical fidelity based on the total variation distance (TVD). To calculate TVD, numerical features are quantiled into 20 groups. Let $[x_1, x_2, x_3, ..., x_{N_\mathcal{F}}]$ be a record/sample (i.e., a row of tables), ${N_\mathcal{F}}$ is the number of features. We define $k$-way TVD (also abbreviated as $k$-TVD) where $k$ is the number of features that are involved. For example, $k = 1$ denotes the single feature distributions (Equation~\ref{eq:1tvd}). Meanwhile, $k = 2$ means the total variation distances of joint two-feature distributions (Equation~\ref{eq:2tvd}), where $f_i$ is the feature at index $i$; $\mathcal{F}$ stands for the set of all features; $c$, $c_1$ and $c_2$ denote the feature values; $\mathcal{D}$ and $\mathcal{D'}$ represent for the synthetic and testing datasets. Inductively, we calculate 3-way TVD, 4-way TVD, and 5-way TVD.

\begin{equation}
    1\text{-way TVD}(\mathcal{D}, \mathcal{D'}) = \dfrac{1}{2N_\mathcal{F}} \sum_{f_i\in \mathcal{F}} \sum_{c \in f_i}  \ \left| p(x_i = c|\mathcal{D}) - p(x_i = c|\mathcal{D'}) \right|
    \label{eq:1tvd}
\end{equation}

\begin{equation}
\begin{split}
   2\text{-way TVD}(\mathcal{D}, \mathcal{D'}) = \dfrac{1}{2N_\mathcal{F}\left(N_\mathcal{F} - 1\right)} \sum_{f_i\in \mathcal{F}} \sum_{f_j \in \mathcal{F} \backslash f_i} \sum_{c_1 \in f_i} \sum_{c_2 \in f_j}  |p(x_i = c_1, x_j = c_2| \mathcal{D})  \\ -  p(x_i = c_1, x_j = c_2| \mathcal{D'})| 
\end{split}
\label{eq:2tvd}
\end{equation}

\subsection{Machine learning downstream performance.} 
We evaluate the ML downstream tasks using XGBoost \citep{xgboost} which has achieved state-of-the-art performance in many benchmarks for tabular data \citep{SHWARTZZIV202284}. We perform a grid search for 5-fold cross validation to achieve the robust and realistic testing performance. The grid search hyperparameters include number of estimators \{100, 200, 300\}, maximum tree depth \{3, 5, 10, 20\}, and learning rate \{0.01, 0.05, 0.1\}.

\subsection{Fairness metrics.} 
\label{sec:app-fairness-metric}
\textbf{Demographic Parity}, also known as statistical parity, is a fairness criterion where  particular attributes or model predictions are equally distributed across different demographic groups. The demographic parity difference of a mechanism $\mathcal{M}$ is calculated in Equation~\ref{eq:dpm}, where $x$ is the features of a sample and $a$, $b$ are demographic groups. Similarly, the demographic parity difference of a dataset can be formalized as Equation~\ref{eq:dpd} where $Y(x)$ represents for the label of $x$.

\begin{equation}
   \mathcal{M}\_\text{DPDiff} = |p(\mathcal{M}(x) = 1| A = a) - p(\mathcal{M}(x) = 1| A = b)|
    \label{eq:dpm}
\end{equation}

\begin{equation}
   \text{Data\_DPDiff} = |p(Y(x) = 1| A = a) - p(Y(x) = 1| A = b)|
    \label{eq:dpd}
\end{equation}

\textbf{Equalized Odds} can guarantee a machine learning model performs equally well for different groups. Let $TPR_a$, $TPR_b$, $FPR_a$, and $FPR_b$ be the True Positive Rate and the False Positive Rate for groups $a$, $b$. The Equalized Odds difference can be defined as the follow:
\begin{equation}
   \text{EO\_DPDiff} = \text{max}(|TPR_a - TPR_b|, |FPR_a - FPR_b|)
    \label{eq:eom}
\end{equation}

\section{Implementation details}
\label{sec:app-impl}
For the GAN-based methods, we implemented using SmartNoise SDK \url{https://github.com/opendp/smartnoise-sdk} which is a part of OpenDP library \citep{Shoemate_OpenDP_Library}. For RAP, RAP++, and GSD, we use the code provided the authors. For the baselines, we used recommended hyperparameters in the papers and default hyperparameters of their github repositories (if no recommendation in the papers). \texttt{DP-LLMTGen}'s implementation is based one Huggingface~\citep{wolf-etal-2020-transformers} for LLMs, Pytorch~\citep{10.5555/3454287.3455008}, and Opacus~\citep{opacus} for DPSGD. Regarding to the hyperparameters of \texttt{DP-LLMTGen}, for the first fine-tuning stage (i.e., format learning), we fine tune LLMs using Adam optimizer \citep{kingma2017adam} and the learning rate at 1e-4. We fine tune 5 epochs for the Bank, Adult, and Shipping datasets and 10 epochs for the Food and Apple datasets. The reason is the number of samples in the Food and Apple datasets is small, so the LLMs need more epochs to learn. For the second stage, $\alpha$ iws set at 0.65; the learning rate is 5e-4. For the Adult dataset, we fine tune for 2 epochs. The remainder datasets are fined tune with 4 epochs. All the experiments are conducted at least three times by different random train test splits to ensure the robust results. It should be noted that the size of the synthetic datasets matches that of the training datasets.

\section{In-Depth Analyses of Overall Evaluation}
\subsection{Training time of \texttt{DP-LLMTGen}}
The main factors affect to the training time is the dataset size, sample length, and number of epochs. Various datasets have different number of samples. Moreover, after converting to sentences (using Tabular-to-Text decoding), the sentence length (i.e, number of tokens) depends on the number of features and categorical/numerical values. It is worth noting that, given the same dataset, there is no significant difference of training time when using SGD or DPSGD. It is because, LLMs have billions of parameters, there are only a dozen of millions of trainable parameters when using LoRA. Most of the training time is spent on the forward pass of the LLMs rather than the backward pass for LoRA parameters.

\begin{table}[h]
    \centering
    \caption{Training time of \text{DP-LLMTGen} on a single H100 GPU}
    \vspace{1.5mm}
    \begin{tabular}{c|c}
       \textbf{Dataset}  &  Training time for \textbf{1} epoch (Llama 2 7b)  \\ \hline
        Bank & 15 minutes \\ 
        Adult & 1 hour 45 minutes \\
        Food & 1 minute\\
        Apple & 12 minutes \\
        Shipping & 16 minutes 
    \end{tabular}
    \label{tab:training-time}
\end{table}

\subsection{Distance to closest record histograms}
\label{sec:dcr}
\begin{figure}[h]
\centering
\begin{subfigure}{.245\textwidth}
  \centering
  \includegraphics[width=\linewidth]{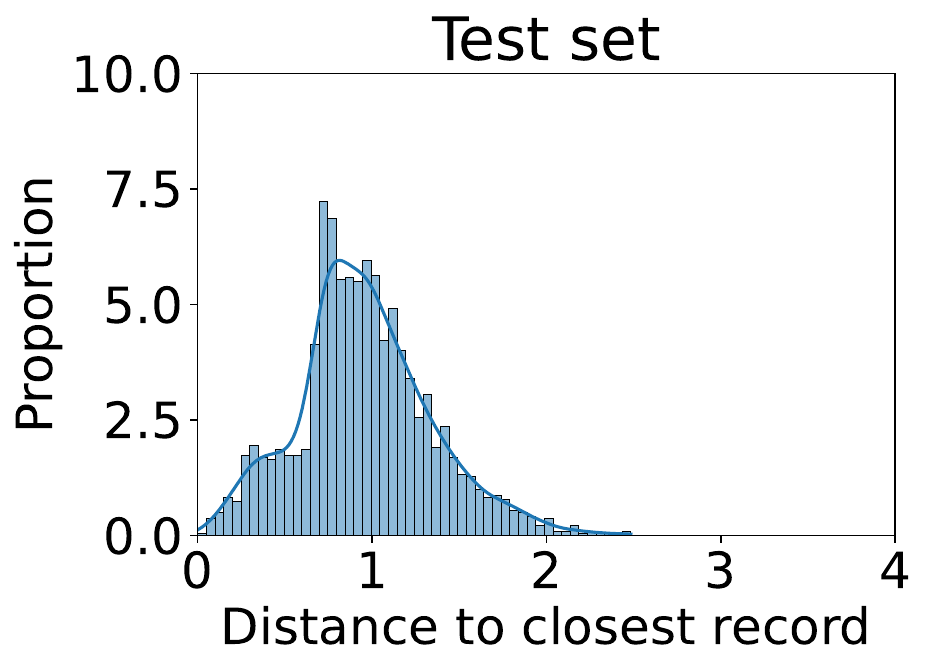}
\end{subfigure}%
\begin{subfigure}{.245\textwidth}
  \centering
  \includegraphics[width=\linewidth]{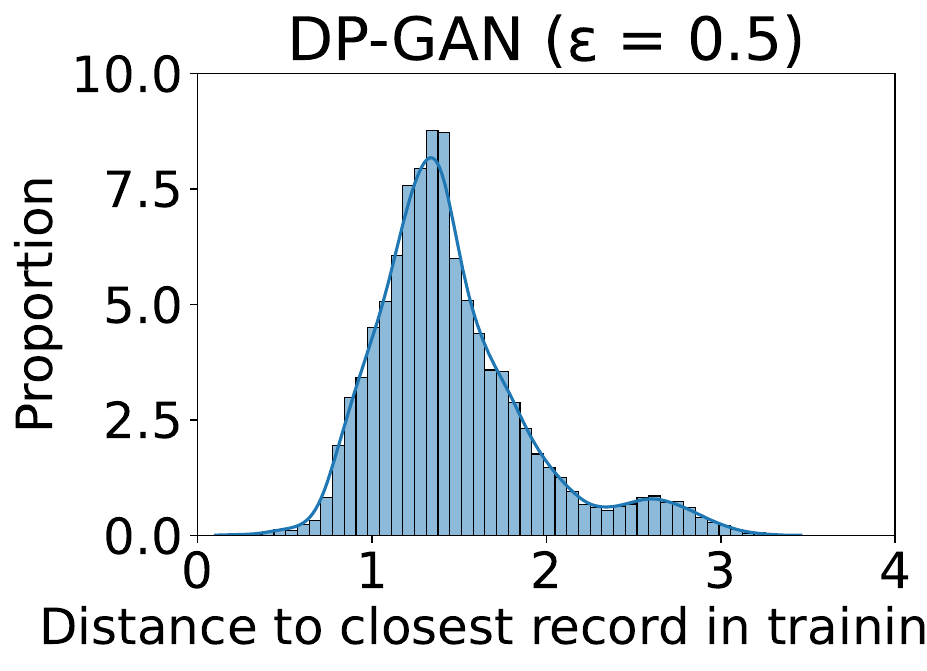}
\end{subfigure}
\begin{subfigure}{.245\textwidth}
  \centering
  \includegraphics[width=\linewidth]{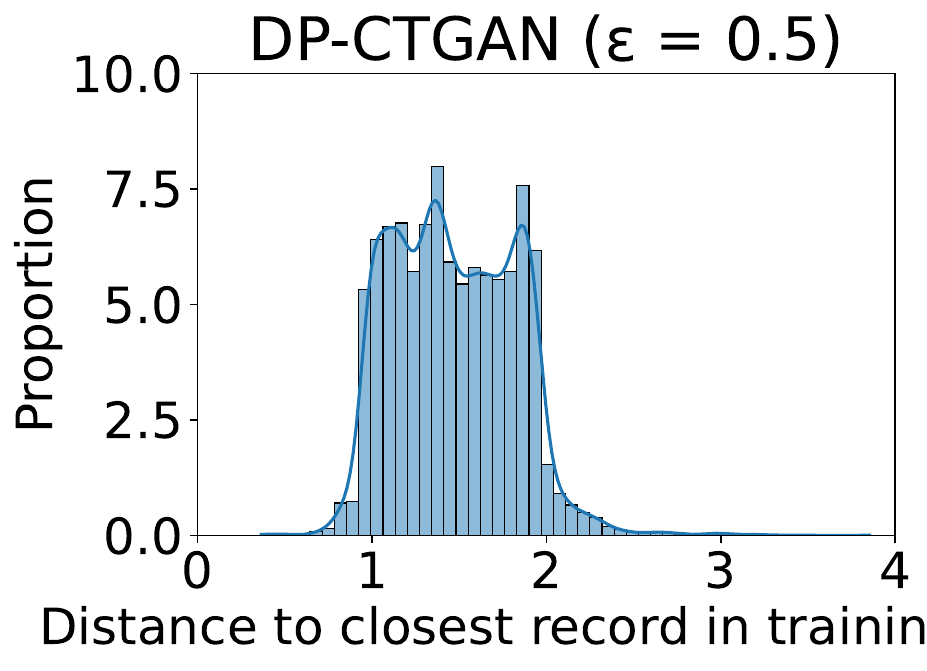}
\end{subfigure}%
\begin{subfigure}{.245\textwidth}
  \centering
  \includegraphics[width=\linewidth]{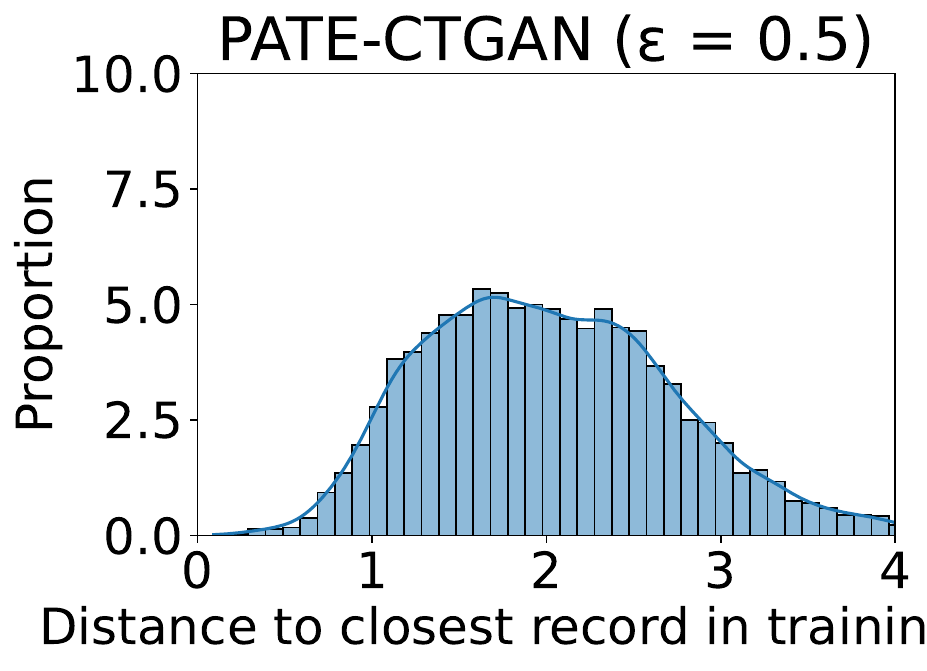}
\end{subfigure} \\
\begin{subfigure}{.245\textwidth}
  \centering
  \includegraphics[width=\linewidth]{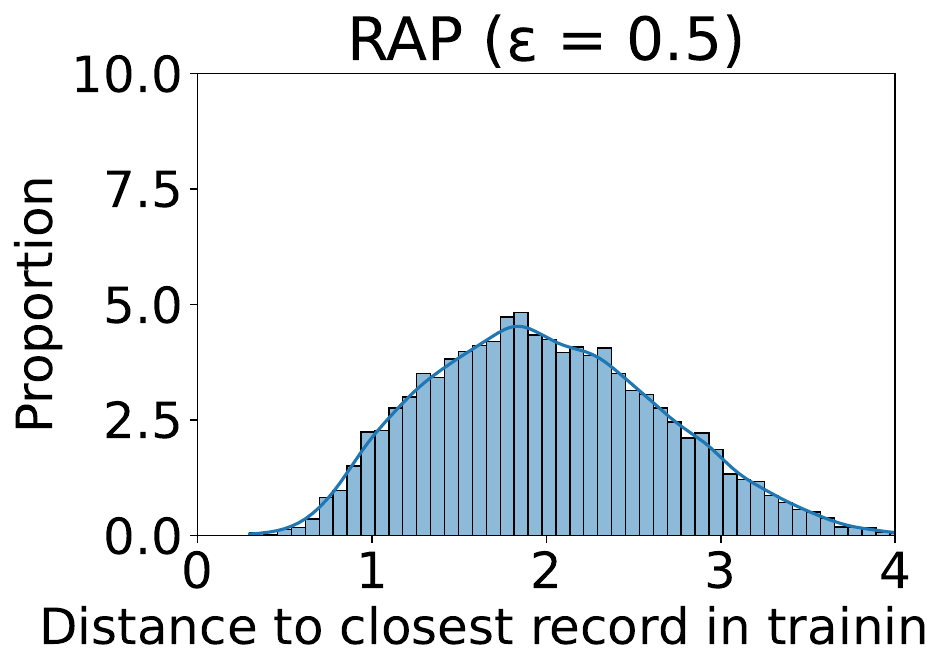}
\end{subfigure}
\begin{subfigure}{.245\textwidth}
  \centering
  \includegraphics[width=\linewidth]{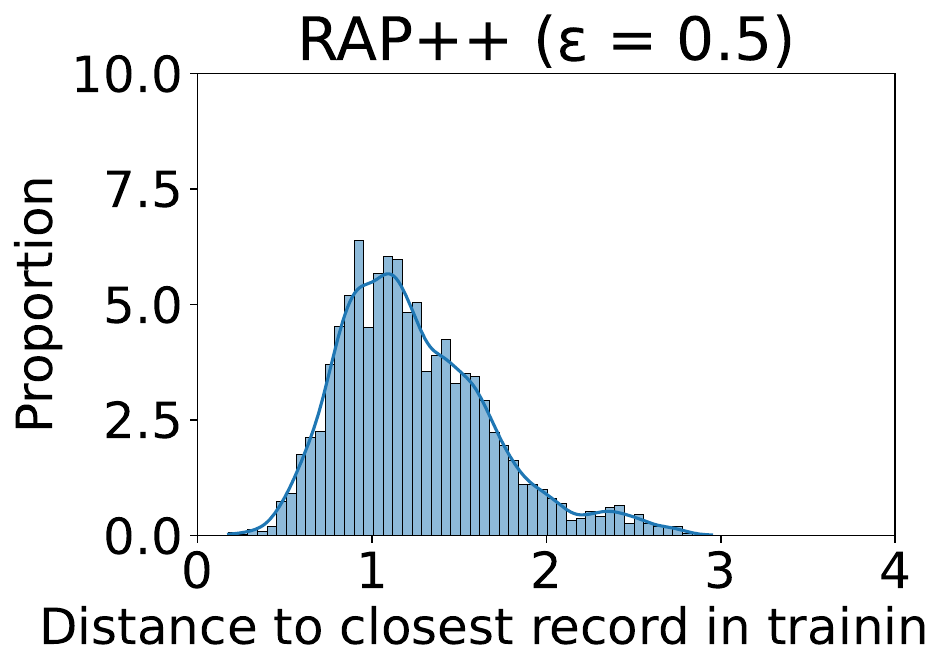}
\end{subfigure}
\begin{subfigure}{.245\textwidth}
  \centering
  \includegraphics[width=\linewidth]{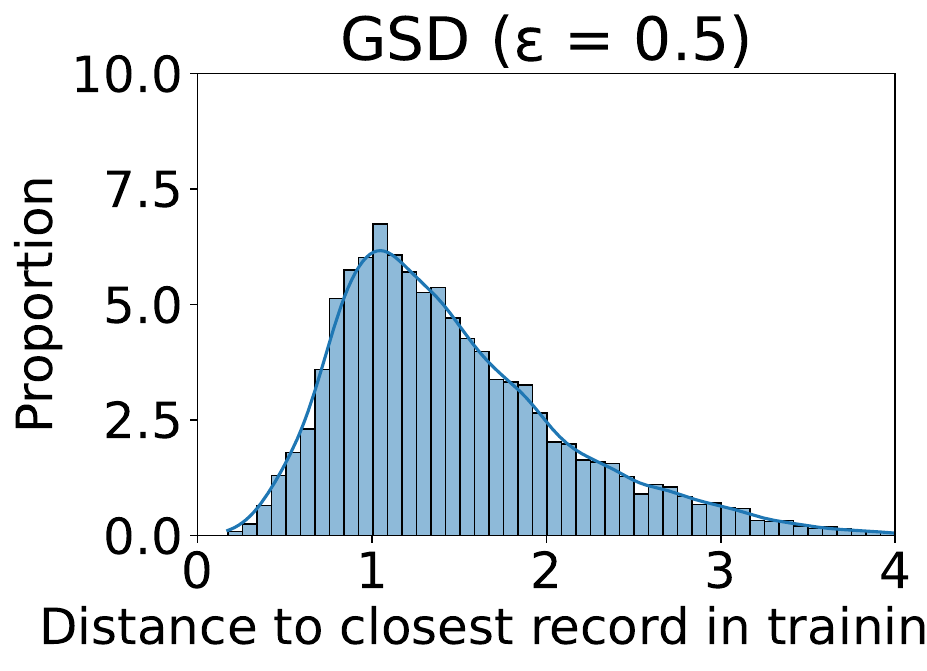}
\end{subfigure}
\begin{subfigure}{.245\textwidth}
  \centering
  \includegraphics[width=\linewidth]{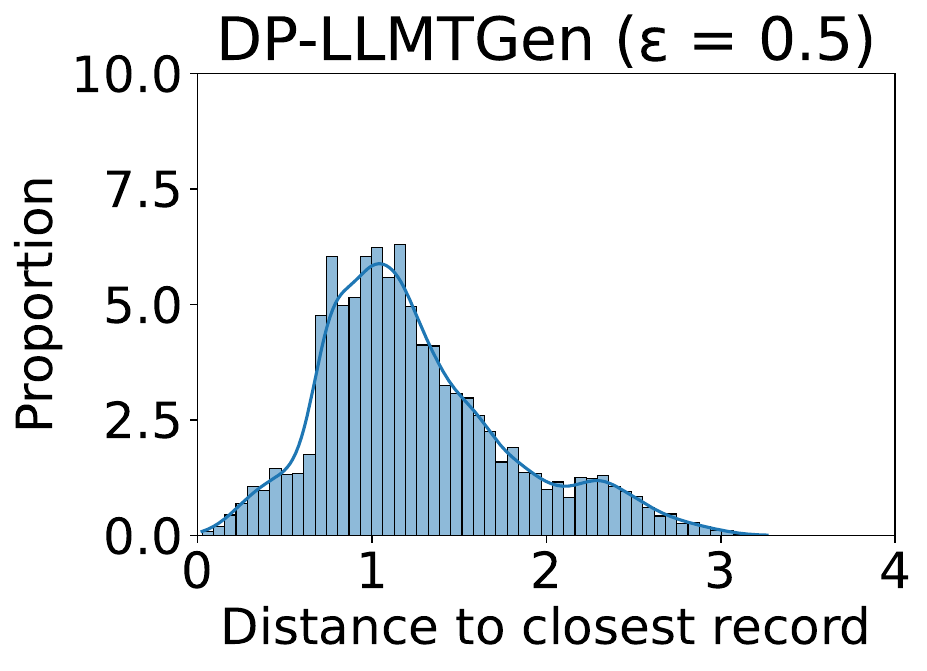}
\end{subfigure}
\caption{Histograms of the distance to closest record between the testing/synthetic and training sets for the Shipping dataset. The privacy budget is set at $\epsilon = 0.5$}
\label{fig:dcr0.5}
\end{figure}

\begin{figure}[h]
\centering
\begin{subfigure}{.245\textwidth}
  \centering
  \includegraphics[width=\linewidth]{figures/shipping/dcr/distance_to_closest_record_test.pdf}
\end{subfigure}%
\begin{subfigure}{.245\textwidth}
  \centering
  \includegraphics[width=\linewidth]{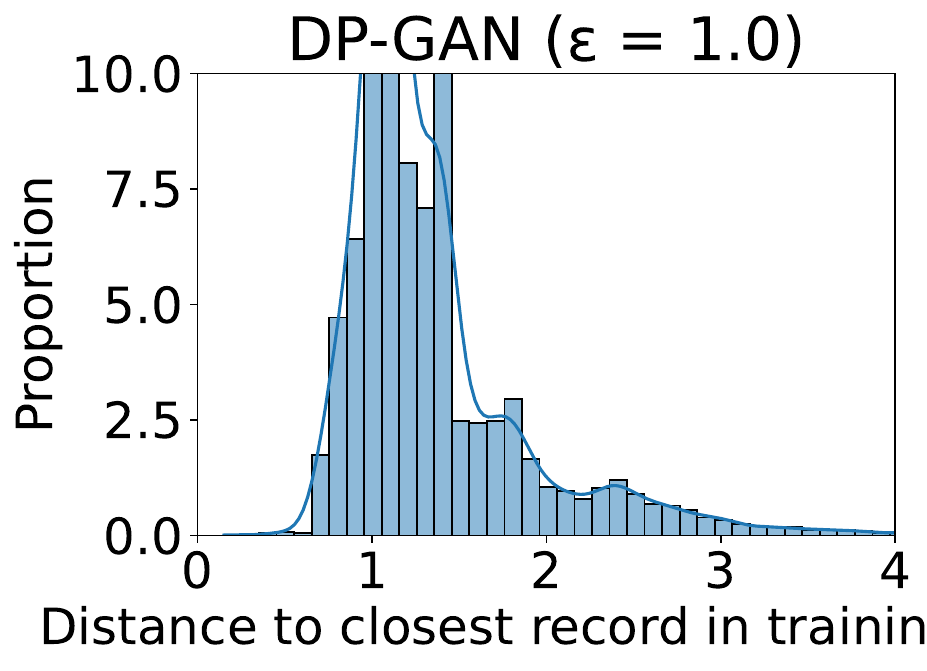}
\end{subfigure}
\begin{subfigure}{.245\textwidth}
  \centering
  \includegraphics[width=\linewidth]{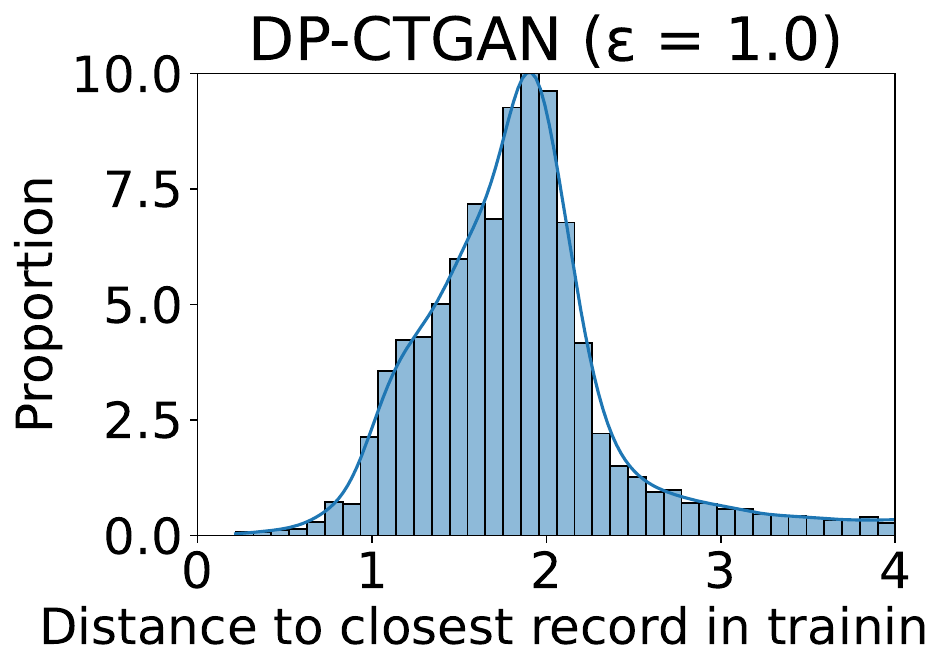}
\end{subfigure}%
\begin{subfigure}{.245\textwidth}
  \centering
  \includegraphics[width=\linewidth]{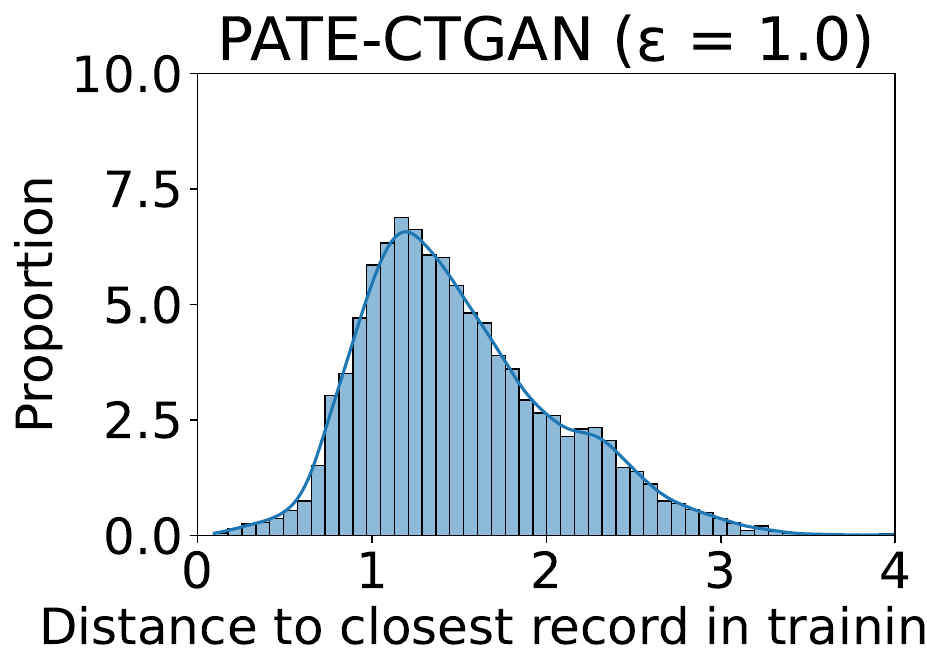}
\end{subfigure} \\
\begin{subfigure}{.245\textwidth}
  \centering
  \includegraphics[width=\linewidth]{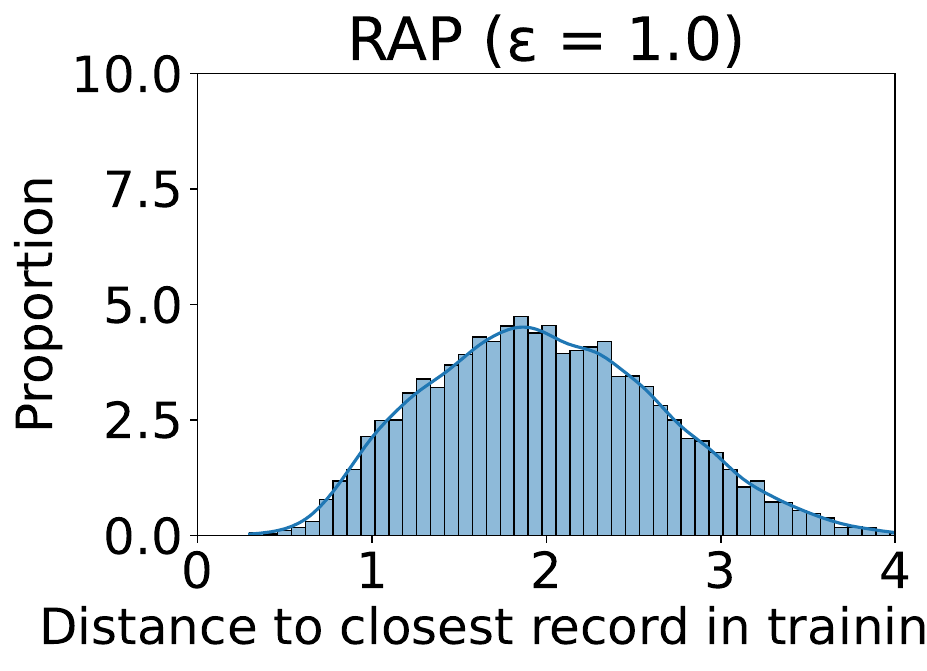}
\end{subfigure}
\begin{subfigure}{.245\textwidth}
  \centering
  \includegraphics[width=\linewidth]{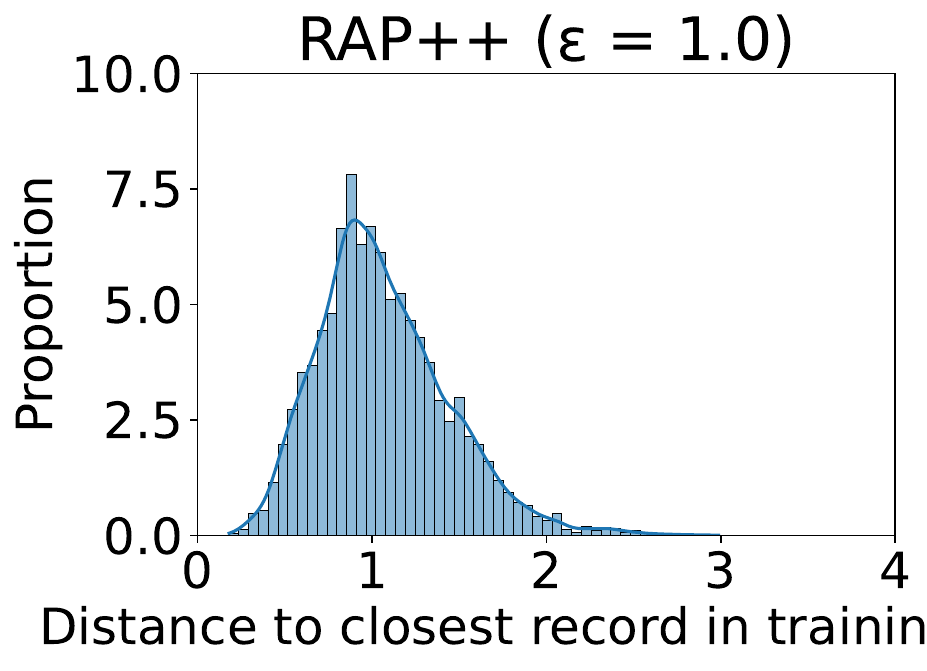}
\end{subfigure}
\begin{subfigure}{.245\textwidth}
  \centering
  \includegraphics[width=\linewidth]{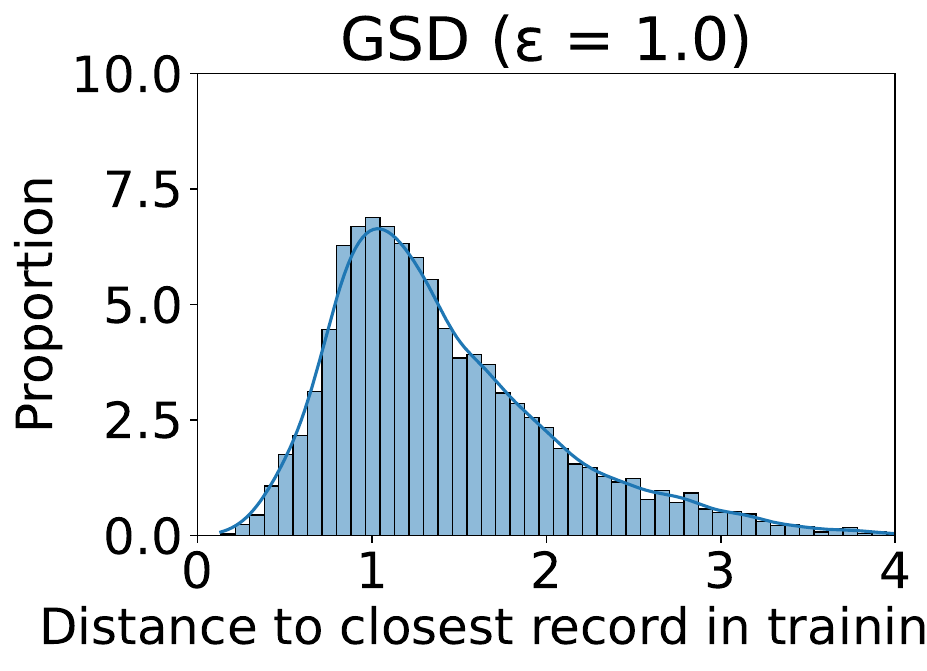}
\end{subfigure}
\begin{subfigure}{.245\textwidth}
  \centering
  \includegraphics[width=\linewidth]{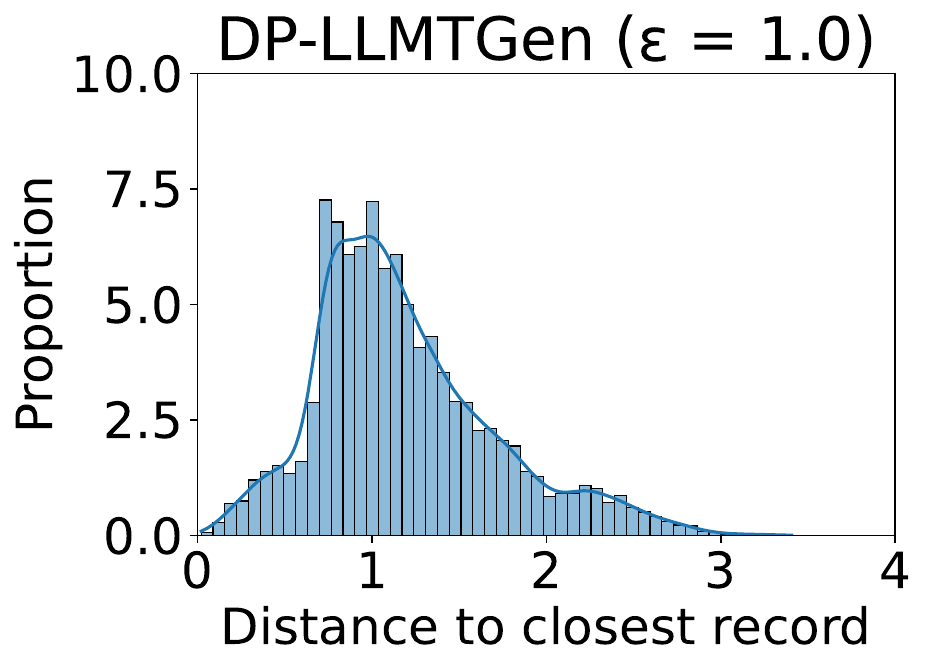}
\end{subfigure}
\caption{Histograms of the distance to closest record between the testing/synthetic and training sets for the Shipping dataset. The privacy budget is set at $\epsilon = 1.0$}
\label{fig:dcr1.0}
\end{figure}

Since LLMs have a strong capability to memorize the fine-tuning datasets, it is essential to verify that LLMs actually generate new samples or just replicate the training datasets. In this study, to demonstrate that LLMs are capable of generating novel samples, we calculate and distance from each record of the synthetic datasets to the closest record in the corresponding training datasets. We normalize the datasets before distance calculation to ensure equal importance of all features. A distance at 0 represents for a replication from the training datasets to the generation. Figures~\ref{fig:dcr0.5} and~\ref{fig:dcr1.0} depicts the histograms of distances for the Shipping dataset under different privacy budgets. Generally, there is no replication of training data when using \texttt{DP-LLMTGen}. When the privacy constraints are more relaxed, all methods tend to generate samples that are closer to each other than those produced under a stricter privacy budget. Additionally, \texttt{DP-LLMTGen} and RAP++ generates less outliers compared to the other methods.

\newpage
\subsection{Single feature distribution visualizations}
\begin{figure}[h]
\centering
\begin{subfigure}{.46\textwidth}
  \centering
  \includegraphics[width=\linewidth]{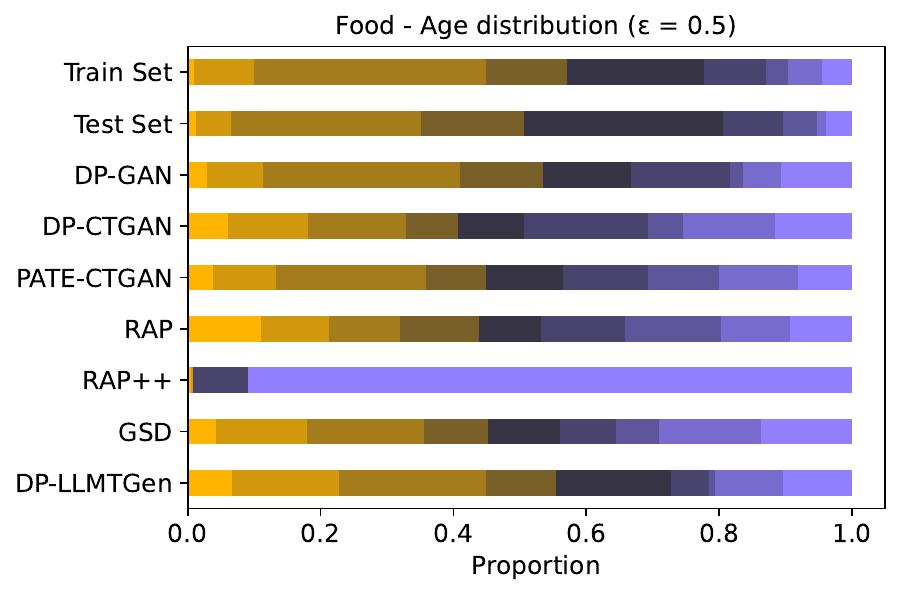}\\
  \includegraphics[width=\linewidth]{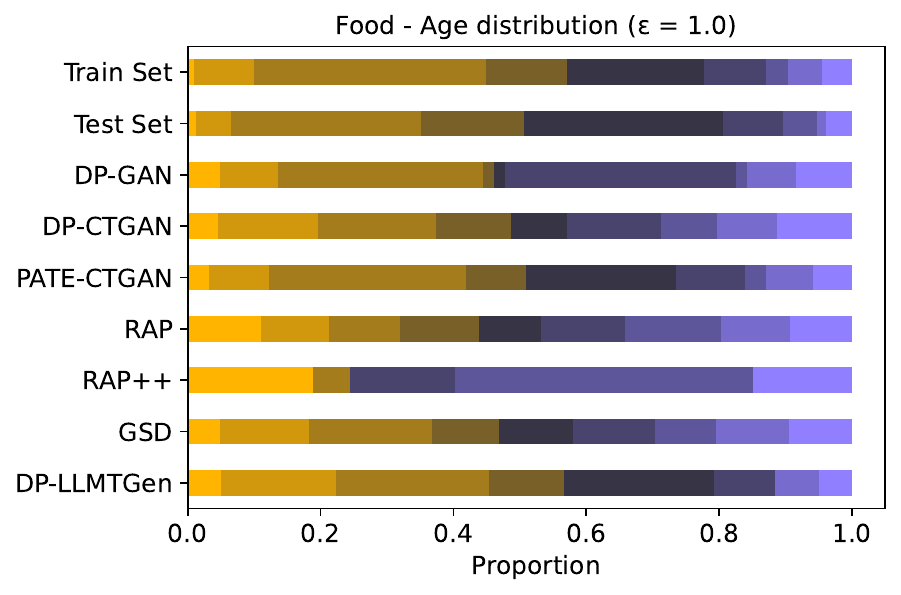}
  \caption{Age distribution}
  \label{fig:1tvd-age}
\end{subfigure}%
\hfill
\begin{subfigure}{.46\textwidth}
  \centering
  \includegraphics[width=\linewidth]{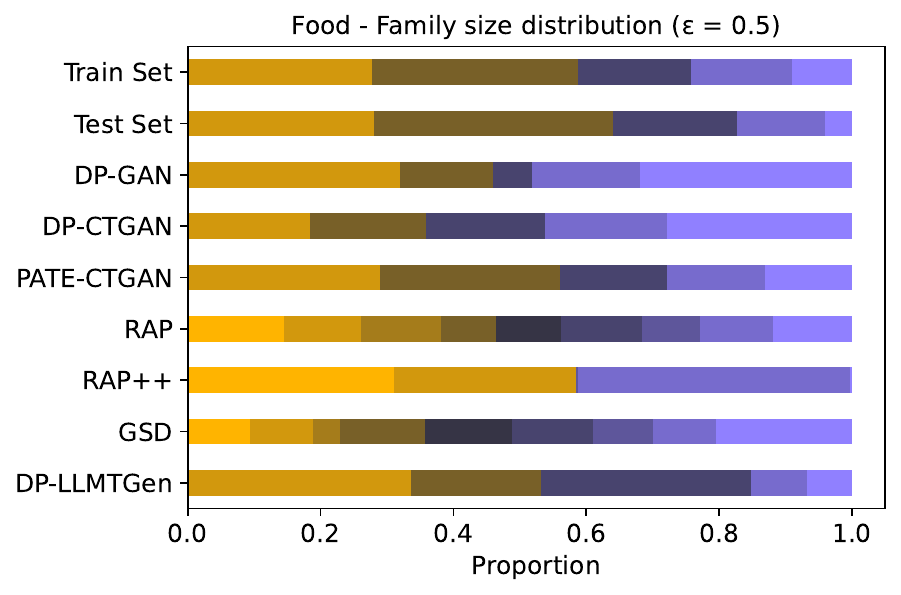}\\
  \includegraphics[width=\linewidth]{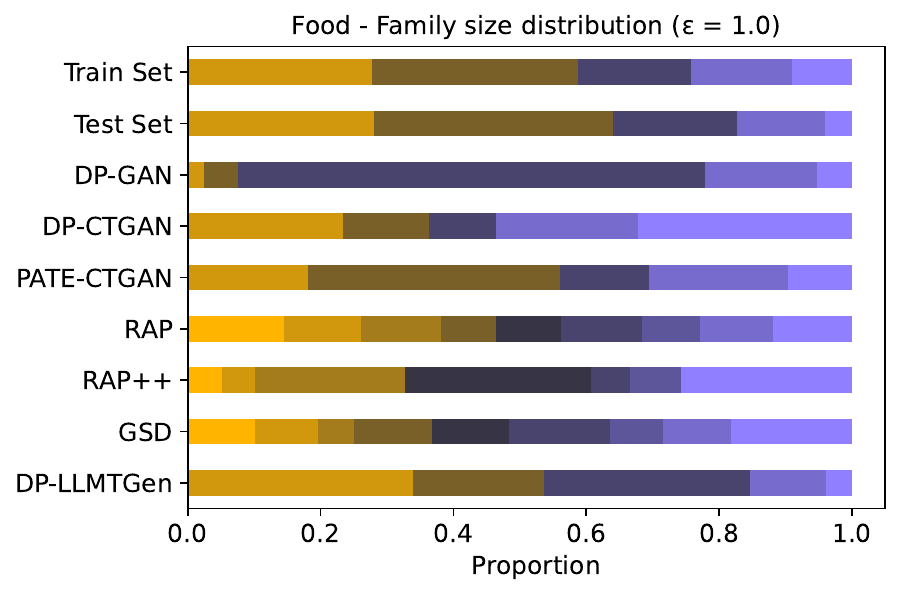}
  \caption{Family size distribution}
\end{subfigure} \\
\begin{subfigure}{.46\textwidth}
  \centering
  \includegraphics[width=\linewidth]{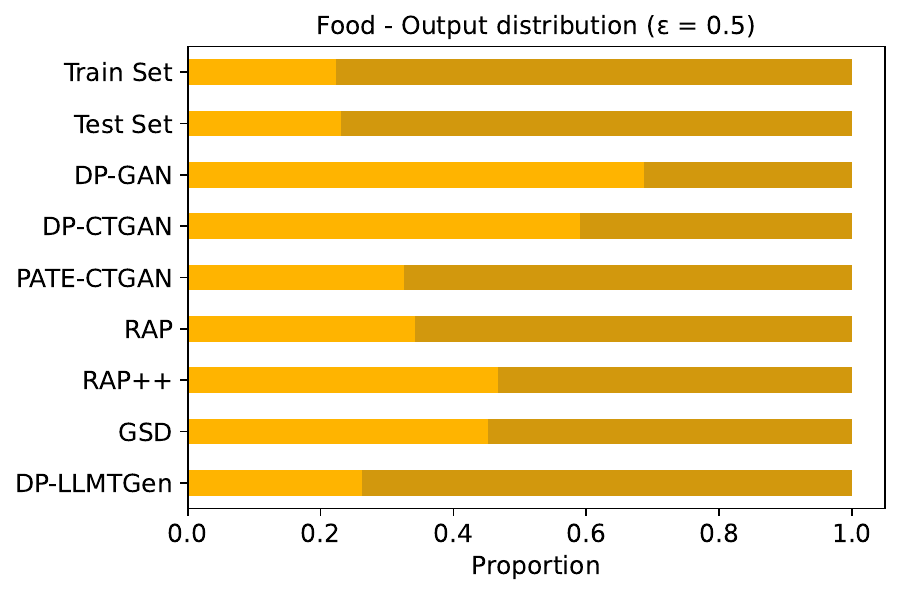}\\
  \includegraphics[width=\linewidth]{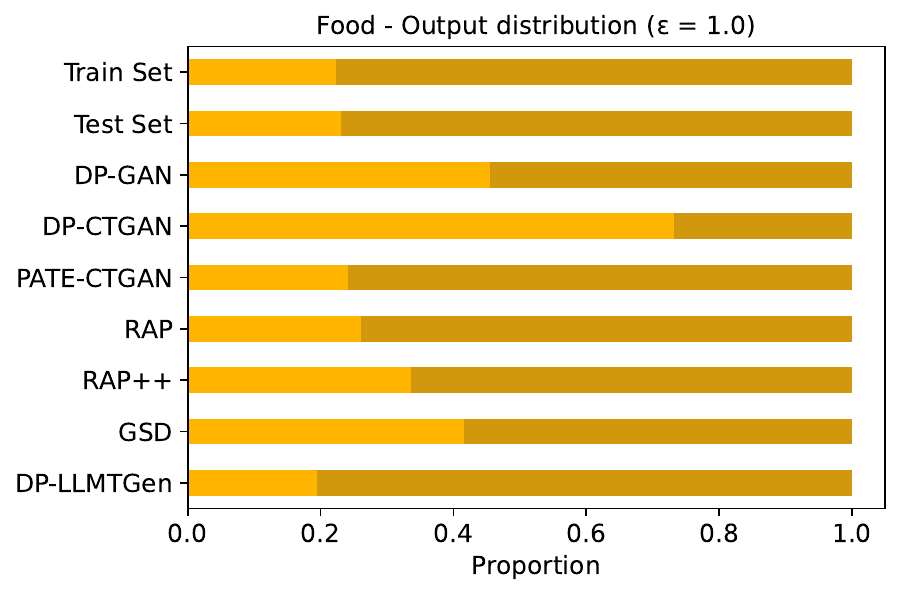}
  \caption{Output distribution}
\end{subfigure}%
\hfill
\begin{subfigure}{.46\textwidth}
  \centering
  \includegraphics[width=\linewidth]{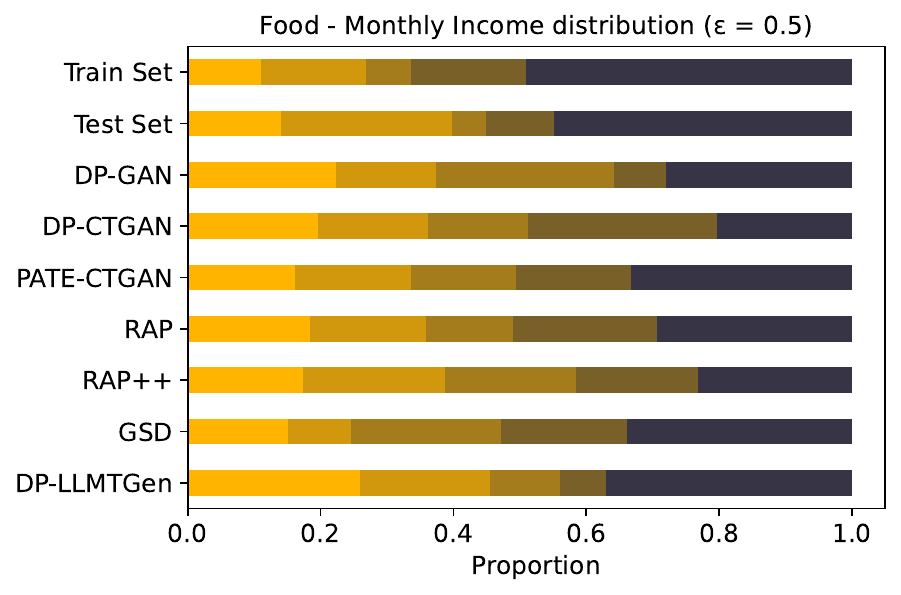}\\
  \includegraphics[width=\linewidth]{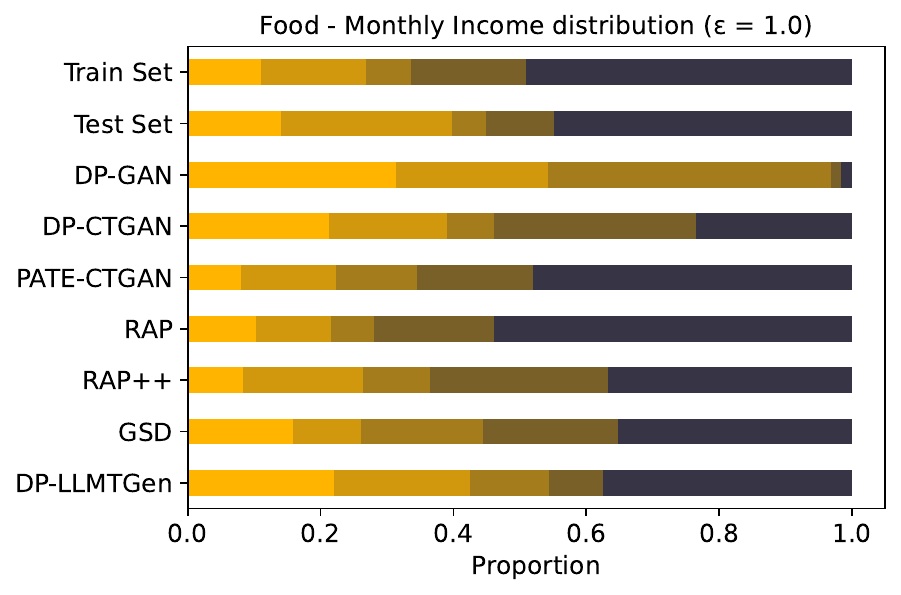}
  \caption{Income distribution}
\end{subfigure}
\caption{[Food dataset] Single feature distributions of synthetic and real sets. Numerical features are binned into 10 groups for this visualization. The colors represent for categorical values and numerical groups.}
\label{fig:food-dist}
\end{figure}

\begin{figure}[h]
\centering
\begin{subfigure}{.48\textwidth}
  \centering
  \includegraphics[width=\linewidth]{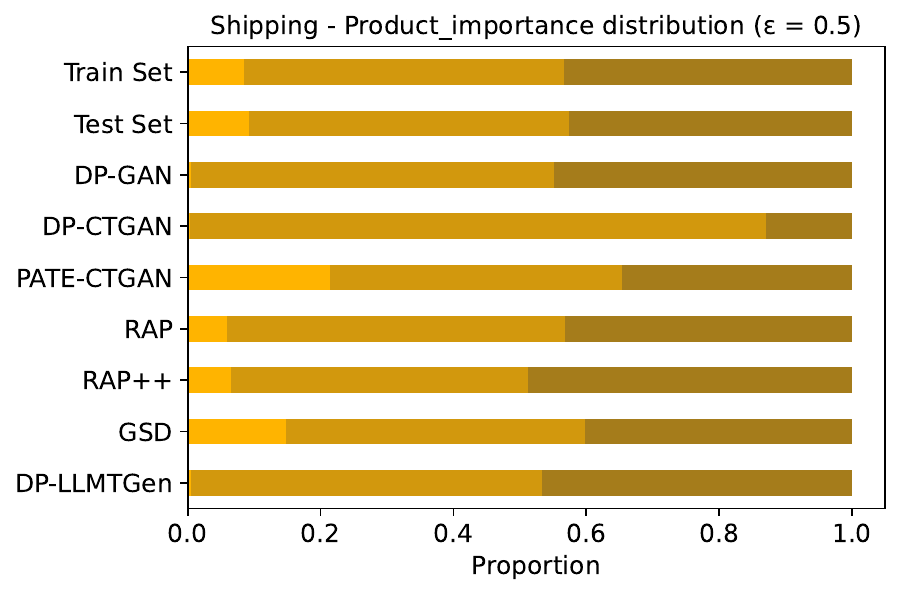}\\
  \includegraphics[width=\linewidth]{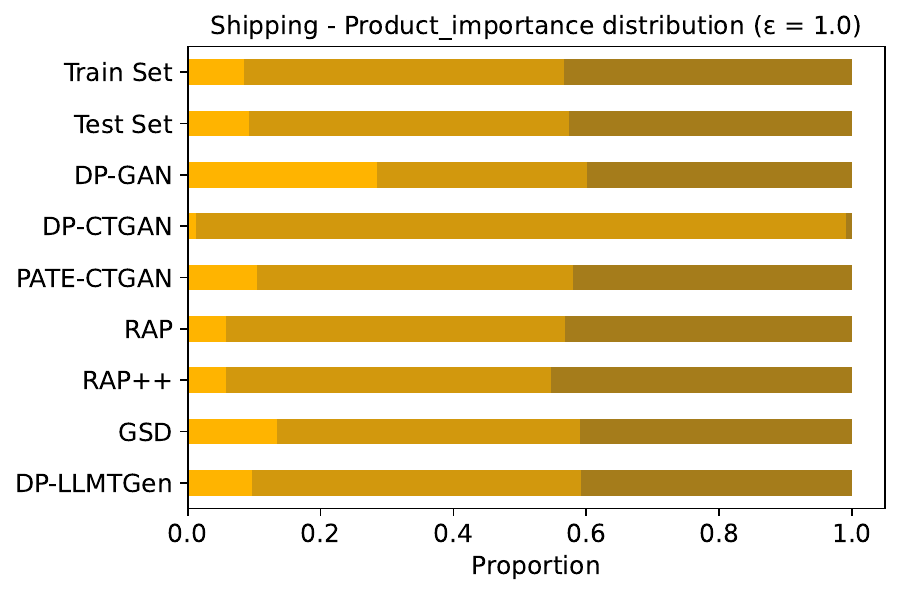}
  \caption{Product importance distribution}
\end{subfigure}%
\hfill
\begin{subfigure}{.48\textwidth}
  \centering
  \includegraphics[width=\linewidth]{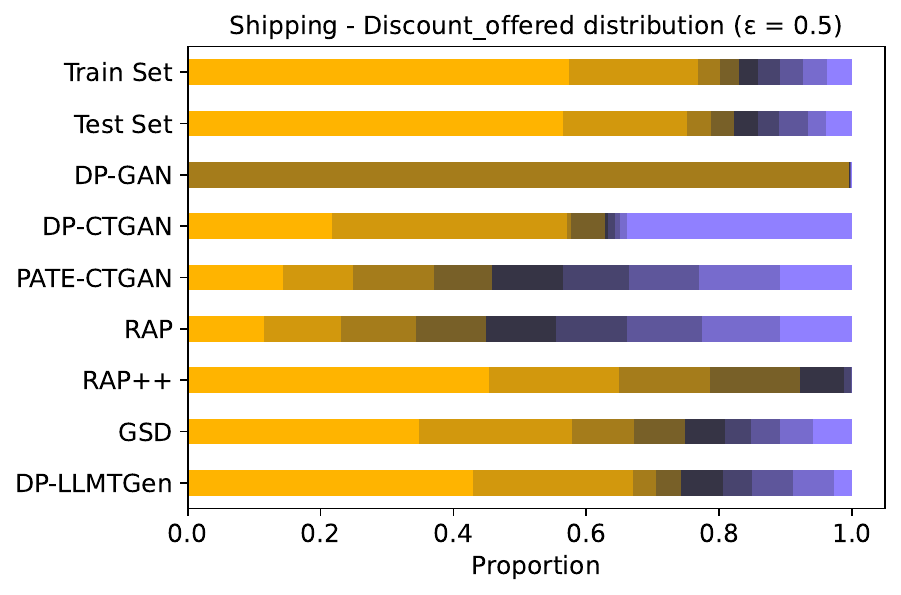}\\
  \includegraphics[width=\linewidth]{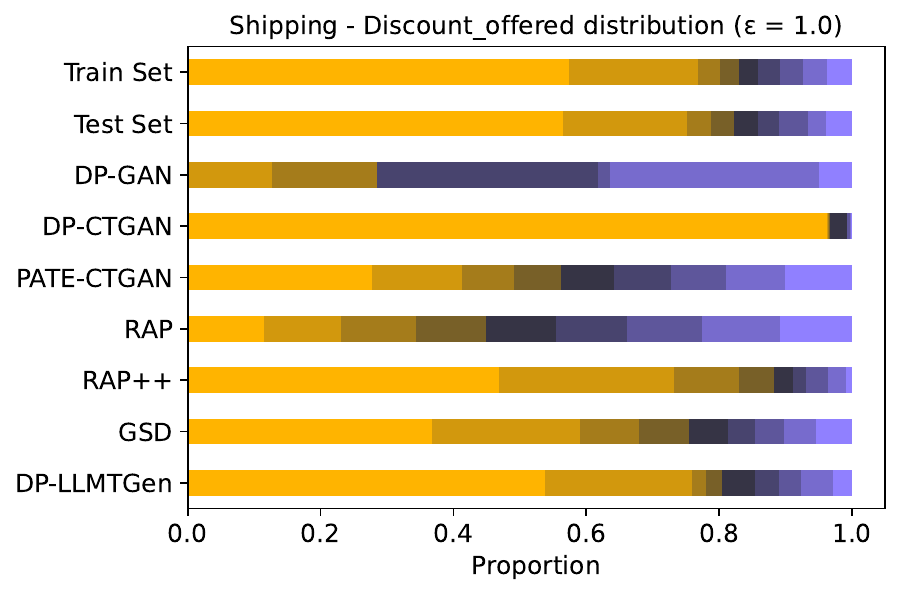}
  \caption{Discount offered distribution}
  \label{fig:1vd-discount}
\end{subfigure} \\
\begin{subfigure}{.48\textwidth}
  \centering
  \includegraphics[width=\linewidth]{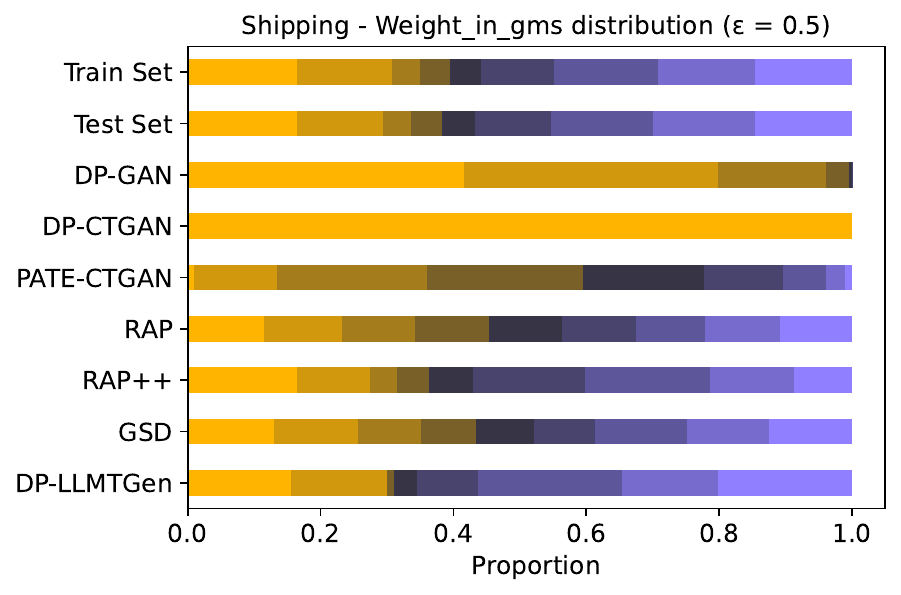}\\
  \includegraphics[width=\linewidth]{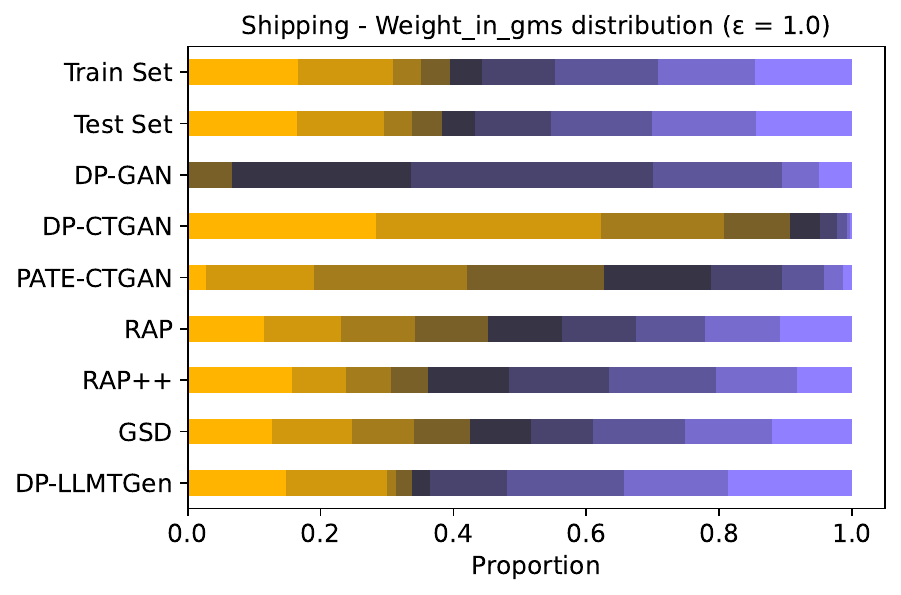}
  \caption{Weight in gms distribution}
\end{subfigure}%
\hfill
\begin{subfigure}{.48\textwidth}
  \centering
  \includegraphics[width=\linewidth]{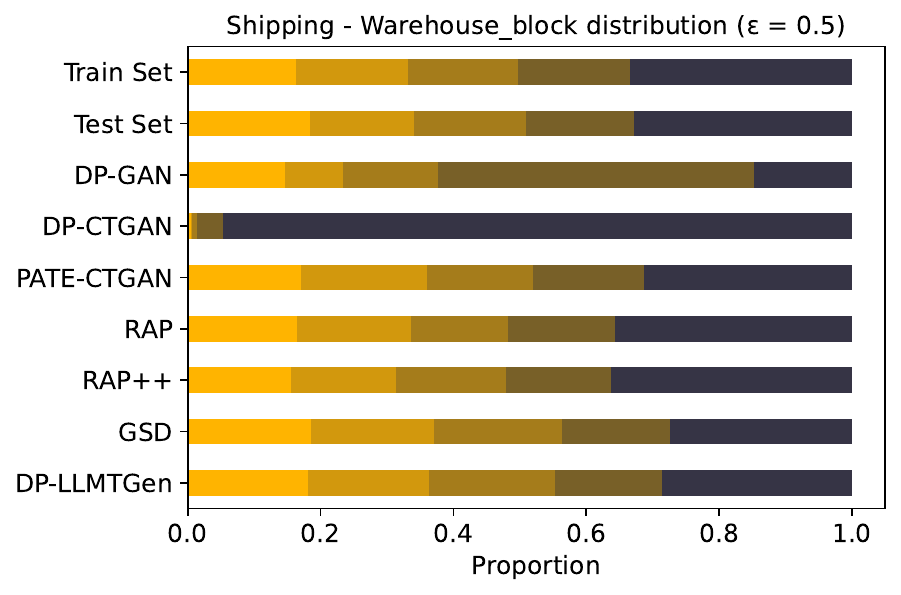}\\
  \includegraphics[width=\linewidth]{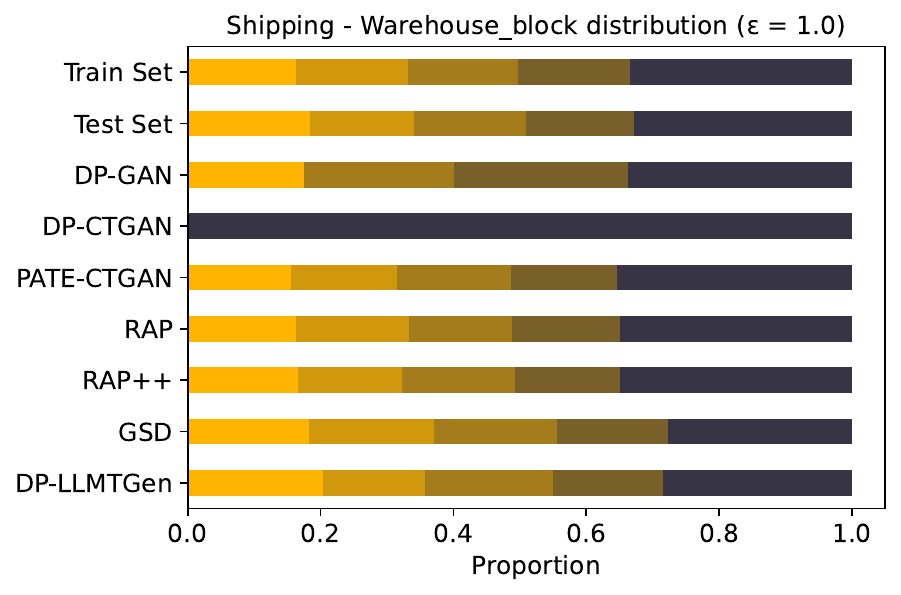}
  \caption{Warehouse block distribution}
\end{subfigure}
\caption{[Shipping dataset] Single feature distributions of synthetic and real sets. Categorical features are binned into 10 groups for this visualization. The colors represent for categorical values and numerical groups.}
\label{fig:shipping-dist}
\end{figure}

\texttt{DP-LLMTGen} produces better single features distributions than the baselines. That leads to a lower 1-way TVD. Especially, \texttt{DP-LLMTGen} replicates well for minority feature categories while RAP, RAP++, and GSD amplify them (Figures~\ref{fig:1tvd-age} and~\ref{fig:1vd-discount}).
\label{sec:1tvd}

\clearpage
\subsection{Joint feature distribution visualizations}
\label{sec:2tvd}

\begin{figure}[h]
\centering
\begin{subfigure}{.32\textwidth}
  \centering
  \includegraphics[width=\linewidth]{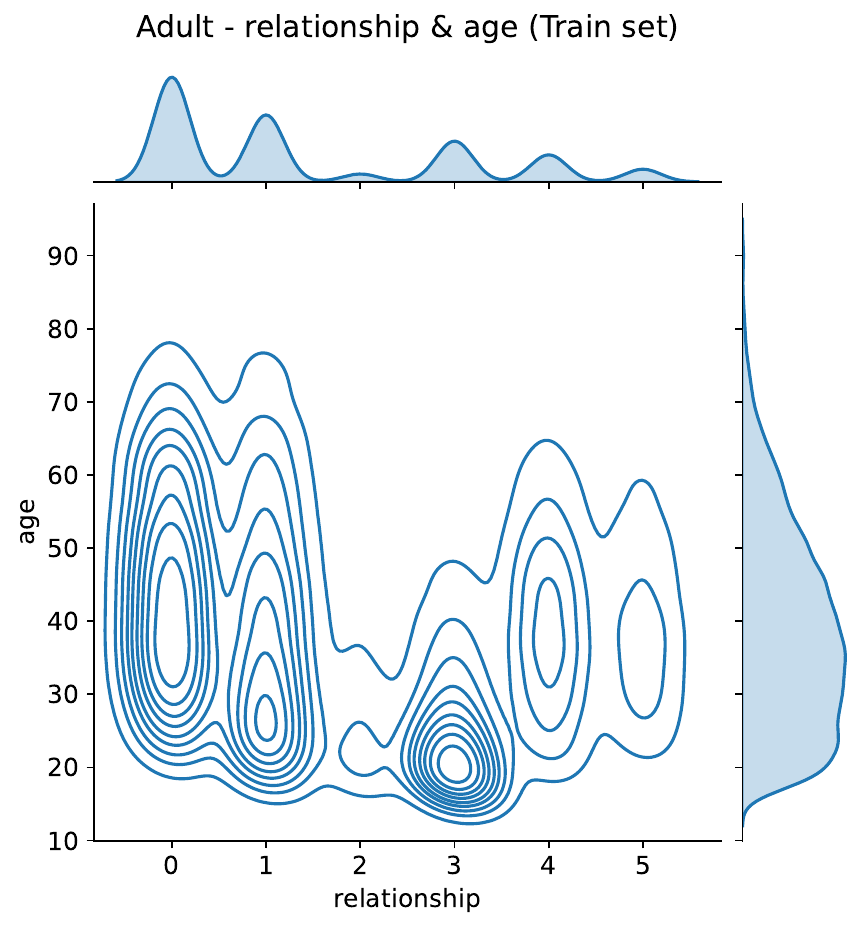}
\end{subfigure}
\begin{subfigure}{.32\textwidth}
  \centering
  \includegraphics[width=\linewidth]{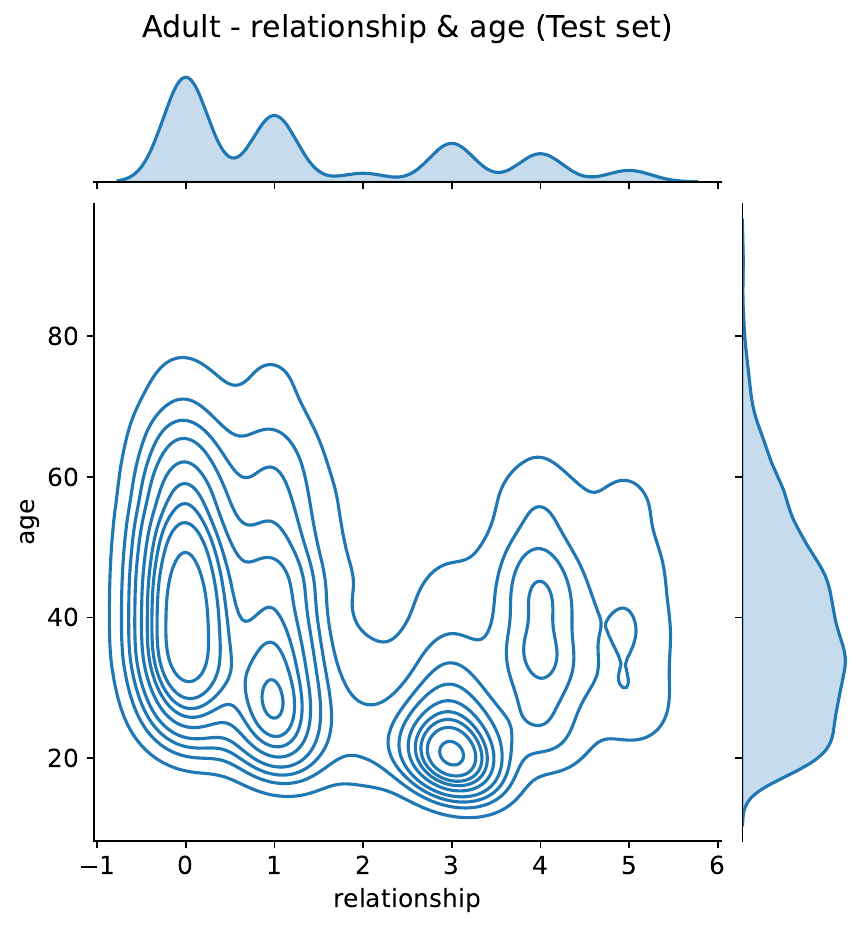}
\end{subfigure}
\begin{subfigure}{.32\textwidth}
  \centering
  \includegraphics[width=\linewidth]{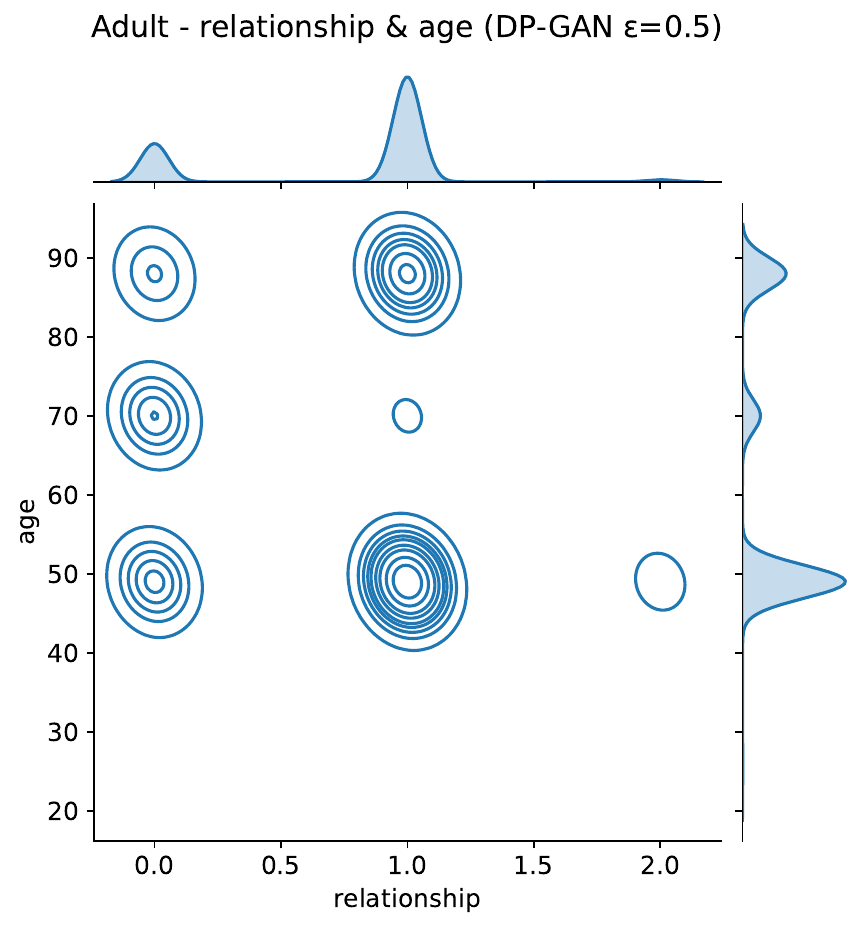}
\end{subfigure} \vspace{5mm}  \\
\begin{subfigure}{.32\textwidth}
  \centering
  \includegraphics[width=\linewidth]{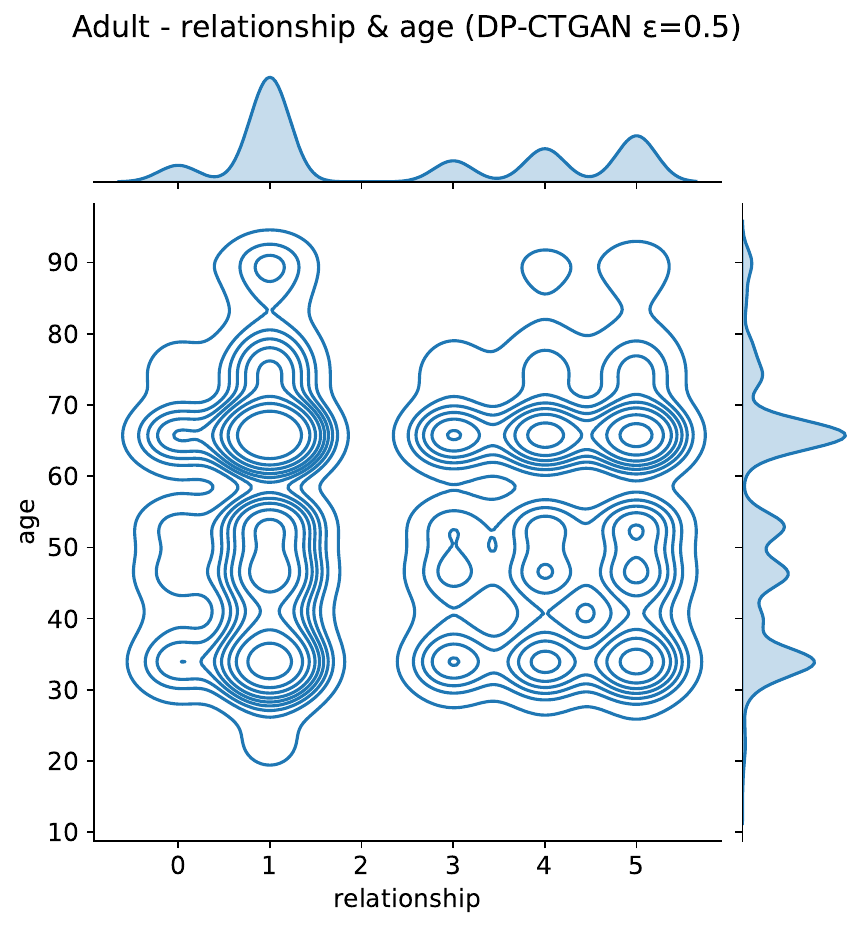}
\end{subfigure}
\begin{subfigure}{.32\textwidth}
  \centering
  \includegraphics[width=\linewidth]{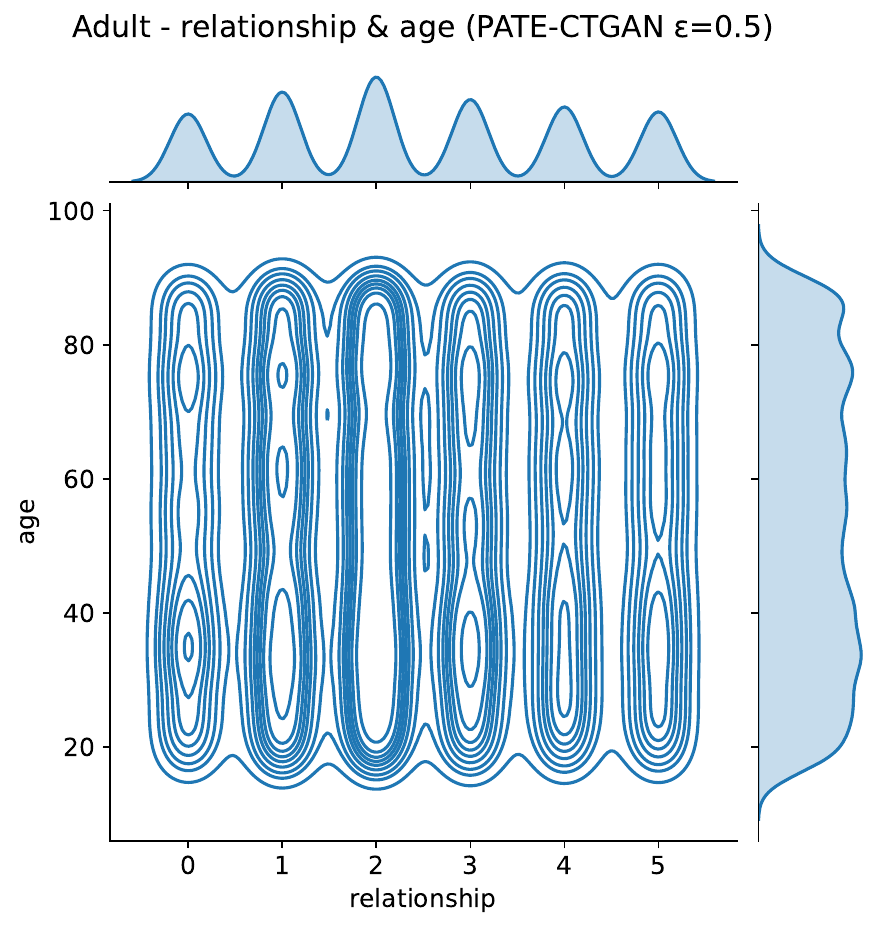}
\end{subfigure} 
\begin{subfigure}{.32\textwidth}
  \centering
  \includegraphics[width=\linewidth]{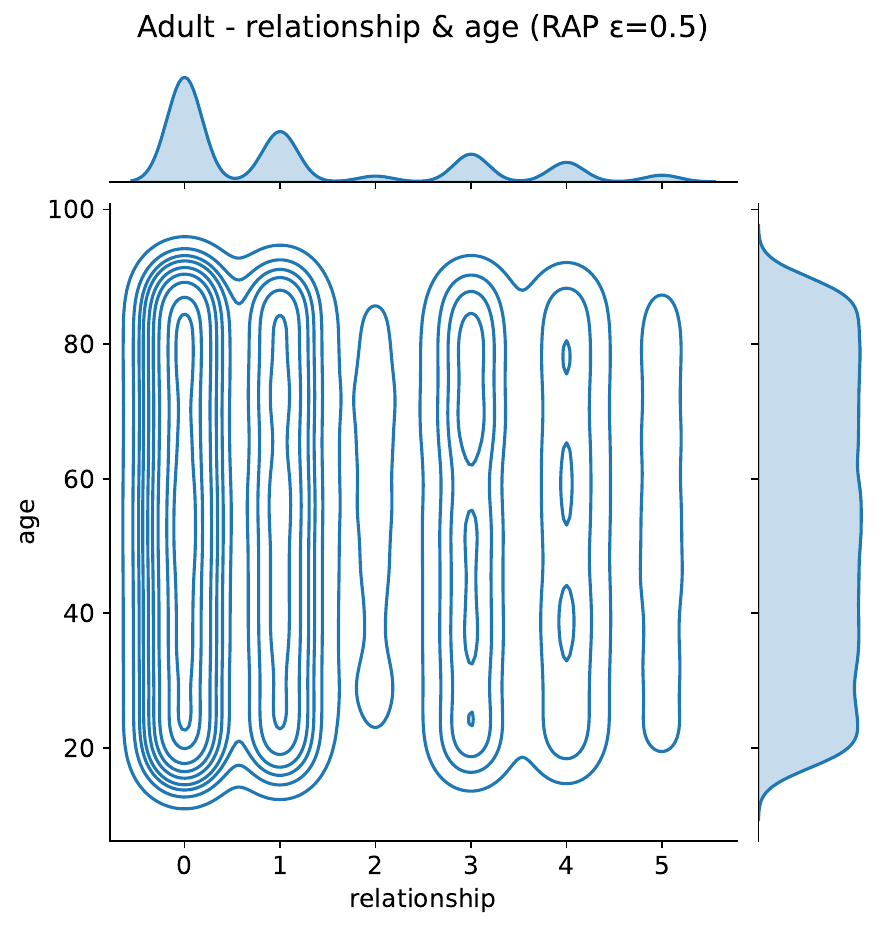}
\end{subfigure} \vspace{5mm} \\
\begin{subfigure}{.32\textwidth}
  \centering
  \includegraphics[width=\linewidth]{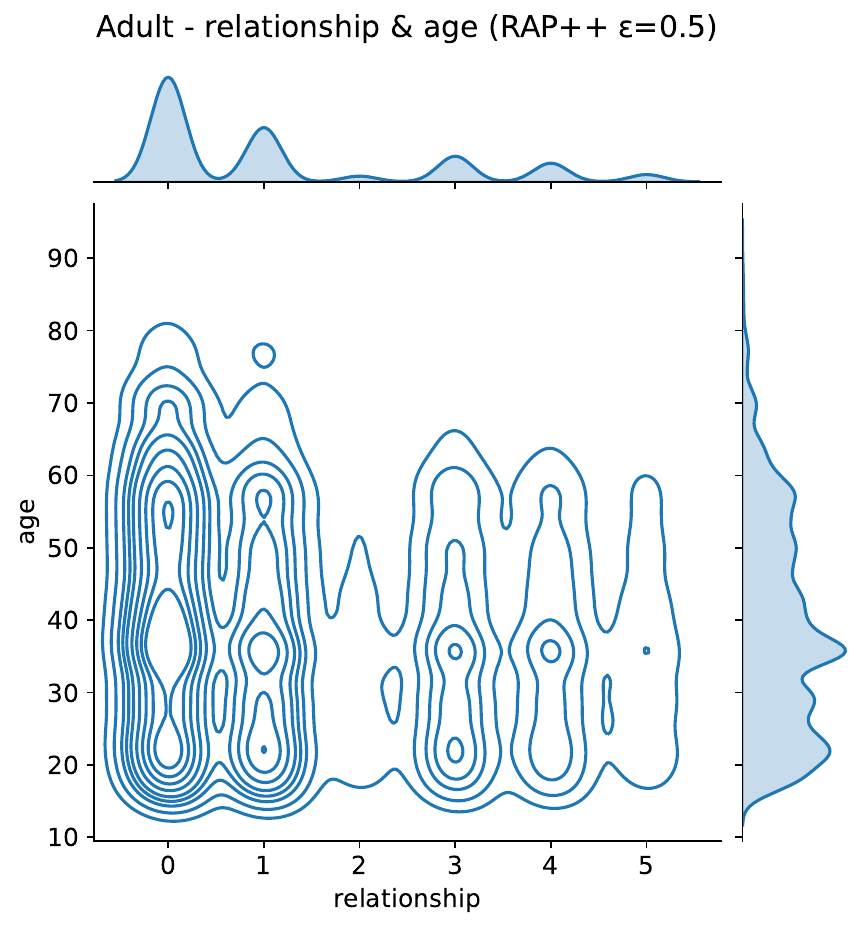}
\end{subfigure} 
\begin{subfigure}{.32\textwidth}
  \centering
  \includegraphics[width=\linewidth]{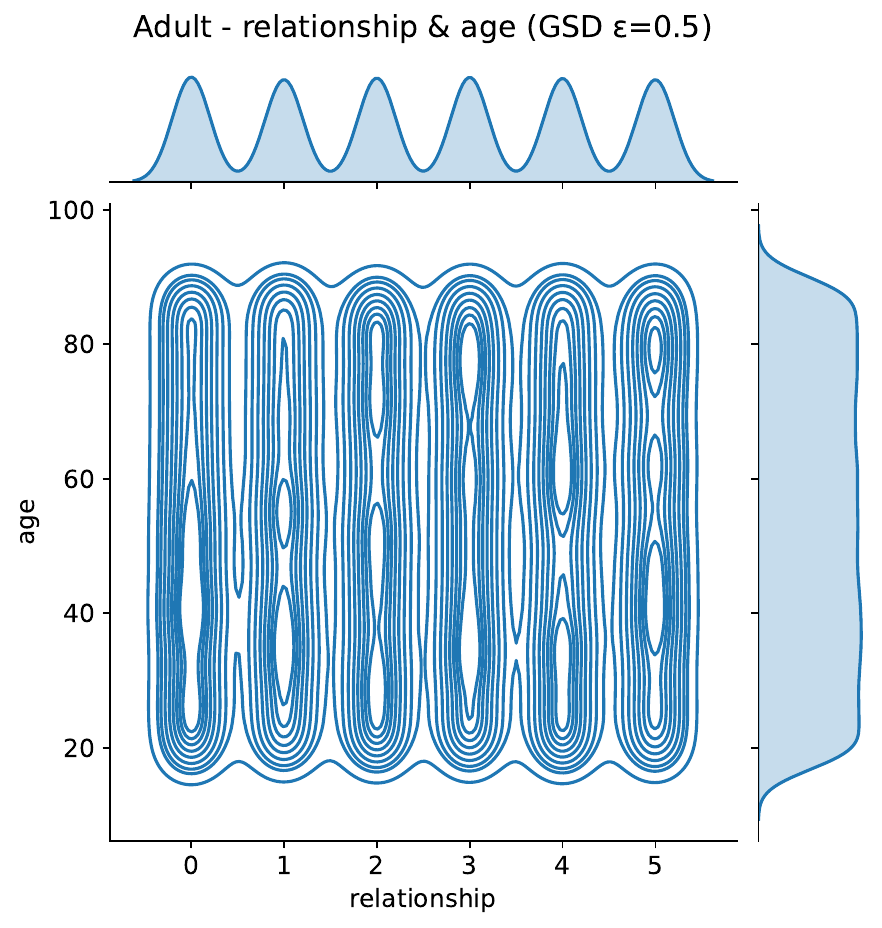}
\end{subfigure}%
\begin{subfigure}{.32\textwidth}
  \centering
  \includegraphics[width=\linewidth]{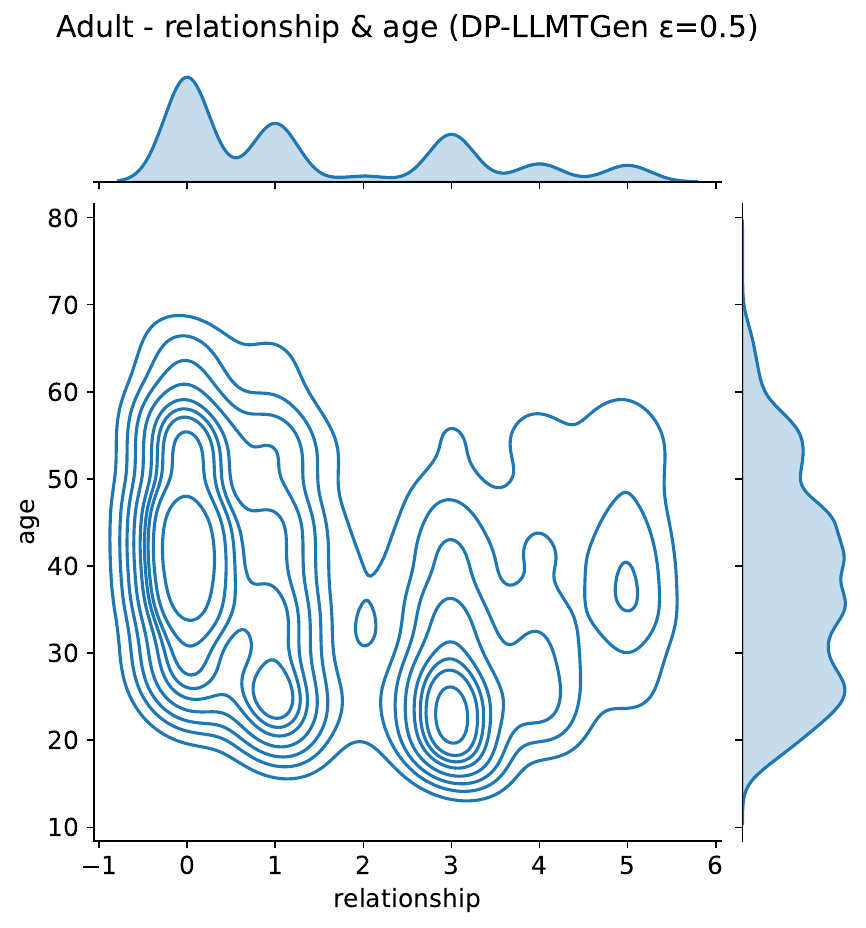}
\end{subfigure} 
\caption{[Adult dataset] Joint distribution plot of two features (age and relationship) of synthetic and real sets.}
\label{fig:relation-age-dist}
\end{figure}

\begin{figure}[h]
\centering
\begin{subfigure}{.32\textwidth}
  \centering
  \includegraphics[width=\linewidth]{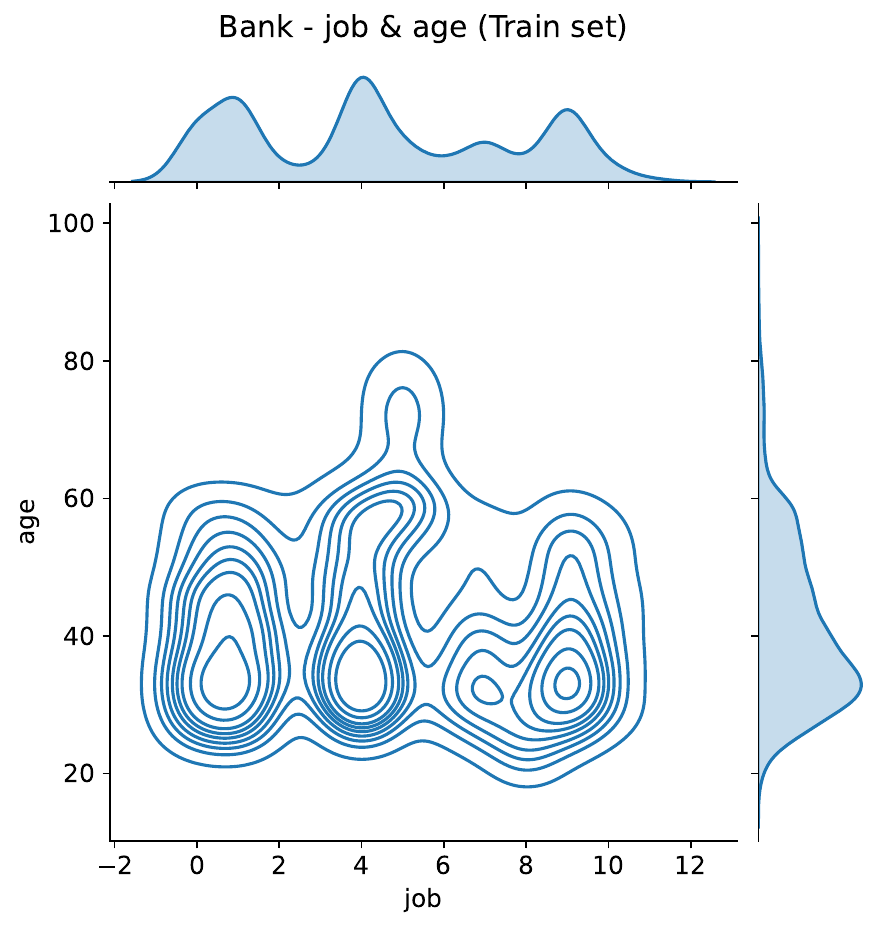}
\end{subfigure}
\begin{subfigure}{.32\textwidth}
  \centering
  \includegraphics[width=\linewidth]{figures/bank/2TVD/job_age_train.pdf}
\end{subfigure}
\begin{subfigure}{.32\textwidth}
  \centering
  \includegraphics[width=\linewidth]{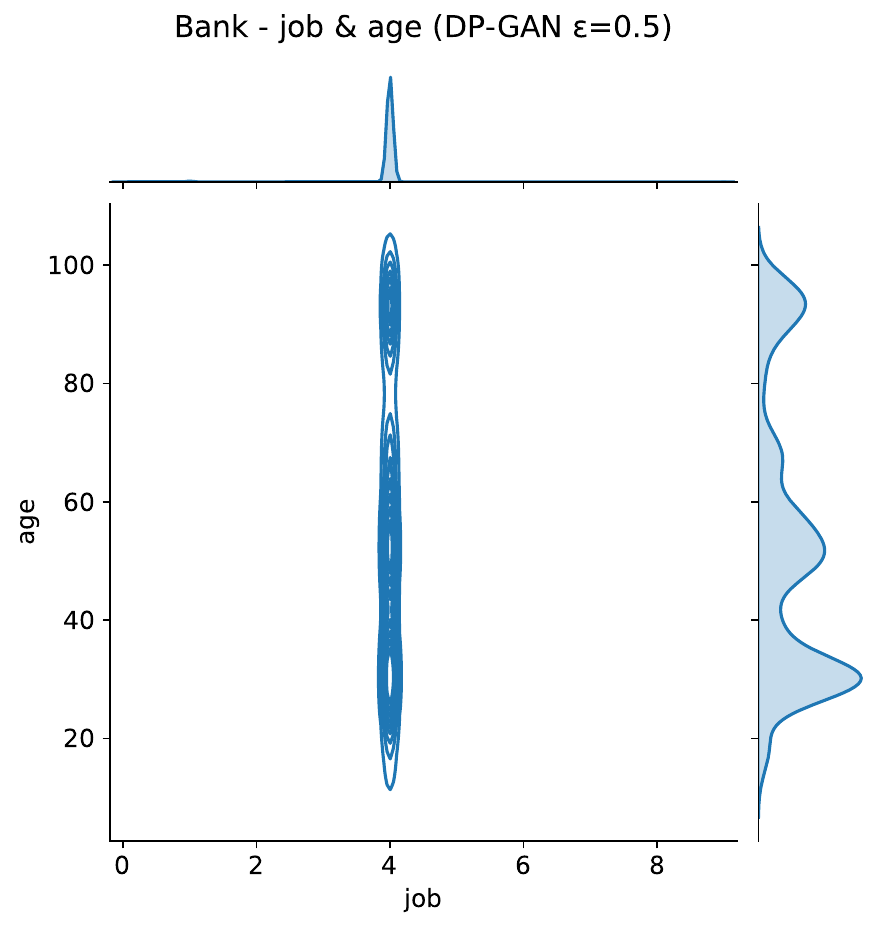}
\end{subfigure} \vspace{5mm}  \\
\begin{subfigure}{.32\textwidth}
  \centering
  \includegraphics[width=\linewidth]{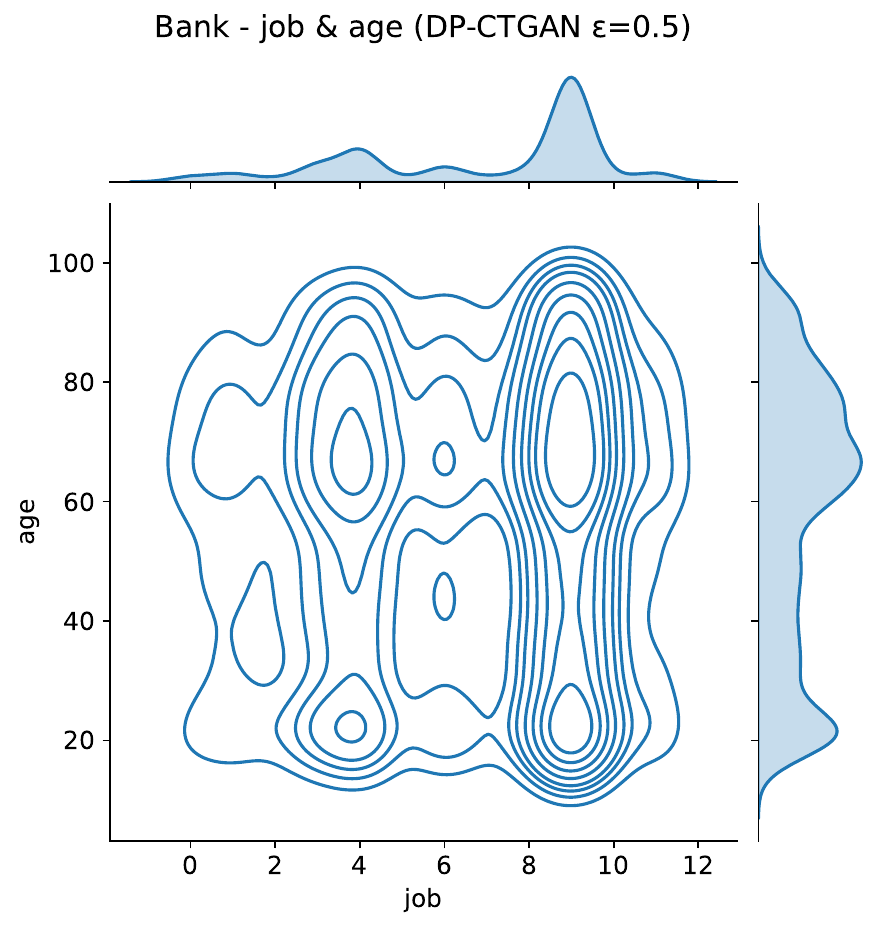}
\end{subfigure}
\begin{subfigure}{.32\textwidth}
  \centering
  \includegraphics[width=\linewidth]{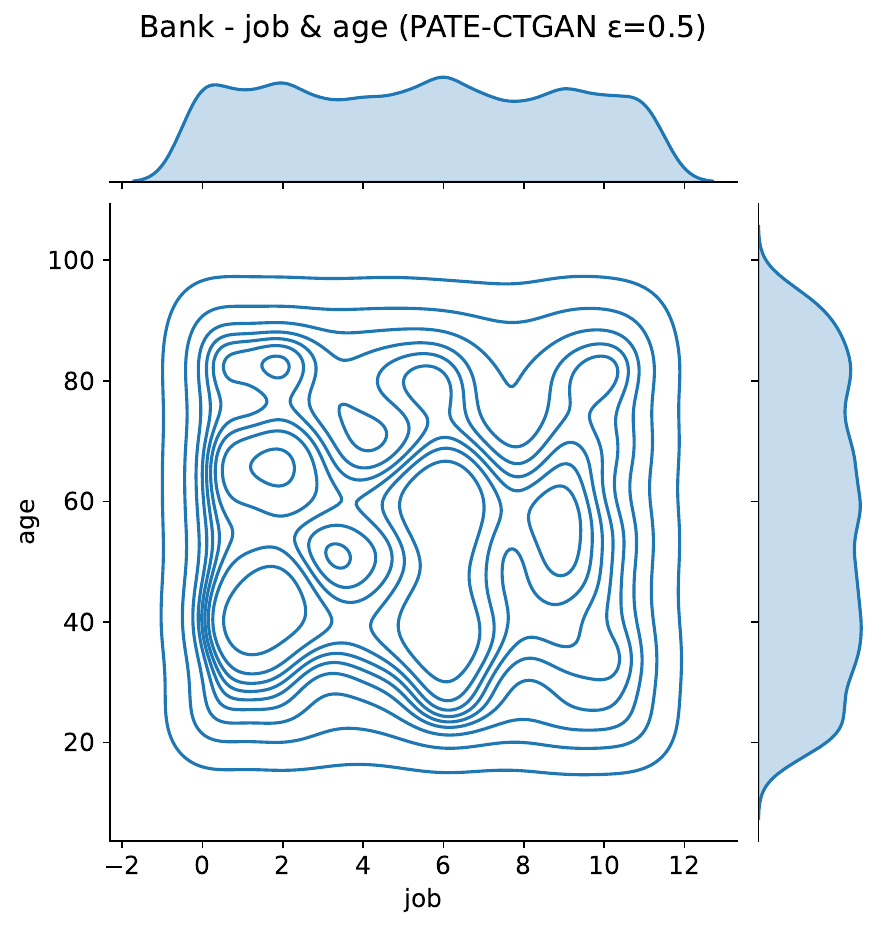}
\end{subfigure} 
\begin{subfigure}{.32\textwidth}
  \centering
  \includegraphics[width=\linewidth]{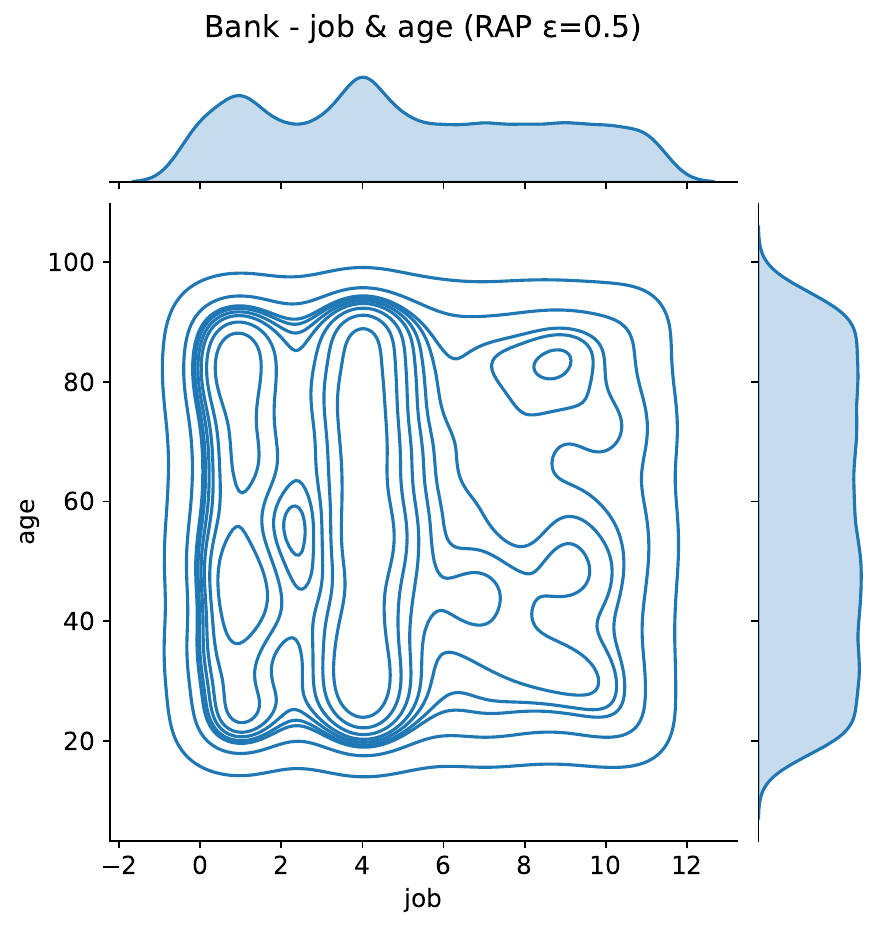}
\end{subfigure} \vspace{5mm} \\
\begin{subfigure}{.32\textwidth}
  \centering
  \includegraphics[width=\linewidth]{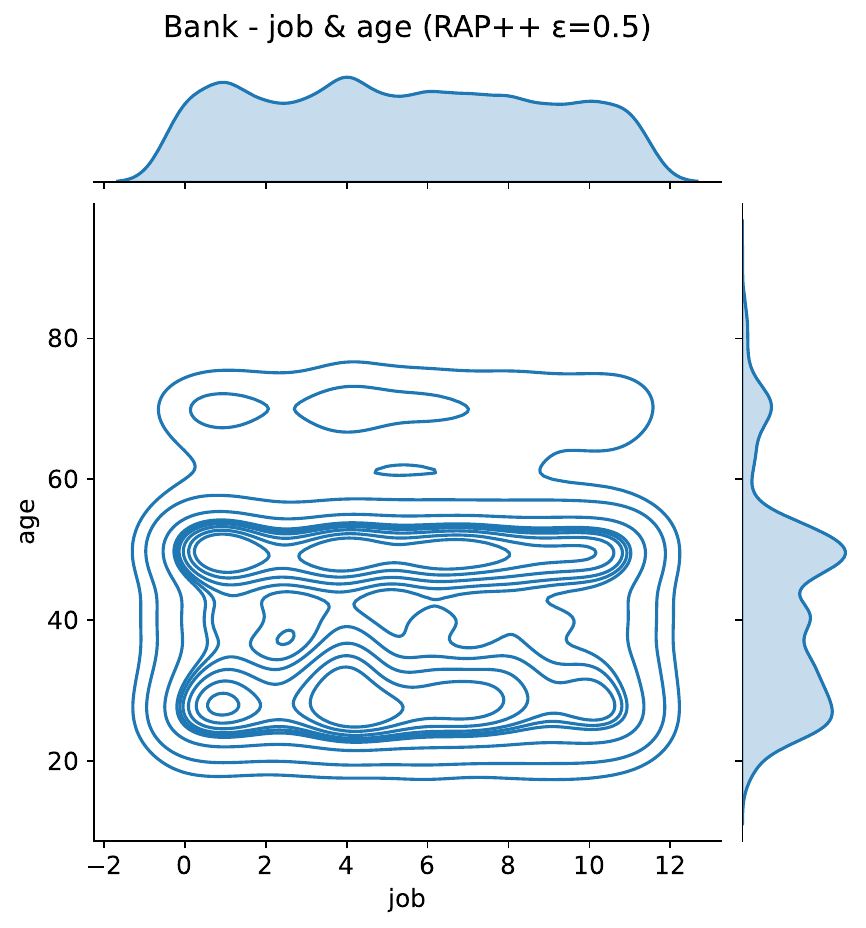}
\end{subfigure} 
\begin{subfigure}{.32\textwidth}
  \centering
  \includegraphics[width=\linewidth]{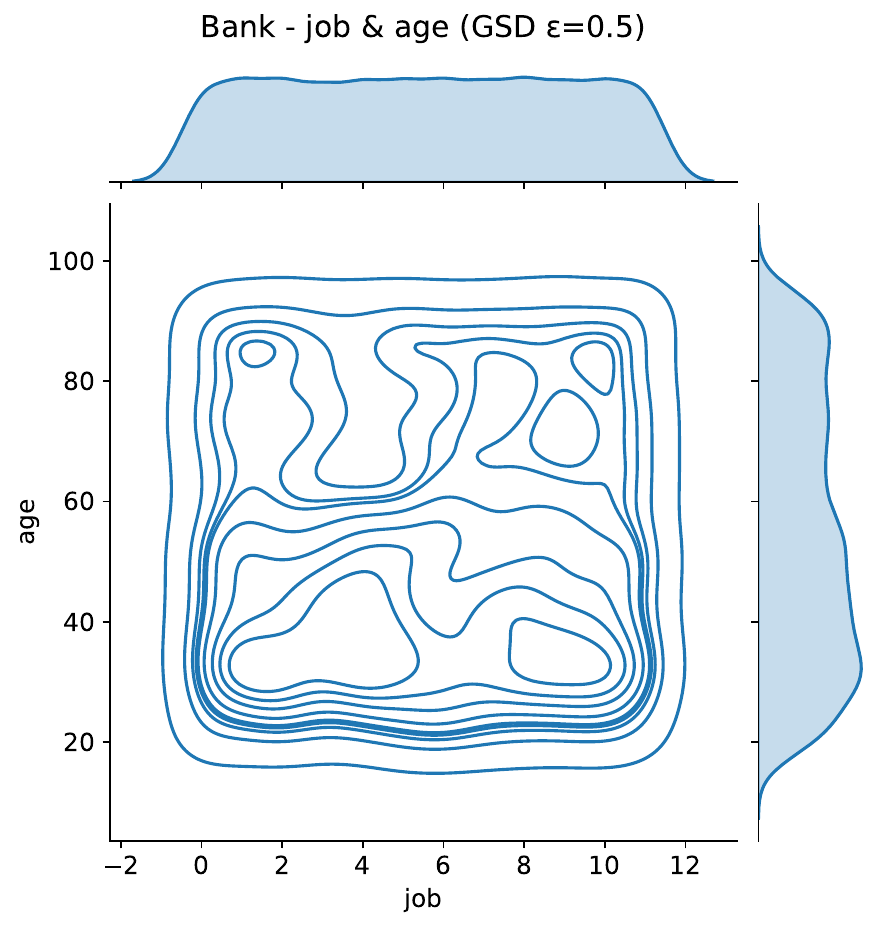}
\end{subfigure}%
\begin{subfigure}{.32\textwidth}
  \centering
  \includegraphics[width=\linewidth]{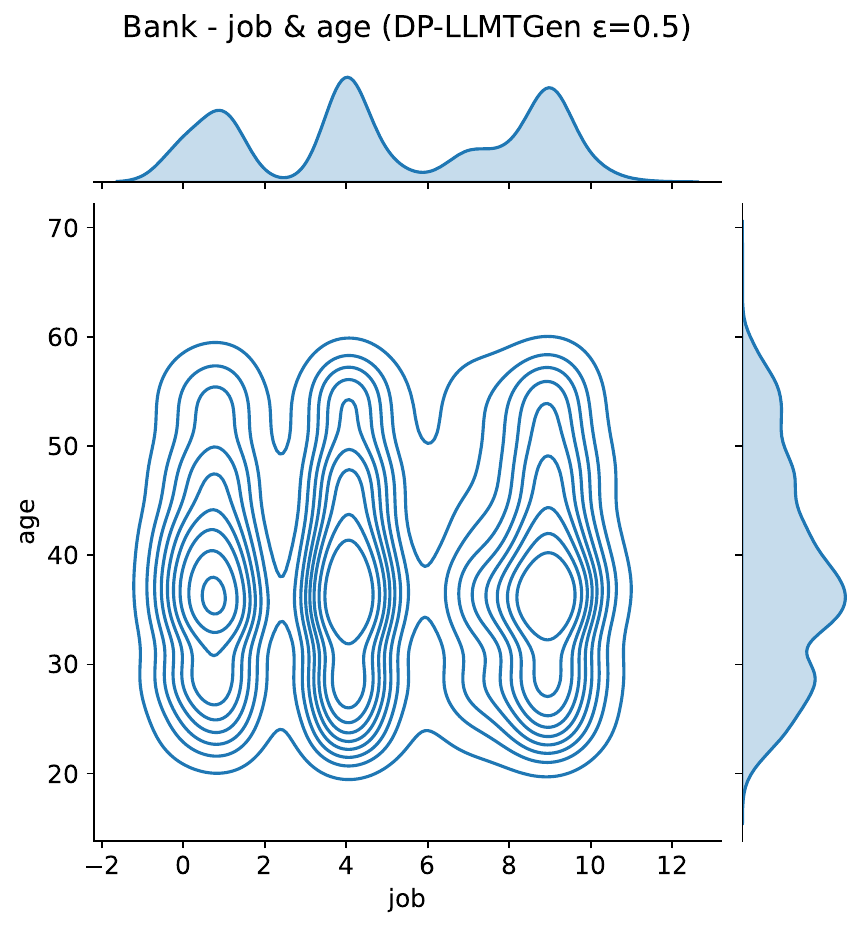}
\end{subfigure} 
\caption{[Bank dataset] Joint distribution plot of two features (age and job) of synthetic and real sets.}
\label{fig:relation-age-dist}
\end{figure}

\clearpage
\section{Additional results for experimental analyses}
Section~\ref{sec:exp-ana} presents some experimental analyses using the Adult subsets with 8192 samples. In this section, we provide results of the same settings but using 4096 samples. Generally, the trends are consistent to those of the previous experiments 8192 samples.

\subsection{Do large language models truly comprehend feature names?}
\begin{figure}[h]
\centering
\begin{subfigure}{.35\textwidth}
  \centering
  \includegraphics[width=\linewidth]{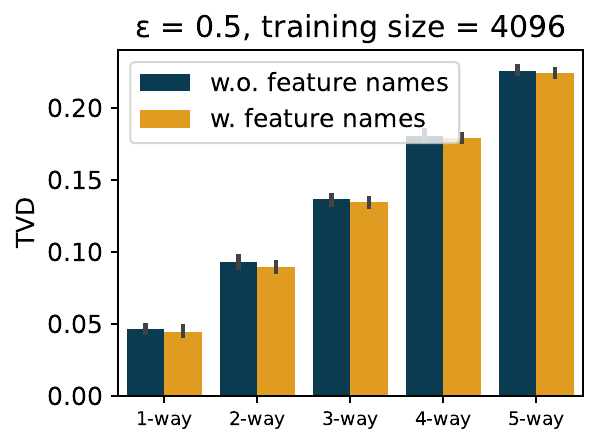}\\
\end{subfigure}%
\hspace{0.5cm}
\begin{subfigure}{.35\textwidth}
  \centering
  \includegraphics[width=\linewidth]{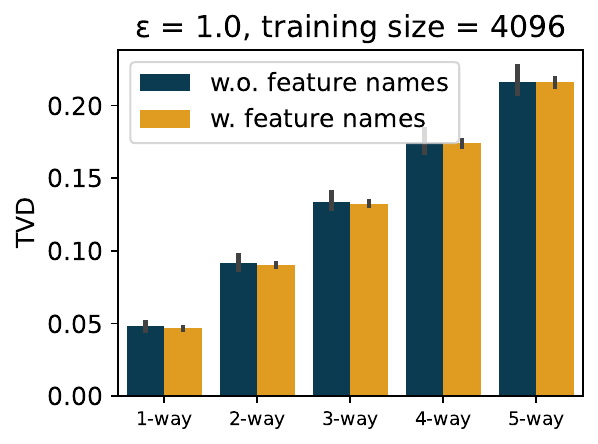}\\
\end{subfigure}%
\caption{Performance of \texttt{DP-LLMTGen} for the Adult subsets of 4096 samples with and without feature names.}
\label{fig:app-noname}
\end{figure}

\subsection{Does dialogue optimizing provide benefit DP-LLMTabGen?}
\begin{figure}[h]
\centering
\begin{subfigure}{.35\textwidth}
  \centering
  \includegraphics[width=\linewidth]{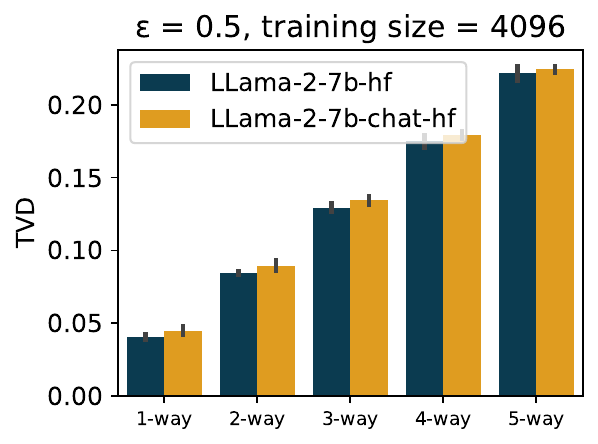}\\
\end{subfigure}%
\hspace{0.5cm}
\begin{subfigure}{.35\textwidth}
  \centering
  \includegraphics[width=\linewidth]{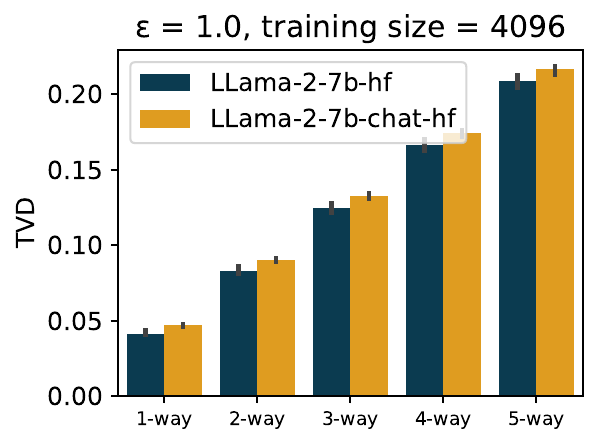}\\
\end{subfigure}%
\caption{Performance of \texttt{DP-LLMTGen} when using Llama-2-7b-hf and Llama-2-7b-chat-hf as the model backbone on the Adult subsets of 4096 samples.}
\label{fig:app-chat-nonchat}
\end{figure}

\subsection{Are Larger LLMs always better?}
\begin{figure}[h]
\centering
\begin{subfigure}{.35\textwidth}
  \centering
  \includegraphics[width=\linewidth]{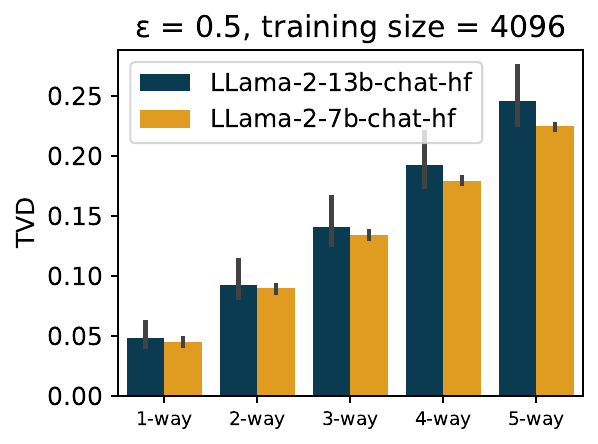}\\
\end{subfigure}
\hspace{0.5cm}
\begin{subfigure}{.35\textwidth}
  \centering
  \includegraphics[width=\linewidth]{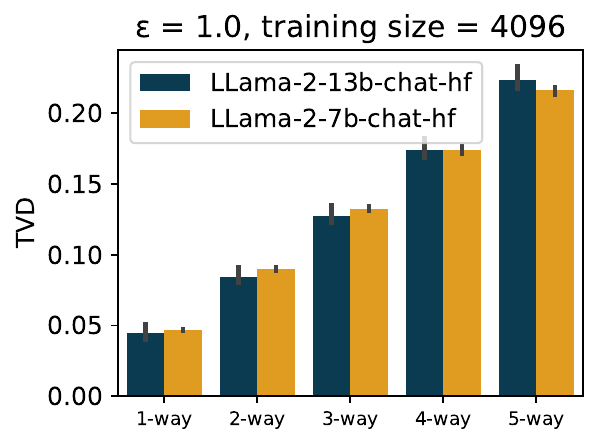}\\
\end{subfigure}%
\caption{Performance of \texttt{DP-LLMTGen} when using Llama-2-7b-chat-hf and Llama-2-13b-chat-hf on three Adult subsets of 4096 samples.}
\label{fig:app-7b-13b}
\end{figure}

\clearpage
\section{Limitations}
\label{sec:app-limit}
\textbf{LLM diversity}. In our previous experiment on the Adult dataset, larger models were not always better. The main reason is the more number of trainable parameters causing bigger gradient clipping factor and larger total noise added. The number of trainable parameters can be adjusted through the parameter-efficient fine-tuning configuration. However, we did not tune this number due to the limitation of computation resources. Therefore, a suitable configuration of parameter-efficient fine-tuning for larger models can provide promising results. Additionally, we acknowledge that there are many other open-source models such as Mistral~\citep{jiang2023mistral} and Gemma~\citep{gemmateam2024gemma}. However, our experiments were mainly based on the Llama-2 models only. It is still unknown about the performance of other model classes for DP tabular data synthesis. Additionally, the higher-quality pre-training corpus can deliver better LLMs given the same number of parameters~\citep{abdin2024phi3}. For example, Phi-3~\citep{abdin2024phi3}, which was pre-trained on a selected corpus (composed of heavily filtered web data and synthetic data), outperformed all other LLMs with the same number of parameters. Llama-3 8b is better Llama-2 70b in multiple benchmarks~\citep{Llama-3}. Given existing evidence that smaller models can deal better with DPSGD training~\citep{remerscheid2022smoothnets}, \texttt{DP-LLMTGen} employing these parameter-efficient LLMs as the model backbone potentially enhances the current performance.

\textbf{ML downstream performance}. \texttt{DP-LLMTGen} using Llama-2 outperformed the baselines in terms of statistical fidelity, but did not entirely surpass them in downstream machine learning performance. Our results exhibited a mismatch between statistical fidelity and ML downstream performance, which is consistent to a previous benchmark~\citep{tao2022benchmarking}. This is because statistical fidelity treats the target feature and other features, equally. However, for ML downstream performance, the target feature can be more important compared to the other features. Only an mistake in the target feature may not lead to a high error of the statistical fidelity but will significantly hurt the ML downstream performance. Currently, \texttt{DP-LLMTGen} does not discriminate target features and other features. Therefore, a possible enhancement could be implementing a weighted loss function to increase the importance of the target feature. Additionally, we evaluated \texttt{DP-LLMTGen} using only the random sampling strategy for the general evaluation. The random sampling does not consider data characteristics and marginal distributions. \texttt{DP-LLMTGen} also offers controllable generation, which has been demonstrated in our fairness-aware experiment. To improve the ML performance, it is possible to generate synthetic datasets with diversity and uncertainty controls by leveraging the controllable generation capability of \texttt{DP-LLMTGen}. Another potential approach is to combine \texttt{DP-LLMTGen}'s controllable generation with active learning or shaley-value methods which can determine which are worthy samples.

\end{document}